\documentclass{article}
\usepackage{for_arxiv,times}

\usepackage{microtype}
\usepackage{graphicx}
\usepackage{subfigure}
\usepackage{booktabs} 
\usepackage{placeins}

\usepackage{hyperref}

\usepackage{amsmath}
\usepackage{amssymb}
\usepackage{mathtools}
\usepackage{amsthm}

\usepackage[capitalize,noabbrev]{cleveref}

\usepackage{color}

\usepackage{array}
\usepackage{multirow}
\usepackage{transparent}
\usepackage{pgf}

\newcommand{\R}{\mathbb{R}}
\newcommand{\E}{\mathbb{E}}
\newcommand{\ve}[1]{\boldsymbol{#1}}
\newcommand{\1}{\mathbf{1}}

\newcommand{\grad}{\nabla}

\DeclareMathOperator{\arctanh}{arctanh}

\theoremstyle{plain}
\newtheorem{theorem}{Theorem}[section]

\newtheorem{corollary}[theorem]{Corollary}

\theoremstyle{definition}
\newtheorem{definition}[theorem]{Definition}

\newtheorem{example}[theorem]{Example}

\theoremstyle{remark}

\iclrfinalcopy

\title{Emergence of Computational Structure in a Neural Network Physics Simulator}

\author{Rohan Hitchcock \\
University of Melbourne \& CSIRO Data61 \\
rhitchcock@student.unimelb.edu.au 
\And
Gary W. Delaney \\
CSIRO Data61 \\
gary.delaney@data61.csiro.au
\AND 
Jonathan H. Manton \\
University of Melbourne \\
jmanton@unimelb.edu.au
\And 
Richard Scalzo \\
CSIRO Data61 \\
richard.scalzo@data61.csiro.au
\And 
Jingge Zhu  \\
University of Melbourne \\
jingge.zhu@unimelb.edu.au
}

\begin{document}

\maketitle

\begin{abstract}
    Neural networks often have identifiable computational structures --- components of the network which perform an interpretable algorithm or task --- but the mechanisms by which these emerge and the best methods for detecting these structures are not well understood. In this paper we investigate the emergence of computational structure in a transformer-like model trained to simulate the physics of a particle system, where the transformer's attention mechanism is used to transfer information between particles. We show that (a) structures emerge in the attention heads of the transformer which learn to detect particle collisions, (b) the emergence of these structures is associated to degenerate geometry in the loss landscape, and (c) the dynamics of this emergence follows a power law. This suggests that these components are governed by a degenerate ``effective potential''. These results have implications for the convergence time of computational structure within neural networks and suggest that the emergence of computational structure can be detected by studying the dynamics of network components.
\end{abstract}

\section{Introduction}

Computational structures, by which we mean components of a neural network which perform 
an interpretable algorithm or task, often emerge in neural networks during training. 
Understanding how and why these structures form would contribute to making large neural networks 
more interpretable, and may enable greater control over what is learned and not learned 
during training.
Clear examples of computational structure have been found in 
large language models \citep{olsson_induction_heads_2022}, vision 
models \citep{cammarata2020circuits_thread} and in toy systems such as \cite{nanda_progress_2023}. 
While some recent progress has been made towards understanding how these emerge \citep{hoogland_ICL1_2024,wangDifferentiationSpecializationAttention2024} the mechanism by which this occurs and the best methods for 
detecting such structures are still open questions. 

In this paper we investigate the emergence of computational structure in a transformer-like
model for simulating the physics of a system of particles under gravity. During training, computational structures emerge 
which have a clear connection to the physical laws and driving behaviour of the target system. 
Specifically we identify 
certain attention heads that learn collision detection between particles, which we call \emph{collision detection heads}. 

Our experiments suggest that the emergence of collision detection behaviour in attention heads coincides with: (a) 
a change in the head's training dynamics, and (b) a change in the geometry of the 
loss landscape near the head's parameters. For (a) we observe that the correlation between 
certain features of the data and the attention scores begins 
to evolve as a power law. For (b) we observe changes in the degeneracy of the loss landscape as measured 
by the local learning coefficient \citep{lauLocalLearningCoefficient2024}.
The emergence of a collision detection head is shown in \cref{fig:heads governed by degenerate potential}. This suggests that the emergence of collision detection heads 
is governed by a degenerate
``effective potential''. 
The specific contributions of this paper are:

\begin{figure}[t]
    \begin{center}
    \includegraphics[width=\linewidth]{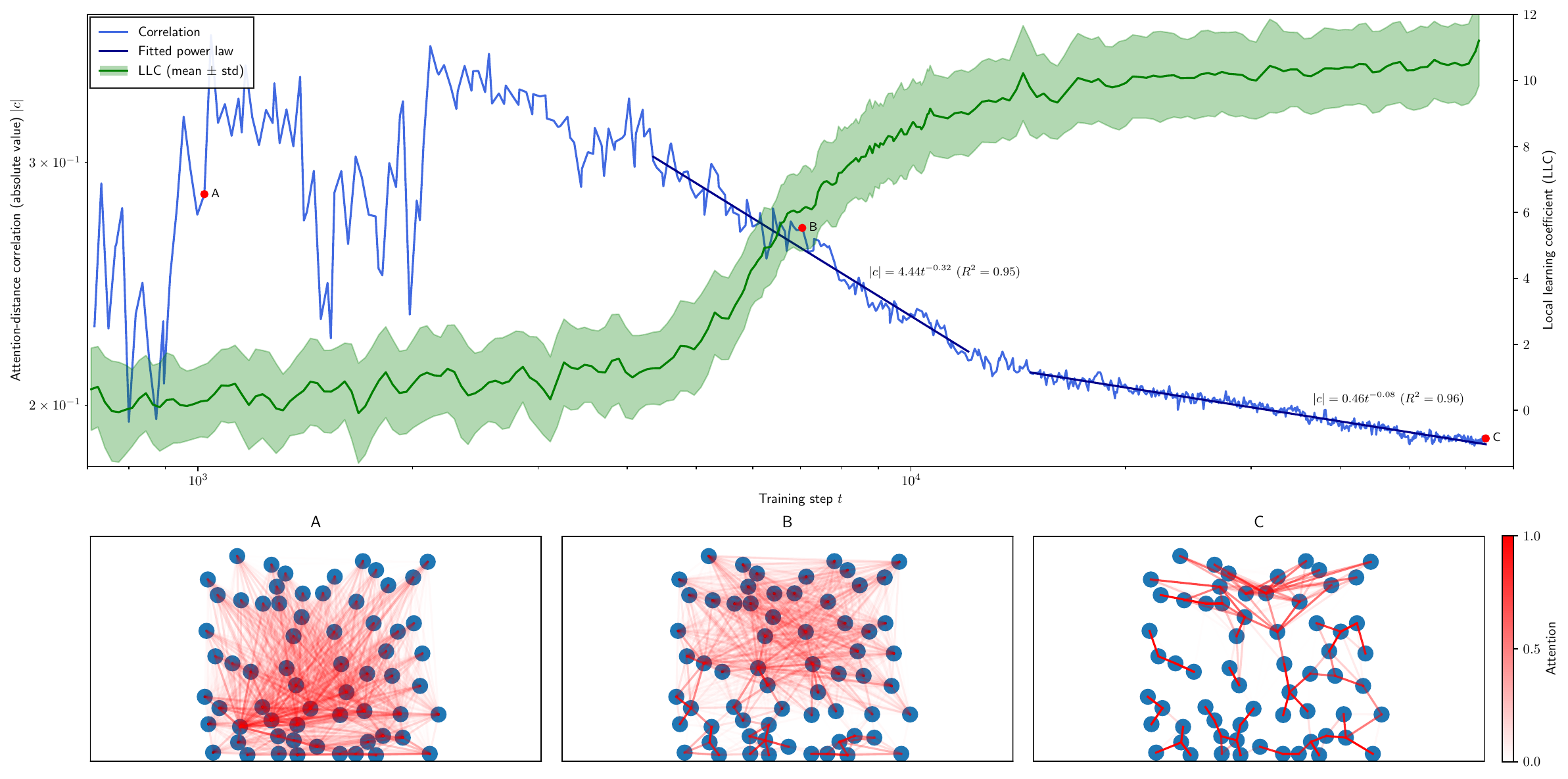}
    \end{center}
    \caption{In collision detection heads we see three simultaneous phenomena: collision 
    detection behaviour emerges (bottom), the attention-distance correlation curve 
    enters a power law regime (top, blue) and the local learning coefficient 
    plateaus (top, green). In this head, we see a transition between 
    two distinct phases of collision detection.
    The behaviour in the first collision detection phase is characterised by collision 
    detection between particles below a certain $y$-level (bottom, B), as
    compared to mostly universal collision detection in the second phase (bottom, C). 
    This transition is reflected in the attention-distance correlation curve as different 
    power law exponents, visualised as different slopes on the log-log plot (top, blue). 
    The local learning coefficient 
    plateaus during the first collision detection phase before continuing up to its 
    final value.} 
    \label{fig:heads governed by degenerate potential}
\end{figure}

\begin{itemize}
    \item It elicits a connection between degeneracy and computational structure in transformer-like models, 
    and the \emph{dynamics} of these components. This has implications for the convergence 
    time of these components: if they evolve as a power law then structures which form earlier 
    in training will be significantly better formed than those which form later. It also suggests that measurements of the dynamics of model 
    components may be a good way to detect the emergence of computational structure in deep learning 
    models. 
    \item It provides evidence for the role of degeneracy and the formation of computational 
    structures in neural networks trained to simulate physics. The connection here is especially interesting 
    as the phenomena in the model associated with the emergence of collision detection heads 
    (power law behaviour and changes in the degree of degeneracy) are closely connected to 
    \emph{second-order phase transitions} in statistical physics.
    This interpretation of our results suggests viewing training neural network physics 
    simulators as a coupled system, where the parameters of the network are coupled to the 
    dynamics of the target system.
\end{itemize}

A running theme through this paper is the effect of degenerate 
critical points on dynamics and the associated emergence of power laws. 
We give a self-contained exposition of this in 
\cref{sec: power laws and degeneracy}.

\subsection{Related work} \label{sec:related work}

\subsubsection{The role of degeneracy in learning}

Complex learning machines like neural networks 
have \emph{degeneracy} in the form of degenerate critical points 
(saddle points or local minima at which the Hessian is singular) in the 
loss landscape or log-likelihood function. 
The importance of this degeneracy in the Bayesian setting has been well-understood 
for some time \citep{watanabe_algebraic_2009}. In strictly singular models 
(including all non-trivial neural network architectures) the minima of the 
log-likelihood function 
are degenerate and non-isolated. \citet{watanabe_algebraic_2009} showed that the 
degenerate geometry of this log-likelihood function, in particular the 
\emph{real log canonical threshold} (RLCT)
\citep[see][Definition 2.1]{watanabe_algebraic_2009},
determine much of the learning behaviour of strictly singular models.

These ideas have recently been applied to neural networks trained using stochastic 
gradient methods. \citet{lauLocalLearningCoefficient2024} establishes the \emph{local 
learning coefficient} (LLC) as a way to quantify the degeneracy in the loss landscape
and proposes a method for estimating the LLC. The LLC is derived 
from results in the Bayesian setting
and under certain conditions it coincides with a RLCT 
\citep{lauLocalLearningCoefficient2024}. 

The LLC has been applied to study transformers trained on natural language and on 
a synthetic in-context regression task \citep{hoogland_ICL1_2024,wangDifferentiationSpecializationAttention2024}. The LLC is shown to identify periods, or \emph{phases}, of training where the model has 
qualitatively different behaviour. There are similar results in toy models 
\citep{chen_dynamical_2023} and in deep linear networks \citep{lauLocalLearningCoefficient2024}. 

Separately, distinct phases of training have also been identified 
by considering the emergence of components of neural networks which perform a 
certain algorithm. These emergent computational structures are often termed 
\emph{circuits}, and they have been identified in models such as large language models 
\citep{olsson_induction_heads_2022}, vision models \citep{cammarata2020circuits_thread} 
and in toy models trained to do modular addition \citep{nanda_progress_2023}. 
In \citet{hoogland_ICL1_2024} the emergence of certain circuits in language models 
is shown to coincide with changes in the local geometry of the loss landscape 
as measured by the LLC.

\subsubsection{Statistical mechanics as a framework for understanding learning}

Statistical mechanics has been used as a theoretical framework to understand 
statistical inference and learning; see for example the reviews and textbooks 
\cite{nishimori_statistical_2001,mezard_information_2009,zdeborova2016statistical,bahri_statistical_2020,huang_statistical_2021}. 
Ideas from statistical mechanics have been applied to neural networks in early work
\cite{seung1992statistical,watkin_statistical_1993} and more recently 
\cite{choromanska2015loss,geiger_jamming_2019,mei_mean_2018,spigler_jamming_2019}. 
When applied to neural networks many of these ideas do not fully account for the 
degenerate nature of the loss landscape. 

Analogies with statistical mechanics are clear in the approach of \citet{watanabe_algebraic_2009}
and are drawn-out in \citet{lamont2019correspondence}.
Analysis of Bayesian learning in \citet{watanabe_algebraic_2009} 
is organised around the concept of the \emph{free energy}
of a statistical model applied to a learning task. This is analogous to the concept 
of the same name from statistical mechanics. A deep theoretical result in 
\citet{watanabe_algebraic_2009} is an asymptotic expansion of the free energy in 
terms of the RLCT of the log-likelihood function. 

\subsubsection{Learned particle simulation}
The transformer-like architecture we use here is similar to a graph neural 
network, particularly a graph attention based architecture \citep{velickovic_gat_2018,brody_gatv2_2022}. 
Graph neural networks (though not using graph attention) have been used 
to simulate particle systems in 
\citet{mei_micro-_2022,zhang_estimation_2023,li_prediction_2023,choi_graph_2024}. 
Graph neural networks have also been used to simulate other physical systems such as in 
\citet{sanchez2020graph_pinns}.

Other than graph neural network approaches, convolutional methods have also been 
used to learn interactions between particles. \citet{lu_machine_2021,mital_bridging_2022} 
train models based on a convolutional layer originally conceived in \citet{ummenhofer_lagrangian_2020}
for fluid simulation. \cite{lai_machinelearningenabled_2022} considers fully connected networks 
trained to detect and resolve collisions between pairs of particles with complex shapes.

\section{Method} \label{sec:method}

We train a model to simulate a system of particles where collisions are governed 
by damped linear spring dynamics \citep[see][]{luding2008introduction}. Given the 
state of such a particle system, the model is trained to predict the state 
of the system after some fixed time interval $\Delta t$. Repeatedly 
applying the model to its own output means the fully trained model can be used 
as a simulator. 

\subsection{Model}

Our model architecture is based on a transformer using multihead attention; 
the main difference is that the 
attention mechanism is used to transfer information between particles rather than 
between elements of a sequence. 
Full details of the model architecture and training process 
are given in \cref{sec:model and training details}. We completed a total of five 
training runs with this model.

\subsection{Understanding the role of attention heads}

To understand how attention heads contribute to detecting collisions between particles
we introduce two measurements of an attention head: the \emph{collision detection score}
and the \emph{attention-distance correlation function}. 

For a given attention head we define its \emph{collision detection score} as the expected 
attention the head assigns within a small, fixed radius of each query particle. The collision 
detection score ranges falls between 0 (for a head that never assigns attention to nearby 
particles) and 1 (for a head which assigns all of its attention to nearby particles). 
The distribution of collision detection scores evolving over training is shown in 
\cref{fig:contact scores}.  

The collision detection score aims to straightforwardly quantify the tendency of an attention head to detect collisions and is used for classifying whether or not a head has collision detection behaviour. We also visualise a head's attention pattern for a given particle state by drawing a 
line between every pair of particles, weighted by the assigned attention score (see \cref{fig:head behaviour}).
These visualisations clearly show when an attention head is detecting particle collisions, 
and qualitatively agree with the collision detection score. 

For each attention head we measure the statistical correlation between its attention scores 
and the distance between particles. Let $\ve w \in \R ^d$ be the parameters of the neural 
network (for example at a checkpoint over training) and consider a single fixed 
attention head within the network. For each pair of particles $i$ and $j$, let 
$\alpha _{ij}(\ve w)$ be the attention score assigned by the head to the 
representation of particle $j$ when updating the representation of particle $i$. We 
consider the correlation between $\alpha_{ij}(\ve w)$ and the distance 
$d_{ij} = \| \ve p_i - \ve p_j\|$ between particles $i$ and $j$.
The \emph{attention-distance correlation function} for the given head 
is
\[
c(\ve w) = \frac{1}{N^2-N} \sum _{i\neq j} \text{corr}(d_{ij}, \alpha _{ij}(\ve w)) ~.
\]
where $\text{corr}(d_{ij}, \alpha _{ij}(\ve w))$ is computed by taking expectations 
with respect to the data distribution.

The attention-distance correlation 
function is inspired by correlation functions in statistical mechanics,
which are often able to detect qualitative changes in a physical system's behaviour. 
We use the dynamics of $c(\ve w)$ --- meaning its behaviour as a function of the training 
step --- as a proxy for the dynamics of a collision detection head (this is discussed further 
in \cref{sec:discussion implied dynamics}, see also 
\cref{corr: holder continuous measurement also evolves as power law}). 
When $c(\ve w)$ is tracked over training we will see that these dynamics 
can be used to identify collision detection heads. 

\subsection{The geometry of attention head weights}

To understand the geometry of the attention head weights we use a parameter-restricted
version of the local learning coefficient (LLC) of 
\citet{lauLocalLearningCoefficient2024}. 
The same method is used to study attention heads in language transformers in 
\citet{wangDifferentiationSpecializationAttention2024}.

The LLC at a parameter vector $\ve w \in \R^d$
is a measurement of the degree of degeneracy of the loss landscape near $\ve w$. We give 
a more detailed exposition of the local learning coefficient in 
\cref{sec:llc theoretical formulation}, but for some intuition the term `degeneracy' is used in 
the same sense as a degenerate critical point; when $\ve w$ is a local minimum of the 
loss function then the LLC will be maximal if the Hessian is non-singular at $\ve w$. 
Smaller values of the LLC indicate that $\ve w$ is more degenerate minimum, in a 
sense which is closely related to the real log canonical threshold (a geometric invariant). When at a local minimum, the LLC can also be understood as relating 
to the rate of volume scaling in the loss landscape: a smaller local learning coefficient means that a greater volume of nearby 
parameters will achieve the same expected loss within a small tolerance $\epsilon$
(see \citet{hoogland_model_complexity_2023,lauLocalLearningCoefficient2024} for more details). 

To understand the geometry of a single attention head we treat all parameters 
other than those in the attention head as fixed and perform LLC estimation 
using the method described in 
\citet{lauLocalLearningCoefficient2024,wangDifferentiationSpecializationAttention2024}. 
This lets us understand the geometry of the loss function considered only as 
a function of the parameters of a single attention head. 
We give specific details about our procedure for estimating the LLC in \cref{sec: llc}. For heads of interest we estimate the parameter-restricted LLC at 
many checkpoints over training.

\section{Results} \label{sec:results}

\begin{figure}[t]
    \begin{center}
        \includegraphics[width=0.85\linewidth]{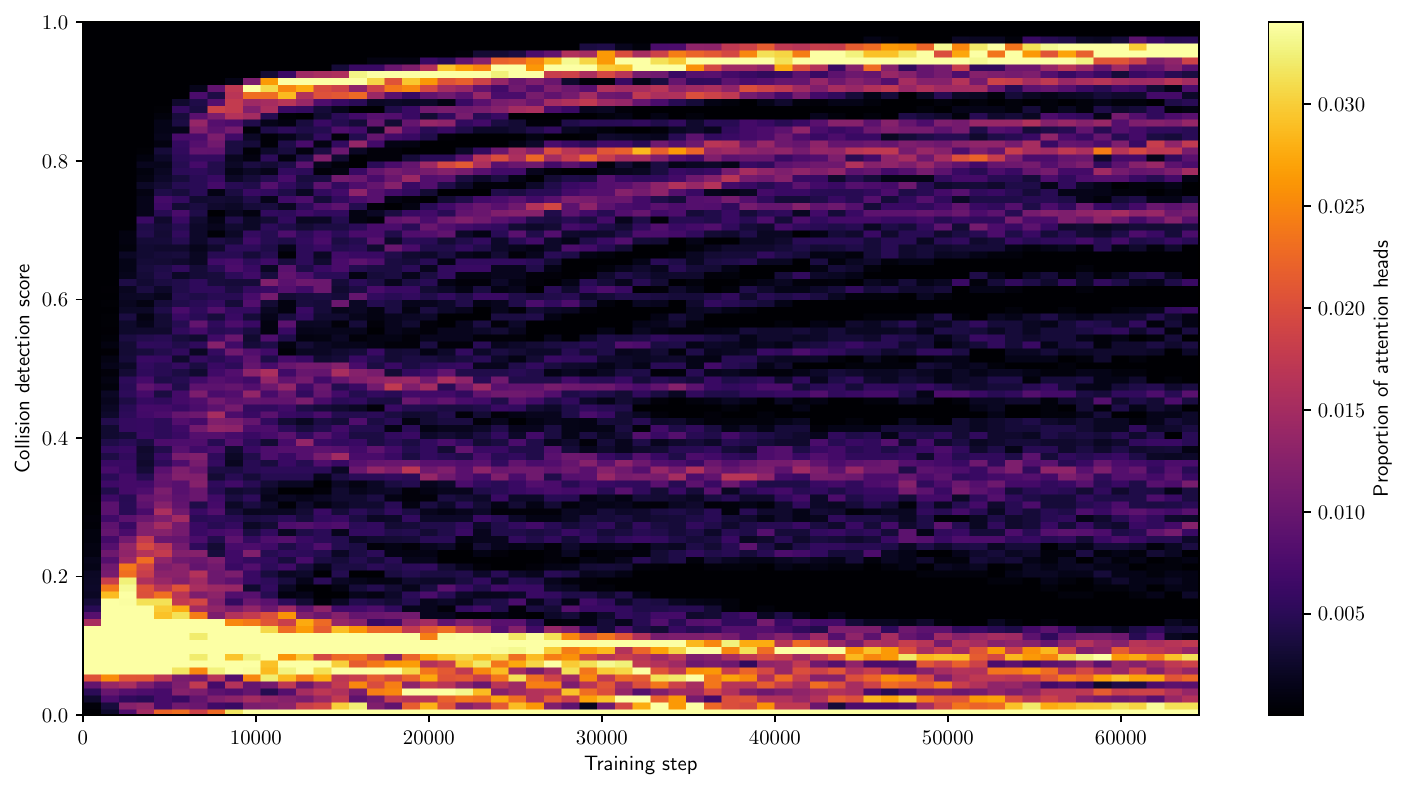}
    \end{center}
    \caption{We track the collision detection score for all attention heads over training
    across all five training runs and aggregate them into a two-dimensional histogram. 
    At the end of training we see a dense cluster of attention heads with collision detection score near $0.95$ and a dense cluster near zero, corresponding respectively to 
    true collision detection heads and heads without collision detection behaviour. Between 
    the two large clusters there are several small clusters corresponding to the 
    partial collision detection heads. To improve the dynamic range of this plot, 
    all density values in the 95\textsuperscript{th} percentile are set to the same 
    maximal colour intensity. Versions of this plot showing only the attention heads 
    from a single training run are given in \cref{sec:additional_contact_score_plots}} 
    \label{fig:contact scores}
\end{figure}

Over the course of training we observe that some attention heads learn to detect 
collisions between particles. We call these heads \emph{collision detection heads}. 
We study how these collision detection heads develop over training. We observe that 
in most cases the following occur simultaneously: (a) collision detection behaviour emerges, 
(b) the attention-distance correlation enters a power law regime, (c) the parameter-restricted 
local learning coefficient plateaus.
These three phenomena are visualised in \cref{fig:heads governed by degenerate potential}. 

\subsection{Collision detection heads}

At the end of training we identify attention heads which assign the majority of their 
attention to particles which are at or near the point of collision, which we call these 
\emph{collision detection heads}. The distribution of the heads' 
collision detection scores at the end of training (\cref{fig:contact scores}) shows several 
clusters of heads with similar collision detection scores. 

\begin{figure}[t]
    \begin{center}
    \includegraphics[width=\linewidth]{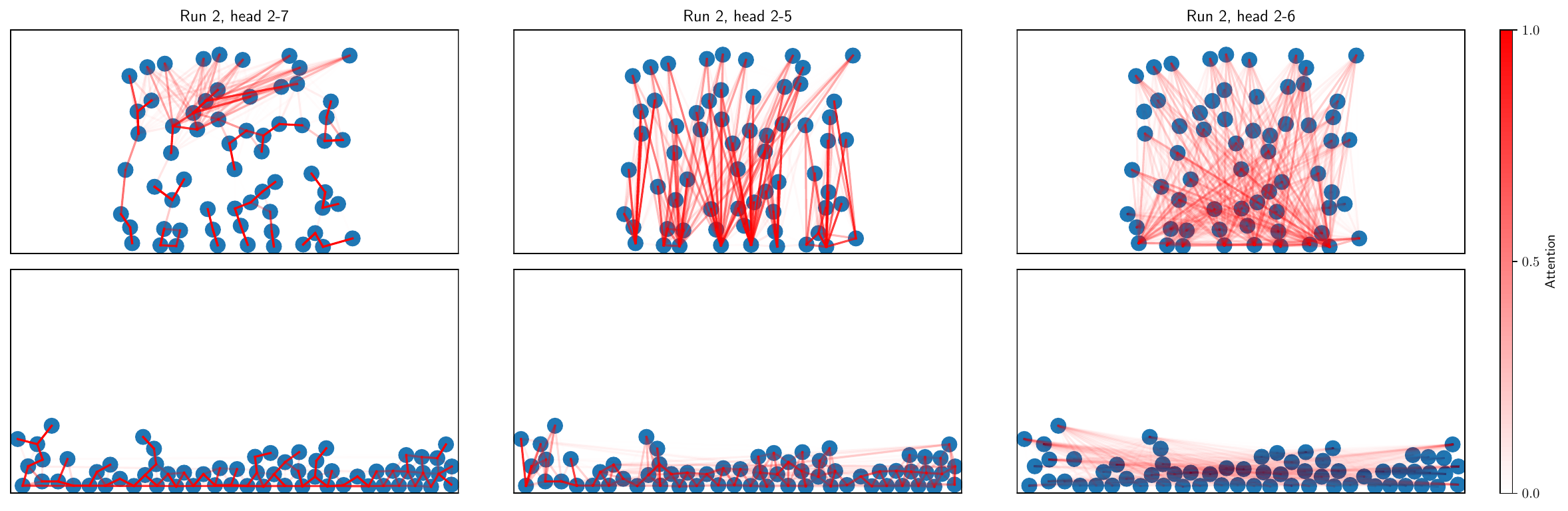}
    \end{center}
    \caption{From left to right, examples of a true contact 
    detection head, a partial collision detection head, and a head which is neither. These have collision detection scores 0.95, 0.78 and 0.02 respectively. We show the heads' behaviour when the 
    particles are in free-fall (top) and when the particles are close to 
    settled (bottom). The attention score assigned by particle $i$ to 
    particle $j$ is indicated by a red line with opacity proportional to 
    the attention score.} \label{fig:head behaviour}
\end{figure}

There is a large cluster of heads 
with a collision detection score of about 0.95, which we refer to as \emph{true collision 
detection heads}. An example of the attention pattern of a true collision detection head 
is shown in \cref{fig:head behaviour}. This visualisation shows 
that the dominant behaviour\footnote{This head has 
some non-collision detection behaviour among a small group of particles near the top of the 
system. All true collision detection heads have some degree of anomalous behaviour,
often for particles near the top of the system.
We don't have a good explanation 
for why this occurs.} of the head is to detect collisions between particles in both particle 
states. We also have a large cluster of attention heads with collision detection 
score close to zero which do not exhibit collision detection behaviour. An example 
attention pattern for such a head is shown in \cref{fig:head behaviour}.

In between these 
two large clusters there are a number of smaller clusters of heads with collision detection 
score between approximately 0.3 and 0.8. We call these \emph{partial collision detection 
heads}. Inspecting the attention patterns for heads in these clusters 
shows that these heads exhibit collision detection behaviour, but only under certain circumstances. 
For example, the partial collision detection head shown in \cref{fig:head behaviour} seems to 
only detect collisions between particles when the particles are in a settled state. 

All training runs develop a mixture of true and partial collision detection heads (see figures in 
\cref{sec:additional_contact_score_plots}), and these heads develop 
in all transformer blocks except the first block (see \cref{fig:head-counts-by-block}, \cref{sec: additional results}). Additional visualisations of head behaviour are shown in \cref{sec:additional-attenrtion-plots}.

\subsection{Attention-distance correlation power laws and the LLC}

We study the development of collision detection heads by looking at how the attention-distance 
correlation function evolves over training.
For collision detection heads the attention-distance correlation 
evolves as a power law. To see this, we plot the attention-distance 
correlation $c(\ve w_t)$ verses the training step $t$ on a log-log plot. Power law sections
can be identified as sections of $\log|c|$ vs $\log t$ with a good 
linear fit.
At the end of its development 
(i.e. at the end of training or before the correlation function stops changing)
the correlation function of a collision detection head evolves as a power law
(see \cref{fig:heads governed by degenerate potential}). This occurs in all but 
six cases. Three of these cases 
have a `power law' with a positive exponent, and three have no obvious power law behaviour. 
Of these six anomalous collision detection heads, 
four appear within the same block for the same training run. There are no attention 
heads for which the attention-distance correlation follows a power law which are not 
collision detection heads.

The emergence of power law behaviour in the attention-distance cross-correlation 
function coincides with changes in the local geometry of the head's parameters, 
as measured by the local learning coefficient. We observe that the local learning 
coefficient plateaus in sections of training where the cross-correlation function 
is governed by a power law. 

\subsection{Transitions between head variants} \label{sec:results transition head variants}

In the case shown in \cref{fig:heads governed by degenerate potential}, we observe a head 
transition between two variants of a collision detection head. Initially a partial collision 
detection head forms: the head detects collisions between particles only near the lower physical 
boundary of the particle system as shown in box B. This is associated with the attention-distance 
correlation function entering 
a power law regime and the LLC plateauing briefly (note that such plateaus in the LLC 
do not occur in heads without two power laws; see \cref{sec:additional corr and llc}). 
After more training, the head transitions to a true collision detection head (shown in box C) and the 
attention-distance function moves into a new power law regime with a different exponent. 
The LLC also changes and plateaus at a higher level than before. Compared to the LLC, the change in the correlation function's 
dynamics has a sharp transition point between the two power law regimes.

\section{Discussion} \label{sec:discussion}

We have shown that collision detection heads --- attention heads which detect collisions 
between particles --- develop in a transformer-like model trained to simulate a particle system. 
This emergence coincides with the attention-distance correlation entering a power 
law regime and a change in the degeneracy of the head's parameters as measured by 
the parameter-restricted local learning coefficient (LLC). 
We observe at least two variants of collision detection heads, and in one case a head 
transitioning between these two variants. The transition between these two variants can 
be identified by a corresponding change in the attention-distance correlation 
power law exponent and by changes in the LLC. The point of transition is especially clear 
when observing the power law exponent (see \cref{fig:heads governed by degenerate potential}).
The observed attention-distance correlation power laws and LLC results suggest that the development of collisions detection heads is governed by a degenerate 
``effective potential'', that this degeneracy results in 
power law development of collision detection heads, and that transitions between variants of collision 
detection heads correspond to moving between different effective potentials. 

\subsection{Implied dynamics of collision detection heads} \label{sec:discussion implied dynamics}

Power laws are characteristic of a dynamical system which is near a degenerate critical point.
The gradient flow of an analytic function can converge at either an exponential or 
power law rate. Which of these two behaviours it exhibits is characterised by the 
\emph{\L{}ojasiewicz exponent} $\theta \in [\tfrac{1}{2}, 1)$ of a nearby critical point; 
if $\theta = \tfrac{1}{2}$ then the system 
converges at an exponential rate, and if $\theta > \tfrac{1}{2}$ the system converges as a power law with 
exponent determined by the value of $\theta$. The power law will also show up in 
measurements of the dynamical system (see \cref{corr: holder continuous measurement also evolves as power law}). The \L{}ojasiewicz exponent also characterises 
a certain kind of degeneracy in critical points, with $\theta = \tfrac{1}{2}$ if the function 
is sufficiently non-degenerate \citep{feehanResolutionSingularitiesGeometric2019}. We discuss 
the \L{}ojasiewicz exponent in more detail in \cref{sec: power laws in dynamical systems}, 
along with its relationship to the LLC.

We hypothesise that the observed power law dynamics in the attention-distance correlation function 
and the corresponding changes in the LLC are explained by a stochastic version of the 
results discussed above. When a collision detection head forms its development 
is governed by degenerate critical points in the loss landscape, which we refer to as 
an ``effective potential''. The transition between head variants shown in 
\cref{fig:heads governed by degenerate potential} (and corresponding 
changes in the LLC and correlation power law) can be explained as the head moving 
between two different effective potentials with different levels of degeneracy. 
This hypothesis is consistent with the results of \citet{hoogland_ICL1_2024} and \citet{chen_dynamical_2023}, 
which associate the development of certain computational structure with degeneracy in the loss 
landscape; here we emphasise the effect of this degeneracy on the \emph{dynamics} of the 
computational structure.

\subsection{Connection to second-order phase transitions}

The emergence of collision detection heads and the associated power laws in many ways 
resembles a physical system at a second-order phase 
transition. Power law scaling in various 
quantities, and in-particular in correlation functions similar to our attention-distance 
correlation function, is characteristic of a second-order phase transition. 
These power laws also arise as a result of degenerate critical points, 
typically in the \emph{free energy} function $f(\omega)$ of a system parameter 
$\omega$, as we explain in \cref{sec:phase-transitions-in-physics}.

The statistical mechanical interpretation of these results is particularly interesting 
in the context of training a neural network to simulate physics.
Correlation functions in statistical mechanics usually 
measure the correlation between a quantity measured at two different 
locations within a physical system. In contrast we measure 
the correlation between two different quantities, one a measurement of the 
network and the other a measurement of the target physical system. This suggests viewing training 
physics simulators as coupled systems, where the neural network is coupled to the 
target physical system via the training process. While for the model discussed in this 
paper this is a simple one-way coupling of the neural network to the target system, more
complex couplings are conceivable, such as when training physics models on sequence data 
or in physics reinforcement learning \citep[see][]{Holl2020Learning}. 

\subsection{Implications and future work}

These results suggest a description of how computational structure emerges in neural networks 
more generally; computational structure is associated to degenerate geometry in the loss 
landscape and, as a result, such structures develop with power law dynamics. Degeneracy 
in the loss landscape is a feature common to all non-trivial neural networks 
\citep[see][]{wei2022singulargood} and it is reasonable to hypothesise that it plays a similar 
role across different model architectures and learning tasks. 
This description of how computational structure emerges is consistent with the stagewise 
development model proposed in \citet{hoogland_ICL1_2024} and \citet{wangDifferentiationSpecializationAttention2024} in the sense that power law dynamics 
emerge suddenly. However, once a computational component has formed it converges 
as a power law. 

This has implications for the convergence time of computational structures in neural networks. 
It suggests computational structure develops slowly in the following sense. Consider 
a quantity $\xi \sim t^{-\alpha}$ converging to zero as a power law in the training step 
$t$ (for example, $\xi$ is the error in a performance metric of a computational component), 
and for simplicity take $\alpha=1$. 
Halving $\xi$ requires training for twice as long. 
In contrast if $\xi \sim 2^{-\alpha t}$ is exponential in $t$ (i.e. if the critical point 
is non-degenerate; see \cref{sec: power laws in dynamical systems}) halving $\xi$ requires just $\tfrac{1}{\alpha}$ additional training steps. 

A consequence of this is that structures 
which form earlier in training will be much better formed compared to those which form later, the implication 
being that promoting the formation of favourable structures and inhibiting the creation of unfavourable 
ones is particularly important early in training. This may contribute to a partial mechanistic explanation of
the results of \citet{achilleCriticalLearningPeriods2019,frankleEarlyPhaseNeural2020,chimotoCriticalLearningPeriods2024} 
which demonstrate that early periods of training can be critical in determining the final performance 
of the network.
We emphasise that this is only a partial explanation because we do not consider how the early development 
of certain structures affects the development of other structures later in training. 

Slow convergence of computational structure also suggests that practices like 
fine-tuning are unlikely to significantly change the computational structures that are present in a
model. This is because the training time of fine-tuning is a small proportion of the total training 
time of the final model. Various works have shown that undesirable 
behaviour can persist through fine-tuning, or that the changes made by 
fine-tuning can easily be undone 
\citep{hubingerSleeperAgentsTraining2024,zhanRemovingRLHFProtections2024,
ji2024language,qi2024finetuning}.
Note, however, that the loss landscape is 
different during the fine-tuning phase of the model and so this topic requires further study. 

Our results also have implications for those hoping to detect the emergence of computational 
structures via measuring degeneracy in the loss landscape. Our results 
show that 
the power law dynamics of a computational component become clear well before the local learning 
coefficient plateaus (\cref{fig:heads governed by degenerate potential}). The point transition between the two collision detection 
phases is also sharply resolved by the correlation dynamics, as compared to the LLC curve.
While the attention-distance correlation function is 
specific to this model and training task, it suggests that the measuring the dynamics of
network components may be a way of detecting emerging computational structures earlier 
than the LLC. 

\subsection{Limitations}

This work studies the emergence of computational structure for a single model; in particular the attention-distance function is specific to this model. 
While more work is needed to investigate this phenomena in other settings, all non-trivial neural networks have degenerate loss landscapes. The
theoretical results 
relating degeneracy to dynamics surveyed in \cref{sec: power laws and degeneracy} provide good 
reason to hypothesise that the power law convergence 
results we observe here should manifest in some form in other settings. 

This work only studies the emergence of collision detection heads in isolation and we do not consider how 
these interact with the rest of the model and the training process. For example we do not consider how the formation of a collision detection 
head affects the formation of other computational structures in the model later in training. 

\subsubsection*{Author Contributions}
RH, GWD, JHM, RS, JZ formulated the research question and interpreted the results. RH implemented the methodology, conducted the experiments, analysed the results, and wrote the initial draft of the manuscript. GWD, JHM, RS and JZ provided guidance and advice, and offered critical feedback throughout the study. All authors contributed to the editing and refinement of the manuscript.

\subsubsection*{Acknowledgements}
RH was supported by the AI for Missions Scholarship from CSIRO.

\bibliography{main}
\bibliographystyle{bibstyle}

\newpage
\appendix

\section{Power laws and degeneracy} \label{sec: power laws and degeneracy}

In this section we explain how power laws arise from degenerate critical points. This 
is intended to be expository; none of the results discussed in this section are new 
to this paper. We give proofs when they are not easily found in the literature.

In \cref{sec: power laws in dynamical systems} we define the \L{}ojasiewicz exponent 
of a critical point and explain how its value characterises the rate of convergence of a 
corresponding gradient flow. In \cref{sec:lojasiewicz and degeneracy} we explain the 
relationship between the \L{}ojasiewicz exponent and degeneracy. In particular we discuss 
its relationship with the real log canonical threshold, as this is an important quantity 
in Bayesian learning and closely related to the local learning coefficient. In 
\cref{sec:phase-transitions-in-physics} we discuss power laws in statistical 
mechanics. Using the Ising model as an example, we explain how 
these also arise from degenerate critical points.

\subsection{Power laws in dynamical systems with degenerate potentials} \label{sec: power laws in dynamical systems}

Whether or not a dynamical system exhibits power law behaviour is determined by the 
value of the \emph{\L{}ojasiewicz exponent} of a nearby critical point. For an 
analytic function, the \L{}ojasiewicz exponent $\theta$ of a local minimum 
is a value in $[\tfrac{1}{2}, 1)$; if $\theta = \tfrac{1}{2}$ then a gradient 
flow $\ve x(t)$ starting sufficiently close to that local minimum will converge at an 
exponential rate, and if $\theta > \tfrac{1}{2}$ then $\ve x(t)$ will converge as 
a power law. 

Let $U \subseteq \R^d$ be an open set and $f : U \to \R$ be an analytic function. Consider a 
critical point $\ve x^* \in U$ of $f$. To simplify the discussion suppose that $f(\ve x) \geq f(\ve x^*)$ 
for all $\ve x\in U$, and 
if $\grad f (\ve x) = 0$ then $f(\ve x) = f(\ve x^*)$. Let $Z = \{\ve x \in U \mid f(\ve x) = f(\ve x^*)\}$
be the set of minima of $f$ in $U$. In order to define the \L{}ojasiewicz exponent 
we need to state the following theorem, which is due to
\citet{lojasiewicz1959inequality1,lojasiewicz1959inequality2,lojasiewicz1959inequality3,lojasiewicz1959inequality4}. 

\begin{theorem}[\L{}ojasiewicz inequalities] \label{thrm:lojasiewicz_inequalities}
    There exist constants $\theta \in [\tfrac{1}{2}, 1)$ and $C, C' > 0$ and an open 
    neighbourhood $V$ of $\ve x^*$ such that:
    \begin{enumerate}
        \item $\| \grad f(\ve x) \| \geq C |f(\ve x) - f(\ve x^*) | ^{\theta}$ for all $\ve x \in V$. \label{eqn:lojasiewicz_gradient_inequality}
        \item $|f(\ve x) - f(\ve x^*)| \geq C' d(\ve x, Z) ^{\frac{1}{1-\theta}}$ for all $\ve x \in V$. \label{eqn:lojasiewicz_distance_inequality}
    \end{enumerate}
    where $d(\ve x, Z) = \inf \{\|\ve x - \ve y\| \mid \ve y \in Z\}$ is the distance between $\ve x$ and $Z$. 
\end{theorem}

\begin{definition}[\L{}ojasiewicz exponent]
    Let $\theta_0$ be the smallest value of $\theta$ 
    satisfying \eqref{eqn:lojasiewicz_gradient_inequality} 
    in \cref{thrm:lojasiewicz_inequalities}. That is, $\theta_0$ is the 
    infimum of all the $\theta \in [\tfrac{1}{2}, 1)$ such that there exists 
    a neighbourhood $V$ of $\ve x^*$ and $C > 0$ satisfying 
    \[
        \| \grad f(\ve x) \| \geq C |f(\ve x) - f(\ve x^*) | ^{\theta} \text{ for all } \ve x \in V.
    \]
    We call $\theta_0$ the \emph{\L{}ojasiewicz exponent} of $\ve x^*$.     
\end{definition}

The \L{}ojasiewicz exponent characterises the rate at which a gradient flow of $f$ --- that is, a solution to $\ve x'(t) = -\grad f(\ve x(t))$  --- converges 
to the critical point $\ve x^*$. When the \L{}ojasiewicz exponent $\theta _0 = \tfrac{1}{2}$
the gradient flows converge exponentially quickly to $Z$, and when $\theta _0 > \tfrac{1}{2}$
they converge as a power law.

\begin{corollary}\label{corr:gradient flow convergence rate}
    Let $\ve x(t)$ be a gradient flow of $f$ starting sufficiently close to $\ve x^*$ and 
    $\theta \in [\tfrac{1}{2}, 1)$ be any value such that inequalities 
    \eqref{eqn:lojasiewicz_gradient_inequality}
    and \eqref{eqn:lojasiewicz_distance_inequality} in \cref{thrm:lojasiewicz_inequalities} 
    hold. Then:
    \begin{enumerate}
        \item If $\theta = \frac{1}{2}$ then $d(\ve x(t), Z) \leq Ae^{-at}$ for all $t$, for some constants $A, a > 0$. 
        \item If $\theta > \frac{1}{2}$ then $d(\ve x(t), Z) \leq A(t + B) ^{-\alpha}$ for all $t$, 
        where $\alpha = \frac{1-\theta}{2\theta -1 } > 0$ and $A, B$ are constants, $A > 0$. 
    \end{enumerate}
    \begin{proof}
        Without loss of generality suppose that $\ve x^* = \ve 0$ and $f(\ve x^*) = 0$. Define 
        $F(t) = f(\ve x(t))$, and note that 
        \[
        F'(t) = \grad f(\ve x(t)) \cdot \ve x'(t) = -\|\grad f(\ve x(t)) \| ^2 ~.
        \]
        When $\theta = \tfrac{1}{2}$ we have 
        \[
            -F'(t) \geq C F(t) ~.
        \]
        for some $C> 0$ by \cref{thrm:lojasiewicz_inequalities}.
        Rearranging (recall that we have assumed $f(\ve x) \geq f(\ve x^*) = 0$) and 
        integrating both sides gives 
        \[
            -\int _0 ^t \frac{F'(s)}{F(s)} ds \geq Ct 
        \]
        Computing the integral and rearranging again gives 
        \[
            F(t) \leq Ae^{-a't}
        \]
        for some $A, a' > 0$. Applying inequality \ref{eqn:lojasiewicz_distance_inequality} from 
        \cref{thrm:lojasiewicz_inequalities} gives the desired result. 

        For $\theta > \tfrac{1}{2}$ we instead consider 
        \begin{align*}
            F(t) ^{1-2\theta} &= \int _0 ^t \frac{d}{ds}\left(F(s) ^{1-2\theta} \right) ds + F(0) ^{1-2\theta}\\
            &= \int _0 ^t (1 - 2\theta) F'(s) F(s) ^{-2\theta} ds + D\\
            &= (2\theta - 1) \int _0 ^t \| \grad f(\ve x(s)) \| ^2 f(\ve x(s))^{-2\theta} ds + D
        \end{align*}
        where $D = F(0) ^{1-2\theta}$. \cref{thrm:lojasiewicz_inequalities} gives 
        \begin{align*}
            F(t)^{1-2\theta} &\geq (2\theta - 1) \int _0 ^t Cf(\ve x(s))^{2\theta} f(\ve x(s)) ^{-2\theta} ds + D \\
            &\geq A' t + D
        \end{align*}
        where $A' = (2\theta -1)C > 0$. Rearranging and applying inequality 
        \ref{eqn:lojasiewicz_distance_inequality} from \cref{thrm:lojasiewicz_inequalities} 
        gives the desired result.
    \end{proof}
\end{corollary}

In our results in the main paper, however, we do not observe power laws in a distance function, but rather 
in some other measurement of our system. The following corollary shows that 
when $\ve x(t)$ converges as a power law, so does any sufficiently nice scalar 
measurement $Q(\ve x(t))$. 

\begin{corollary} \label{corr: holder continuous measurement also evolves as power law}
    Let $Q: U \rightarrow \R$ be $\beta$-H\"{o}lder continuous function and 
    let $\ve x(t)$ be a gradient flow of $f$ starting sufficiently close to $\ve x^*$. Let 
    $\theta \in [\tfrac{1}{2}, 1)$ be any value such that inequalities 
    \eqref{eqn:lojasiewicz_gradient_inequality}
    and \eqref{eqn:lojasiewicz_distance_inequality} in \cref{thrm:lojasiewicz_inequalities} 
    hold, and $Q^* = \lim _{t\to\infty } Q(\ve x(t))$. Then for some $T \in \R$ sufficiently large  
    \begin{enumerate}
        \item If $\theta = \frac{1}{2}$ then $|Q(\ve x(t)) - Q^*| \leq A'e^{-a\beta t}$ for all $t > T$. 
        \item If $\theta > \frac{1}{2}$ then $|Q(\ve x(t)) - Q^*| \leq A'(t + B') ^{-\alpha \beta }$ for all $t > T$, 
    \end{enumerate}
    where $a$, $\alpha$  and $B$ are the constants from \cref{corr:gradient flow convergence rate}
    and $A'> 0$. 
    \begin{proof}
        Recall that $Q$ being $\beta$-H\"older continuous means that for all $\ve x, \ve y \in U$
        \[
            | Q(\ve x) - Q(\ve y) | \leq C' \| \ve x - \ve y \| ^\beta
        \]
        for some $C' > 0$. Let $\ve z = \lim _{t\to \infty} \ve x(t)$. Since $Z$ 
        is a closed set we have $\ve z \in Z$, and also there exists a $T$ such that 
        for all $t > T$ we have $d(\ve x(t), \ve z) = d(\ve x(t), Z)$. Combining the 
        inequalities in \cref{corr:gradient flow convergence rate} and the inequality 
        from $Q$ being H\"older continuous yields the desired result. 

        Note that a slight variation on this argument shows that if $Q(\ve z) = Q^*$ for all $\ve z \in Z$
        then the stated inequalities hold for all $t$ where $\ve x(t)$ is defined. 
    \end{proof} 
\end{corollary}

\begin{example}[see\ \citeauthor{strogatzNonlinearDynamicsChaos2015} \citeyear{strogatzNonlinearDynamicsChaos2015}, Chapter 3.4] \label{example:pitchfork bifurcation}
    Consider the function $f: \R \to \R$ where $f(x) = \tfrac{r}{2}x^2 + \tfrac{1}{4}x^4$ and $r \in \R$
    is a constant. The 
    corresponding gradient flow equation is
    \begin{equation} \label{eqn: pitchfork bifurcation equation}
        x'(t) = - r x(t) - x(t)^3 ~.
    \end{equation}
    For $r \geq 0$ the function $f$ has a single critical point: a global minimum 
    at $x = 0$. When $r < 0$ the function $f$ has three critical points: a local maximum at $x = 0$
    and two local minima located symmetrically about $x = 0$. When $r \neq 0$ the 
    critical points are all non-degenerate, but when $r = 0$ the critical point 
    $x = 0$ is degenerate. For $r \neq 0$ we have $x'(t) \approx -r x(t)$ in a neighbourhood of $x = 0$. 
    Solving this approximate version of the differential equation gives 
    \[
        |x(t)| \approx A e^{-rt}
    \]
    for some $A > 0$ depending on the initial conditions. When $r = 0$ we instead 
    solve $x'(t) = -x(t) ^3$ to obtain 
    \[
        |x(t)| = \tfrac{1}{\sqrt{2}}(t + B) ^{-\tfrac{1}{2}}
    \]
    where $B$ depends on the initial conditions. We can use the exponent $\alpha = \frac{1}{2}$
    to find an \textit{ansatz} for the \L{}ojasiewicz exponent. We solve 
    $\frac{1}{2} = \frac{1 - \theta_0}{2\theta_0 - 1}$ to obtain $\theta_0 = \tfrac{3}{4}$. 
    The first \L{}ojasiewicz inequality in \cref{thrm:lojasiewicz_inequalities}
    for $x^* = 0$ becomes 
    \[
        |x^3| \geq C | \tfrac{1}{4} x^4 | ^\theta ~.
    \]
    Setting $C = 4$ and $\theta = \tfrac{3}{4}$ we obtain equality for all $x \in \R$.
    From this it is straightforward to prove that the \L{}ojasiewicz exponent of 
    $f(x) = \tfrac{1}{4} x^4$ is $\tfrac{3}{4}$.
\end{example}

\subsubsection{Relationship to degeneracy} \label{sec:lojasiewicz and degeneracy}

The \L{}ojasiewicz exponent characterises a certain kind of degeneracy in critical points. 
When the critical point is sufficiently non-degenerate the \L{}ojasiewicz
exponent is equal to $\tfrac{1}{2}$. Specifically, if the function $f$ is \emph{Morse-Bott} 
then the \L{}ojasiewicz exponent of any critical point of $f$ is equal to $\tfrac{1}{2}$ 
\citep[Theorem 2.1]{feehanResolutionSingularitiesGeometric2019}. A function is Morse-Bott 
if its set of critical points form a manifold and the Hessian of $f$ is non-degenerate in 
directions normal to this critical manifold. Put another way, 
the set of critical points of an analytic function $f$ is 
in-general an analytic set which 
may have singularities, and the condition of being Morse-Bott means that the set of 
critical points of $f$ does not have singularities. Combining with the results discussed in 
\cref{corr:gradient flow convergence rate}, this means that power law behaviour 
is characteristic of a function $f$ with singularities in its set of critical points.
In the Bayesian learning context 
(i.e. when $f$ is the log-likelihood function of a statistical model) the condition 
of being Morse-Bott corresponds 
to the case that the statistical model is \emph{minimally singular}.

Another measurement of degeneracy relevant to us is the \emph{real log canonical threshold} (RLCT)
\citep[see][Definition 2.7]{watanabe_algebraic_2009}. The RLCT is relevant because it is closely 
related to the local learning coefficient (their relationship is discussed in \citet[Appendix A]{lauLocalLearningCoefficient2024}
and also in \cref{sec: llc}). The RLCT characterises non-degeneracy in the same way as the \L{}ojasiewicz exponent: 
if $f$ is Morse-Bott then the RLCT of its set of critical points is maximal. For a function $f$ 
which is not Morse-Bott the relationship between the RLCT and the \L{}ojasiewicz exponent is complex 
and, to the best of the authors' knowledge, not fully understood. 

An interesting result related to the \L{}ojasiewicz exponent is that of 
\citet[Theorem 3]{feehanResolutionSingularitiesGeometric2019} which proves 
the \L{}ojasiewicz inequalities for an analytic function in 
\emph{normal crossing form} \citep[see][Definition 1.1]{feehanResolutionSingularitiesGeometric2019}, 
and hence for all analytic functions via Hironaka's resolution of singularities 
theorem \citep{hironakaResolutionSingularitiesAlgebraic1964}; a statement of this 
theorem is also given in
\citet[Theorem 2.3]{watanabe_algebraic_2009}). A consequence of this result 
is an upper bound 
for the \L{}ojasiewicz exponent of $f$ in terms of properties of a resolution of 
singularities of $f$. Normal crossing functions and resolution 
of singularities are key components of the proof in \citet{watanabe_algebraic_2009} 
that the RLCT of the log-likelihood 
function is an important quantity in Bayesian learning.

\subsection{Power laws in statistical physics} \label{sec:phase-transitions-in-physics}

Power laws are characteristic of statistical mechanical systems at second order phase 
transitions. In this section we explain, using the example of the Ising model on 
a fully connected graph, that these power laws arise for the same reason as power 
laws in dynamical systems: degenerate critical points in a certain function. For
the necessary background on statistical mechanics and phase transitions we refer 
readers to \citet{yeomans_statistical_1992,hartmann_phase_2005}. 

\subsubsection{Example: the Ising model on a fully connected graph} \label{sec:ising model example}

We consider the Ising model on a fully connected graph with $N$ nodes.
This example is adapted from \citet[Chapter 5.3]{hartmann_phase_2005}, and 
is considered primarily because it is one of the few cases where 
calculations can be done explicitly. 
A state of this system is a vector $\ve \sigma = (\sigma_1, \ldots, \sigma _N)$ where 
$\sigma _i \in \{-1, +1\}$ is the value at node $i$. The system is governed by 
the Hamiltonian 
\[
    H(\ve \sigma ) = -\frac{1}{N}\sum_{i<j} \sigma _i \sigma _j - h \sum _{i=1} ^N \sigma _i
\]
where $h \in \R$ is a hyperparameter of the system. States are distributed according 
to the Boltzmann distribution
\[
    p(\ve \sigma ) = \frac{1}{Z} \exp(-\beta H(\ve \sigma)) 
    \qquad \text{where} \qquad Z = \sum _{\ve \sigma '} \exp(-\beta H(\ve \sigma '))
\]
where $\beta > 0$ is another hyperparameter. 

We consider the behaviour of this system in the limit as $N \to \infty$. 
Physically speaking, the Ising model 
is a simplified model of magnetism in a solid. The nodes represent a small volume 
of the solid, which are magnetised in one of two polarities (``spin up'' $\sigma _i = +1$, 
or ``spin down'' $\sigma _i = -1$). 
The magnetisation of a particular state $\ve \sigma$ is given by the average of these values 
\[
    m(\ve \sigma) = \frac{1}{N} \sum _{i=1} ^N \sigma _i ~.
\]
We can rewrite the Hamiltonian as a function of the magnetisation $m \coloneqq m(\ve \sigma)$
like so
\[
    H(m) = -\frac{N}{2} m^2 - hN m + \frac{1}{2} ~.
\]

The hyperparameter $h$ represents an external magnetic field, and $1/\beta$ is 
the thermal temperature of the solid. This system has phase transitions in $(h, \beta)$
where the expected magnetisation $\E [m(\ve \sigma)]$ changes suddenly as the $h$ and 
$\beta$ are varied. For fixed $\beta > 1$ the system has a phase transition at $h = 0$ 
where the expected magnetisation changes sign to align with $h$. Physically this 
corresponds to the solid magnetising with the same polarity as the external magnetic 
field. This is a first-order phase transition. More interestingly, when $h=0$ the 
system has a phase transition at $\beta = 1$, where $\E[ |m(\ve \sigma)|]$ changes 
from a non-zero value for $\beta > 1$ to  $|\E [|m(\ve \sigma)|] = 0$ for $\beta < 1$. 
Unlike in the previous case, where $\E [m(\ve \sigma)]$ was discontinuous as a function 
of $f$, only the derivatives of $\E [m(\ve \sigma)]$ with respect to $h$ and $\beta$ are
discontinuous \citep[see][Figure 2.1]{yeomans_statistical_1992}. 
Physically this corresponds to the Curie point of a material, where a magnet suddenly 
becomes demagnetised once it is heated past a certain temperature. The point 
$(h, \beta) = (0, 1)$ is a second order phase transition of the system. For more 
details on the phase structure of the Ising model see \citet[Chapter 2]{yeomans_statistical_1992}. 

We can understand these phase transitions by studying the partition function $Z$ 
or equivalently the free energy $F = \tfrac{-1}{\beta} \log(Z)$. This is because 
the expected magnetisation is a derivative of the free energy 
\[
    \E[m(\ve \sigma)] = \frac{-1}{N} \frac{\partial F}{\partial h} ~.
\]
We now consider the partition function $Z$. For large $N$ one can show
\[
    Z \approx \int _{-1} ^1 \exp(-\beta N f(m)) dm
\]
where 
\[
    \beta f(m) = \tfrac{1}{2}\beta m^2 + hm - \frac{1 + m}{2} \log \left(\frac{1 + m}{2}\right) - \frac{1 - m}{2} \log \left(\frac{1-m}{2}\right)
\]
For details of this derivation see \citet[Chapter 5.3]{hartmann_phase_2005}. 
Incidentally, $f(m)$ is (approximately, for large $N$) the free energy of a state with 
magnetisation $m$
\[
    f(m) \approx \tfrac{1}{N} H(m) - \tfrac{1}{\beta} \log (\mathcal{N}(m))
\]
where $\mathcal{N}(m)$ is the number of states with magnetisation $m$, and so 
$\log (\mathcal{N}(m))$ is the entropy.\footnote{Our definition of Ising model
implicitly sets units so that the Boltzmann constant $k_B = 1$.}

For large $N$ we can approximate $Z$ using a saddle point approximation 
\[
    Z \approx \exp(-\beta N f(m^*) + \mathcal{O}(\log N))
\]
where $m^*$ is an isolated global minimum\footnote{The validity of this 
saddle point approximation depends on $f(m)$ having isolated global minima. We 
have not shown this is the case, but it does.} of $f(m)$. The derivative of $f(m)$ is 
\[
    f'(m) = \tfrac{1}{\beta}\arctanh(m) - m - h
\]
where $\arctanh$ is the inverse to $\tanh$. Solving $f'(m) = 0$ lets us find 
$m^*$. This is best done graphically using the methods described in 
\citet[Chapter 3]{strogatzNonlinearDynamicsChaos2015}. We focus on the two 
interesting cases of $h = 0$ and $\beta < 1$ and $h = 0$ and $\beta > 1$. In 
the former we have a single solution $m = 0$ which is a local minimum. In the 
latter we have three solutions: two local minima $m = \pm M(\beta)$ for some $M(\beta) \in \R$, 
and a local maximum at $m = 0$. Note that the $m = 0$ changes from a local minima 
as $\beta$ is varied past $\beta = 1$. 

We now show that in a neighbourhood of $m = 0$ and $\beta = 1$ the system $f'(m) = 0$
resembles that of \cref{example:pitchfork bifurcation}. Rearranging we find 
\[
    0 = \tanh(\beta m) - m~.
\]
In a neighbourhood of $m = 0$ we can use the Taylor expansion of $\tanh(\beta m)$ 
to obtain
\[
    0 \approx \beta m - \beta ^3 m^3 ~.
\]
Setting $T = \tfrac{1}{\beta}$ (the temperature), the critical value is $T_c = 1$. 
We define $r = \frac{T - T_c}{T_c} = T - 1$, which gives $\beta = \frac{1}{r + 1}$. 
In a neighbourhood of $r = 0$ we have $\beta \approx 1 - r$. Substituting this approximation 
and dropping the higher order terms we find 
\[
    0 \approx -rm - m^3 ~.
\]
Note the similarities with \eqref{eqn: pitchfork bifurcation equation} in 
\cref{example:pitchfork bifurcation}. 
Writing $m$ in terms of $r$ we find the power law relationship 
\[
    |m| \sim |r|^{\frac{1}{2}}
\]
when $m$ is in a neighbourhood of $m=0$ and $r$ is in a neighbourhood of $r = 0$. 
Other power laws exist in a neighbourhood of this critical point and are discussed 
in \citet[Chapter 2.6]{yeomans_statistical_1992}.

\subsubsection{Limitations in the analogy to deep learning}

While the above example is instructive, there is an important difference with 
the nature of the critical points encountered in the Ising model 
and similar statistical mechanical systems and those encountered in deep learning. 
In \cref{sec:ising model example} we relied on a saddle point approximation 
of $Z$. For this to be valid this required that the critical points of $f(m)$ were 
isolated. For neural networks this is not the case and a more careful treatment 
is required, such as the approach developed in \citet[Chapter 4]{watanabe_algebraic_2009}. 
We emphasise however that the results discussed in \cref{sec: power laws in dynamical systems}
do not depend on critical points being isolated and so we should still have an 
expectation of power law behaviour in a neighbourhood of sets of non-isolated 
critical points.

\section{Local learning coefficient estimation} \label{sec: llc}

In this section we give an overview of the theoretical and practical aspects of 
estimating the local learning coefficient. In \cref{sec:llc theoretical formulation}
we explain the formulation of the 
local learning coefficient via a specially constructed Bayesian posterior 
distribution. Nothing discussed in \cref{sec:llc theoretical formulation} 
is original to this paper: the connection between Bayesian learning and degenerate 
geometry is a deep result established in \cite{watanabe_algebraic_2009} with 
estimation methods first proposed in \cite{watanabe2013widely}. The specialisation 
of these results to local (as opposed to global) geometry and to deep learning 
is due to \cite{lauLocalLearningCoefficient2024,chen_dynamical_2023}. 

In \cref{sec:llc practical details} we discuss the details of the local learning 
coefficient estimation results in this paper, including the estimation hyperparameters 
and details of the hyperparameter selection process and additional 
results. 

\subsection{Theoretical formulation} \label{sec:llc theoretical formulation}

In this section consider a neural network $f : \R^N \times W \rightarrow \R^M$
which takes inputs in $\R^N$, outputs in $\R^M$ and has parameters $W \subseteq \R^d$. 
Let $q(\ve x, \ve y)$ be a distribution on $\R^N \times \R^M$ and 
$D_n = \{(X_1, Y_1), \ldots, (X_n, Y_n) \}$ be a dataset of input-output pairs 
drawn independently from $q(\ve x, \ve y)$. 
We consider the task of training the neural network using the mean-squared 
error loss function:
\[
    \ell(\ve w, \ve x, \ve y) = \|\ve y - f(\ve x, \ve w) \|^2
\]
The goal is find parameters which minimise the expected loss 
\begin{equation}
    L(\ve w) = \E_{(X, Y)\sim q(\ve x, \ve y)} \ell(\ve w, X, Y) ~. \label{llc-apdx:eqn:expected loss regression}
\end{equation}

Formally speaking, local learning coefficient estimation amounts to computing the 
expectation with respect to a Bayesian posterior. For a given parameter $\ve w_0 \in W$
(for example a training checkpoint of our neural network), we construct a Bayesian 
learning problem which is distinct from, but related to, the problem of training 
our neural network on the given task. Our Bayesian learning problem is constructed 
in such a way that it tells us something about the local geometry of $L(\ve w)$ near 
the parameter $\ve w_0$; if $\ve w_0$ happens to be a local minimum it specifically tells 
about degeneracy of $\ve w_0$ as a critical point of $L(\ve w)$. That Bayesian learning can tell 
us anything about the degeneracy of critical points is a deep theoretical result 
established in \citet{watanabe_algebraic_2009,watanabe2013widely}.

For a given parameter $\ve w_0 \in W$ a Bayesian learning problem is constructed as 
follows.
We choose a (small) compact neighbourhood $W' \subseteq W$ of $\ve w_0$ and a prior 
$\varphi(\ve w)$ with support $W'$.\footnote{In practice however $\varphi(\ve w)$ is 
chosen to be a sharp Gaussian prior which does not have compact support.}
Assume that the input distribution $q(\ve x)$ is known; the learning problem is 
to recover the true 
conditional distribution $q(\ve y\mid \ve x)$ derived from $q(\ve x, \ve y)$. 
For each $\ve w \in W$ we define a conditional distribution 
\[
    p(\ve y\mid \ve x, \ve w) \propto \exp(-\tfrac{1}{2} \| \ve y - f(\ve x; \ve w) \| ^2) ~.
\]
This definition is chosen so that the negative log-likelihood of $p(\ve y\mid \ve x, \ve w)$ 
as a function of $\ve w$ coincides with the expected loss $L(\ve w)$ in 
\eqref{llc-apdx:eqn:expected loss regression}. Define 
\[
    W_0 ' = \{\ve w \in W' \mid L(\ve w) = L_0 \} \qquad \text{where}~ L_0 = \min _{\ve w \in W'} L(\ve w) ~.
\]
Subject to some technical assumptions \citep[see][Appendix A]{lauLocalLearningCoefficient2024}, including that $L(\ve w)$ is an analytic function 
of $W$, the \emph{real log canonical threshold} (RLCT) of the 
analytic variety $W_0'$ --- which we denote $\lambda (W_0')$ --- can be estimated using \citet[Theorem 4]{watanabe2013widely} as
\begin{equation} \label{llc-apdx:eqn:local learning coefficient}
    \hat \lambda(W_0') \coloneqq n\beta \left(\E^{(\beta)}_{\ve w} [L_n(\ve w)] - L_n(\ve w_0)\right)
\end{equation}
where $L_n(\ve w) = \tfrac{1}{n}\sum _{i=1} ^n \ell(\ve w, X_i, Y_i)$, $\beta > 0$ and 
$\E_{\ve w} ^{(\beta)}$ denotes expectation with respect to the tempered Bayesian posterior 
distribution at inverse temperature $\beta$:
\begin{equation} \label{llc-apdx:eqn:tempered posterior}
    p^{(\beta)}(\ve w \mid D_n) \propto \varphi(\ve w) \exp(-\beta n L_n(\ve w)) ~.
\end{equation}
If $\ve w_0$ is a global minimum of $L(\ve w)$ on $W'$ and a further technical assumption 
is met (roughly speaking that $\ve w _0$ is the most degenerate critical point of $L(\ve w)$
in $W'$ in the sense measured by the RLCT) then $\lambda(W_0')$ coincides with 
the local RLCT of $\ve w_0$ as a critical point of $L(\ve w)$ 
\citep[Appendix A]{lauLocalLearningCoefficient2024}. 

Estimating the local learning coefficient in \eqref{llc-apdx:eqn:local learning coefficient}
amounts to sampling from the tempered posterior distribution in \eqref{llc-apdx:eqn:tempered posterior}. 
This is done using Markov-chain Monte Carlo methods, specifically the Stochastic Gradient Langevin Dynamics (SGLD) algorithm 
\citep{welling2011bayesian} which uses minibatches of the dataset $D_n$ to compute 
each update rather than the entire dataset. For practical reasons a sharp Gaussian 
prior centred at $\ve w_0$ 
\[
    \varphi(\ve w) \propto \exp(-\tfrac{\gamma}{2} \| \ve w - \ve w_0\|^2) \qquad \text{where}~\gamma > 0
\]
is used rather than a prior with compact support. Here $\gamma$ is a hyperparameter of the local learning 
coefficient estimation procedure. Starting at $\ve w_0$ SGLD generates a sequence 
of parameters $\ve w_0, \ve w_1, \ve w_2, \ldots, \ve w_T$ where $\ve w_{t+1} = \ve w_t + \Delta \ve w_t$
and 
\begin{equation} \label{llc-apdx:eqn:sgld update rule}
    \Delta \ve w_t = -\frac{\epsilon}{2} (\gamma (\ve w - \ve w_0) + n \beta \nabla L_m (\ve w_t)) + \eta _t(\epsilon)
\end{equation}
where $\epsilon > 0$ is the step size, $m$ is the batch size and $\eta _t(\epsilon) \sim \text{Normal}(\ve 0, \epsilon I)$ 
are iid for $t = 0, \ldots, T-1$. After a burn-in period of $B$ steps, the 
expectation $\E^{(\beta)}_{\ve w} [L_n(\ve w)]$ is approximated as 
\[
    \E^{(\beta)}_{\ve w} [L_n(\ve w )] \approx \frac{1}{T - B} \sum ^T _{t = B} L_m(\ve w_t) ~.
\]
 
\subsection{Empirical details} \label{sec:llc practical details}

In this paper we consider \emph{weight restricted} local learning coefficient 
estimation, in which all parameters in a model are considered fixed except 
for those within a certain architectural component of the model (in our case 
an attention head). This lets us focus on the geometry of specific components 
of our model, rather than how the geometry changes as a function of all trainable 
parameters. In the notation of \cref{sec:llc theoretical formulation} this 
amounts to choosing our parameter space $W$ to consist only of the parameters 
of a given attention head, with the remainder of the parameters of the model 
fixed for the purpose of estimating the local learning coefficient. 

As described in \cref{sec:llc theoretical formulation}, we estimate 
the local learning coefficient using SGLD using components 
of the \texttt{devinterp} Python library \citep{devinterp2024}. Referring to the 
update rule of SGLD given in \eqref{llc-apdx:eqn:sgld update rule}, the hyperparameters
of the estimation procedure are the step size $\epsilon > 0$, the localisation 
strength $\gamma > 0$, the batch size $m$, the number of steps $T$, the number 
of burn-in steps $B$, and the gradient factor $\tilde{\beta} \coloneqq n\beta$.
Since our training data can be generated 
as-needed from a ground truth physics simulation we generate a sufficiently large 
dataset so that samples are never reused by SGLD. When producing the LLC results in 
\cref{fig:heads governed by degenerate potential} and 
\cref{sec:additional corr and llc} we complete $C$ independent estimation 
runs --- or chains --- at each training checkpoint. We also apply a small amount of 
smoothing to these LLC curves by taking averages over a small, fixed-width sliding 
window. The hyperparameters used to obtain our 
main local learning coefficient results are given in \cref{table:llc hyperparameters main}. 
\begin{table}[h]
    \caption{Hyperparameters for local learning coefficient results.}
    \label{table:llc hyperparameters main}
    \vskip 0.15in
    \begin{center}
    \begin{small}
    \begin{sc}

    \begin{tabular}{lc}
    \toprule
    Hyperparameter & Value \\
    \midrule
    Step size $\epsilon$ & 0.00075 \\
    Localisation $\gamma$ & 1 \\
    Gradient factor $\tilde \beta$ & 1385 \\ 
    Batch size $m$ & 256 \\
    Total steps $T$ & 10000 \\
    Burn-in steps $B$ & 9000 \\
    Number of chains $C$ (\cref{fig:heads governed by degenerate potential}) & 8 \\
    Number of chains $C$ (\cref{sec:additional corr and llc}) & 4 \\
    \bottomrule
    \end{tabular}

    \end{sc}
    \end{small}
    \end{center}
    \vskip -0.1in
\end{table}

\subsubsection{Hyperparameter selection process}

Hyperparameters were chosen by inspecting the \emph{trace plots} of different 
SGLD runs. The trace plot of a sequence of SGLD steps $\ve w_0, \ve w_1, \ldots, \ve w_T$ 
is the graph of the step $t$ versus the loss $L_m(\ve w_t)$. Healthy trace plots should show that $L_m(\ve w_t)$
converges to fluctuate around a fixed positive value as $t$ becomes large, and 
should be free from pathologies like large spikes and erratic fluctuations. Sweeps 
were conducted over $\epsilon$, $\gamma$ and $\tilde \beta$, where the batch size 
$m$ and total steps $T$ were increased as-needed based on observations from previous 
sweeps. The number of burn-in steps $B$ was chosen so that the loss 
trace had converged. For each set of hyperparameters SGLD was performed at a number of checkpoints 
over training with the goal of finding a single set of hyperparameters which worked 
well over all training checkpoints. 

Finding hyperparameters that worked for both earlier and later checkpoints was 
a particular challenge. The loss trace for earlier checkpoints tended to converge 
much slower. Normally this could be resolved by increasing the learning rate, but 
this caused divergence in later checkpoints. This was resolved by increasing the 
batch size $m$ to be four times greater than the batch size used during training 
(this allowed the later checkpoints to tolerate a higher learning rate) and increasing 
the total number of steps $T$. Even still, we could not find a single set of hyperparameters 
which worked for all collision detection heads over all training checkpoints.

\section{Model and training details} \label{sec:model and training details}

In this section we give the details of the model architecture, training process 
and training data generation. \cref{fig:model_architecture} shows a detailed 
diagram of the model architecture and \cref{table:model architecture hyperparameters}
summarises the model's hyperparameters. To specify physical boundaries we use virtual `boundary particles' with 
positions $\ve b_1, \ldots, \ve b_M \in \R^D$. These do not have a velocity and are 
not moved by collisions with the particles in the system.

We trained this model architecture 
for 64000 steps on input-output pairs of states, where the output state is 
$\Delta t = 0.005$ after the input state. Training hyperparameters are given 
in \cref{table:training hyperparameters}. 

The training data is generated as follows. 
A $D$-dimensional particle state with $N$ particles 
is a tensor $\ve x = [\ve p \ \ve v] \in \R^{N\times 2D}$ consisting of the positions $\ve p \in \R^{N \times D}$
and velocity $\ve v \in \R^{N \times D}$ of every particle in the system.
We randomly initialise a particle state $\ve x_0 = [\ve p_0 \ \ve v_0]$ in a fixed 
rectangular box and 
then simulate it for $1024$ steps using damped linear spring dynamics to resolve collisions between particles 
(\citet{luding2008introduction}; see \cref{table:data hyperparameters} 
for the simulation hyperparameters). The number of steps was chosen as after this 
point the particle system has generally dissipated all of its energy and particles 
are no longer moving. The resulting sequence of states 
$\ve x_0, \ve x _1, \ldots, \ve x_{1024}$ is then turned into input-output 
pairs $(\ve x_0, \ve x_1), (\ve x_1, \ve x_2), \ldots, (\ve x_{1023}, \ve x_{1024})$. 
This process is then repeated for a different random initial state. Samples 
from multiple sequences are shuffled together before training. We generate 
enough training data so that the model never sees the same data sample more than 
once.

\begin{table}[h]
    \caption{Hyperparameters for training process.}
    \label{table:training hyperparameters}

    \vskip 0.15in
    
    \begin{center}

    \begin{small}
    \begin{sc}
    
        \begin{tabular}{lc}
            \toprule
            Hyperparameter & Value\\
            \midrule
            Optimiser & Adam \\
            Learning rate & 0.001 \\
            Weight decay & 0 \\
            Batch size & 64 \\
            Number of training steps & 64000 \\
            Number of runs & 5 \\
            \bottomrule
        \end{tabular}

    \end{sc}
    \end{small}
    \end{center}
    
    \vskip -0.1in
\end{table}

\begin{table}[h]
    \caption{Hyperparameters for data generation.}
    \label{table:data hyperparameters}
    \vskip 0.15in
    
    \begin{center}

    \begin{small}
    \begin{sc}
    
        \begin{tabular}{lc}
            \toprule
            Hyperparameter & Value \\
            \midrule
            Spatial dimension $D$ & 2 \\
            Number of particles $N$ & 64 \\
            Particle diameter & 1 \\
            Particle mass & 1 \\
            Time step $\Delta t$ & 0.005 \\
            Collision spring constant & 1000 \\
            Collision damping coefficient & 10 \\
            Acceleration due to gravity $\|\ve g\|$ & 9.8 \\
            Box height & 14 \\
            Box width & 28 \\
            Initial particle velocity distribution & Standard normal \\
            Number of simulated steps from initial state & 1024 \\
            \bottomrule
        \end{tabular}

    \end{sc}
    \end{small}
    \end{center}
    
    \vskip -0.1in
\end{table}

\subsection{Model architecture} \label{sec:model architecture appendix}

\begin{figure}[h]
    \centering
    \footnotesize
    \def\svgwidth{0.77\linewidth}
    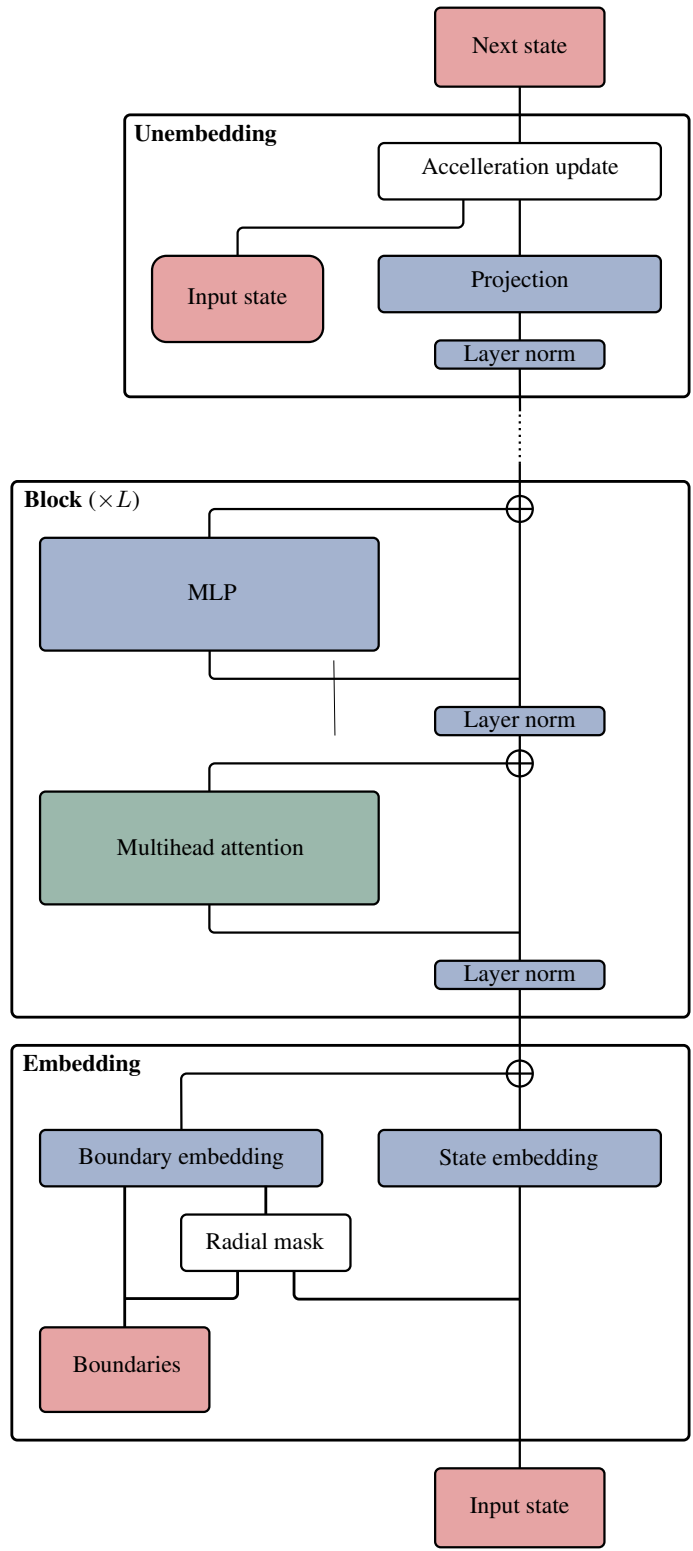
    \caption{The model architecture. Layers which operate on particle embeddings 
    in-parallel are shaded in blue. Layers which mix information between particle 
    embeddings are shaded in green. Data is shaded in red. Operations without any 
    trainable parameters are left unshaded.}
    \label{fig:model_architecture}
\end{figure}

The model architecture is summarised in \cref{fig:model_architecture} and we 
describe its components in more detail below. The values of hyperparameters 
of the architecture are given in \cref{table:model architecture hyperparameters}. 
The input data of the model consists of the states of $N$ particles, 
each of which has a position $\ve p_i \in \R^D$ and a velocity $\ve v_i \in \R^D$. We also have 
the positions of $B$ \emph{boundary particles} $\ve b_1, \ldots, \ve b_B \in \R^D$. 
These are virtual particles used to indicate the presence of physical barriers in 
the system and they are densely placed along physical boundaries in the system. 

\begin{table}[h]
    \caption{Hyperparameters for model architecture.}
    \label{table:model architecture hyperparameters}
    \vskip 0.15in
    
    \begin{center}

    \begin{small}
    \begin{sc}
    
        \begin{tabular}{lc}
            \toprule
            Hyperparameter & Value \\
            \midrule
            Number of blocks $L$ & 4 \\
            Number of heads & 8 \\
            Embedding dimension $E$ & 128 \\
            Boundary mask radius $r$ & 2 \\
            Activation function $\sigma$ & ReLU \\
            \bottomrule
        \end{tabular}

    \end{sc}
    \end{small}
    \end{center}
    
    \vskip -0.1in
\end{table}

\paragraph{Embedding} 
The state of each particle $\ve x_i = [\ve p_i\ \ve v_i] \in \R^{2D}$ is passed through 
two fully-connected layers $\ve x_i \mapsto \ve x_i' \coloneqq W_{e2} \sigma (W_{e1} \ve x_i + \ve c_{e1}) + \ve c_{e2}$
to obtain the \emph{embedded particle state} $\ve x_i' \in \R^E$. Here $W_{e1} \in \R^{E \times 2D}$, $W_{e2} \in \R^{E \times E}$, 
$\ve c_{e1}, \ve c_{e2} \in \R^{E}$ and $\sigma$ is the activation function. Separately, 
for each particle we do a graph convolution to include information about nearby boundaries. 
For each particle $i$ we compute the \emph{embedded boundary data} $\ve b_i ' \in \R^E$
\[
    \ve b_i ' = \sum _{j=1} ^B \1(\|\ve p_i - \ve b_j \| < r) \left[ W_{b2} \sigma (W_{b1} \ve b_j + \ve c_{b1}) + \ve c_{b2} \right]
\]
where $\1(\|\ve p_i - \ve b_j \| < r)$ is the \emph{radial mask}, selecting only
the boundary particles within $r$ units of the current particle, and $W_{b1} \in \R^{E \times D}$, $W_{b2} \in \R^{E \times E}$, 
$\ve c_{b1}, \ve c_{b2} \in \R^{E}$. This embedded boundary data is then used 
to update the embedded particle state:
\[
    \ve x_i ^{(0)} = \ve x_i' + \ve b_i ' ~.
\]

\paragraph{Transformer blocks} 

The embedded state of each particle is then passed through $L$ transformer blocks. 
The main components of each block are the multihead attention layer and the 
two-layer MLP. Layer normalisation is also used as indicated in \cref{fig:model_architecture}. 
In the multihead attention layer each particle attends to the other embedded particle 
states to update its own embedding. Every particle is allowed to attend to every particle
embedding other than its own. The multihead attention layer is almost the same as in 
\citet[Algorithm 5]{phuong_formal_2022} with one modification which means that the total attention 
assigned by a particle can be less than one. This is done by appending the zero vector 
to the key and value vectors before the attention scores are implemented. See the implementation 
of the \texttt{add\_zero\_attn} option of \texttt{MultiheadAttention} in PyTorch \citep{pytorch}. 
The MLP layer operates on each particle embedding and consists of two fully-connected layers separated 
by a non-linearity. 

\paragraph{Unembedding} 
After the transformer blocks and an additional layer normalisation step, the final 
embedding of each particle state $\ve x_i ^{(L)}$ is projected to a vector $\ve a_i \in \R^D$ using a 
single linear layer: $\ve a_i = W_{u} \ve x_i ^{(L)} + \ve c_u$.  
The updated position $\ve p_i'$ and velocity $\ve v_i '$ 
are obtained by 
\begin{equation} \label{eqn:model-pos-vel-update}
    \ve v ' _i = \ve v_i + \Delta t (\ve a_i + \ve g) \qquad \ve p _i ' = \ve p_i + \Delta t \ve v_i ' 
\end{equation}
where $\ve g \in \R^D$ is acceleration due to gravity. We can interpret the output $\ve 
a_i$ of the model as the net acceleration experienced by particle $i$ due to collisions 
with other particles. With this interpretation, \eqref{eqn:model-pos-vel-update} 
corresponds to 
integrating the equations of motions using the first-order 
\emph{semi-implicit Euler method}. When applied to numerical particle simulations this 
integrator has good stability properties 
\citep{samieiAssessmentPotentialsImplicit2013,tuleyOptimalNumericalTime2010}. 

\clearpage

\section{Additional results and methods} \label{sec: additional results}

\begin{figure}[h]
    \centering
    \includegraphics[width=0.7\textwidth]{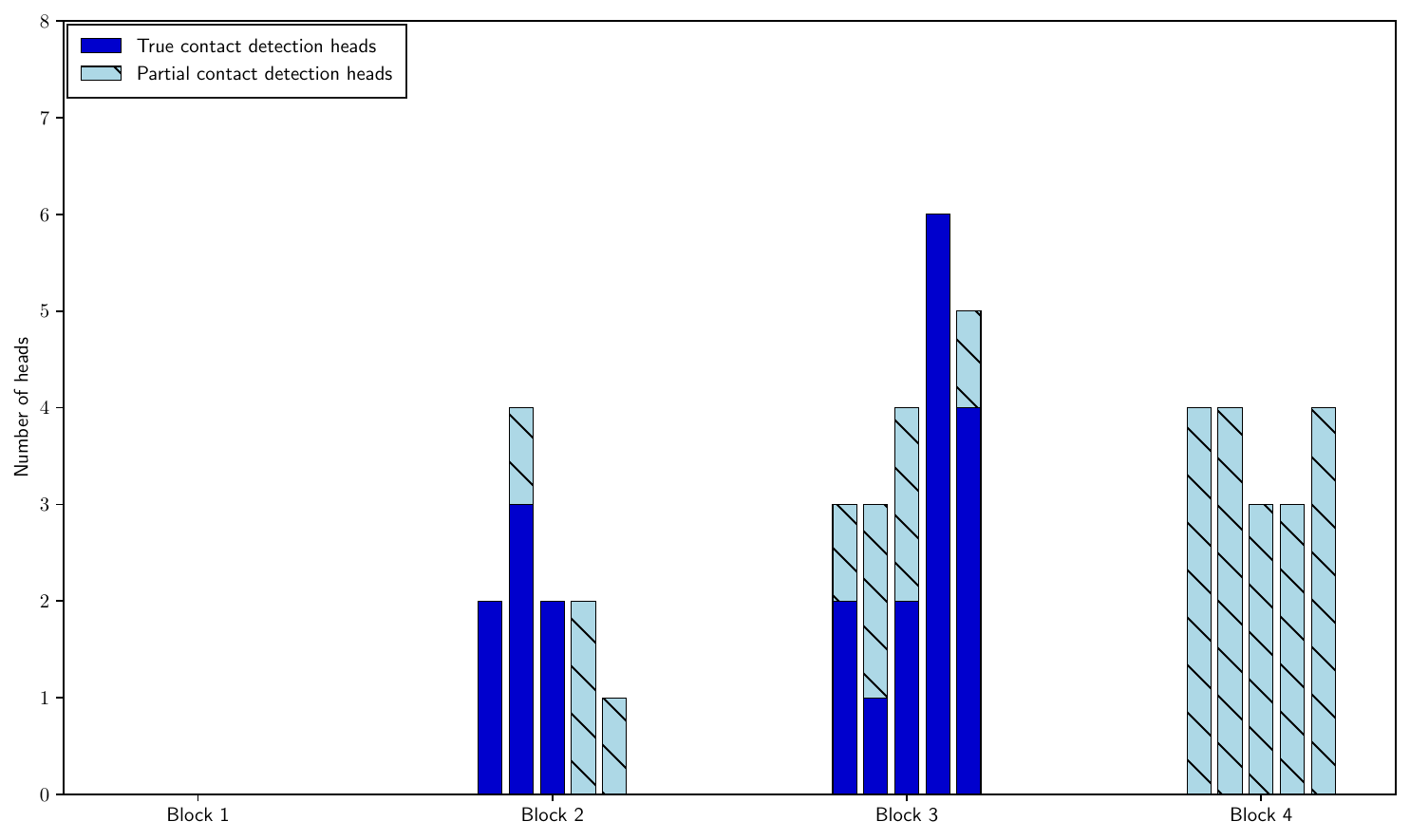}
    \caption{The number of true collision detection heads and partial collision detection heads within each 
    transformer block at the end of training. There are eight attention heads in each block. 
    Data is shown for all five training runs, each as a separate bar.}
    \label{fig:head-counts-by-block}
\end{figure}

In this section we give some additional results. In \cref{fig:head-counts-by-block} we give 
the number of partial and true collision detection heads, broken down by training run and by location 
within the network architecture. In \cref{sec:power law fit statistics appendix} we 
give the fitted power laws for all collision detection heads in the model as well as discussing 
the fitting methodology, in \cref{sec:model-performance} we discuss the performance of the model as a 
particle simulator, in \cref{sec:additional-attenrtion-plots} we give further visualisations of attention along the lines of \cref{fig:head behaviour}, and in section 
\cref{sec:additional corr and llc} we give additional attention-distance correlation and local 
learning coefficient results, with plots similar to \cref{fig:heads governed by degenerate potential}
produced for other collision detection heads. 

\subsection{Power law fit statistics} \label{sec:power law fit statistics appendix}

\begin{figure}[h]
    \begin{center}
    \includegraphics[width=\textwidth]{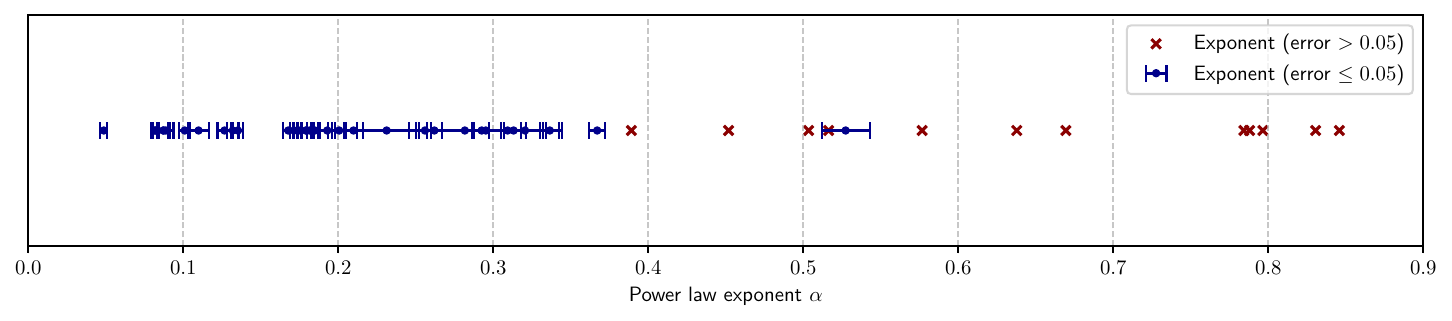}
    \end{center}
    \caption{The values of the exponent $\alpha$ in the fitted power law relationship 
    $|c| = At^{-\alpha}$ between the attention-distance correlation $c$ and the training 
    step $t$. Error bars are omitted from exponents where the estimated error is greater than $0.05$ 
    to improve readability.} \label{fig: power law exponents}
\end{figure}

We identified heads where the attention-distance correlation curve $c(t)$ has power law 
behaviour by inspecting a log-log plot of $|c(t)|$ versus training step $t$. On a log-log 
plot a power law $|c| = At^{-\alpha t}$ appears linear. For the heads where power laws were 
identified we used least-squares linear regression on the transformed plot of $\log(|c|)$ versus 
$\log(t)$ to find parameters $\alpha$ and $A$
for the curve $\log(|c|) = -\alpha \log(t) + \log(A)$. This corresponds to a power law 
$|c| = At^{-\alpha}$. We report the power law sections, corresponding fitted parameters, and 
the $R^2$ value of the linear fit in \cref{table:power law exponents}. We also attempt to 
estimate the error in $\alpha$ by computing a $95\%$ confidence interval for $\alpha$ under the 
following assumption: the true relationship between $\log(|c|)$ and $\log(t)$ is linear, and the 
data used to produce the fit has iid normal errors with zero mean. This error term is also 
reported in \cref{table:power law exponents}. We plot the power law exponents in  
\cref{fig: power law exponents}. This plot shows some 
clustering of power law exponents near certain values.

\begin{table}[h]
    \caption{The results from fitting a power law of the form $|c| = A t^{\alpha}$ 
    where $c$ is the attention-distance correlation of the given head and $t$ is 
    training time in steps. To obtain the fit, least-squares linear regression 
    was applied to find parameters $\log(|c|) = \alpha \log(t) + \log(A)$ for 
    $t$ between the given start and end values. The $R^2$ column is the $R^2$ 
    value associated to this linear regression. The error term in $\alpha$ indicates the 
    $95\%$ confidence interval for $\alpha$ under the assumption that $\log(|c|)$ and 
    $\log(t)$ have a linear relationship and that the data used to produce the 
    fit have iid normal errors with zero mean.}
    \label{table:power law exponents}
    \begin{center}
    \begin{small}
    \begin{sc}
    \renewcommand{\arraystretch}{1.2}  
    \begin{tabular}{ccccccc}
\toprule
Run & Head & Exponent $-\alpha$ & Log-constant $\log(A)$ & $R^2$ & Start step & End step \\
\midrule
0 & (1, 4) & $-0.784 \pm 0.131$ & 5.134 & 0.934 & 4864 & 5734 \\
0 & (1, 6) & $-0.788 \pm 0.128$ & 5.145 & 0.831 & 4596 & 6695 \\
0 & (2, 1) & $-0.321 \pm 0.014$ & 1.491 & 0.954 & 4352 & 12032 \\
0 & (2, 1) & $-0.082 \pm 0.002$ & -0.771 & 0.961 & 14976 & 64000 \\
0 & (2, 5) & $-0.179 \pm 0.003$ & 0.263 & 0.964 & 16000 & 64000 \\
0 & (2, 6) & $-0.256 \pm 0.004$ & 1.317 & 0.975 & 18944 & 64000 \\
0 & (3, 1) & $-0.262 \pm 0.005$ & 1.363 & 0.971 & 22616 & 64000 \\
0 & (3, 2) & $-0.087 \pm 0.004$ & -0.711 & 0.839 & 24320 & 64000 \\
0 & (3, 3) & $-0.313 \pm 0.008$ & 1.827 & 0.950 & 27392 & 64000 \\
0 & (3, 4) & $-0.179 \pm 0.006$ & 0.422 & 0.925 & 26624 & 64000 \\
1 & (1, 0) & $-0.127 \pm 0.005$ & -0.259 & 0.895 & 22784 & 64000 \\
1 & (1, 2) & $-0.831 \pm 0.252$ & 5.107 & 0.703 & 2702 & 3683 \\
1 & (1, 6) & $-0.503 \pm 0.081$ & 2.584 & 0.884 & 4804 & 6272 \\
1 & (1, 7) & $-0.638 \pm 0.076$ & 3.653 & 0.874 & 2118 & 3936 \\
1 & (2, 1) & $-0.087 \pm 0.003$ & -0.730 & 0.904 & 22121 & 64000 \\
1 & (2, 3) & $-0.231 \pm 0.019$ & 0.879 & 0.801 & 32226 & 47998 \\
1 & (2, 6) & $-0.185 \pm 0.008$ & 0.255 & 0.907 & 4992 & 21636 \\
1 & (3, 0) & $-0.527 \pm 0.016$ & 4.288 & 0.958 & 41216 & 64000 \\
1 & (3, 4) & $-0.293 \pm 0.005$ & 1.777 & 0.975 & 20096 & 64000 \\
1 & (3, 6) & $-0.210 \pm 0.006$ & 0.856 & 0.932 & 26368 & 64000 \\
1 & (3, 7) & $-0.171 \pm 0.003$ & 0.212 & 0.976 & 16227 & 64000 \\
2 & (1, 0) & $-0.797 \pm 0.184$ & 5.535 & 0.813 & 6144 & 7552 \\
2 & (1, 1) & $-0.846 \pm 0.145$ & 5.671 & 0.893 & 4397 & 5609 \\
2 & (2, 0) & $-0.182 \pm 0.006$ & 0.438 & 0.923 & 24960 & 64000 \\
2 & (2, 1) & $-0.295 \pm 0.049$ & 1.096 & 0.760 & 8540 & 12416 \\
2 & (2, 5) & $-0.336 \pm 0.006$ & 2.262 & 0.971 & 25856 & 64000 \\
2 & (2, 7) & $-0.309 \pm 0.023$ & 1.204 & 0.897 & 5504 & 11904 \\
2 & (3, 0) & $-0.193 \pm 0.005$ & 0.574 & 0.954 & 28800 & 64000 \\
2 & (3, 1) & $-0.175 \pm 0.004$ & 0.448 & 0.947 & 23168 & 64000 \\
2 & (3, 6) & $-0.168 \pm 0.003$ & 0.364 & 0.964 & 19328 & 64000 \\
3 & (1, 0) & $-0.089 \pm 0.005$ & -0.015 & 0.752 & 19584 & 64000 \\
3 & (1, 2) & $-0.049 \pm 0.002$ & -0.952 & 0.788 & 13696 & 64000 \\
3 & (2, 2) & $-0.516 \pm 0.112$ & 3.240 & 0.715 & 8320 & 11138 \\
3 & (2, 4) & $-0.389 \pm 0.118$ & 1.611 & 0.525 & 3584 & 5994 \\
3 & (3, 2) & $-0.367 \pm 0.005$ & 2.559 & 0.982 & 22121 & 64000 \\
3 & (3, 3) & $-0.135 \pm 0.003$ & 0.405 & 0.936 & 15744 & 64000 \\
3 & (3, 6) & $-0.101 \pm 0.003$ & -0.505 & 0.904 & 22616 & 64000 \\
4 & (2, 1) & $-0.452 \pm 0.078$ & 2.444 & 0.833 & 4397 & 6265 \\
4 & (2, 3) & $-0.577 \pm 0.111$ & 3.522 & 0.727 & 4911 & 7808 \\
4 & (2, 5) & $-0.669 \pm 0.105$ & 3.983 & 0.832 & 2585 & 4224 \\
4 & (2, 7) & $-0.282 \pm 0.036$ & 1.156 & 0.731 & 10752 & 18944 \\
4 & (3, 1) & $-0.201 \pm 0.005$ & 0.545 & 0.953 & 22912 & 64000 \\
4 & (3, 2) & $-0.110 \pm 0.007$ & -0.394 & 0.775 & 29495 & 64000 \\
4 & (3, 5) & $-0.082 \pm 0.003$ & -0.760 & 0.911 & 20608 & 64000 \\
4 & (3, 7) & $-0.132 \pm 0.004$ & -0.094 & 0.930 & 21760 & 64000 \\
\bottomrule
\end{tabular}

    \end{sc}
    \end{small}
    \end{center}
\end{table}

\clearpage

\subsection{Performance of the fully trained model as a particle simulator} \label{sec:model-performance}

\begin{figure}[h]
    \begin{center}
        \includegraphics[width=0.8\linewidth]{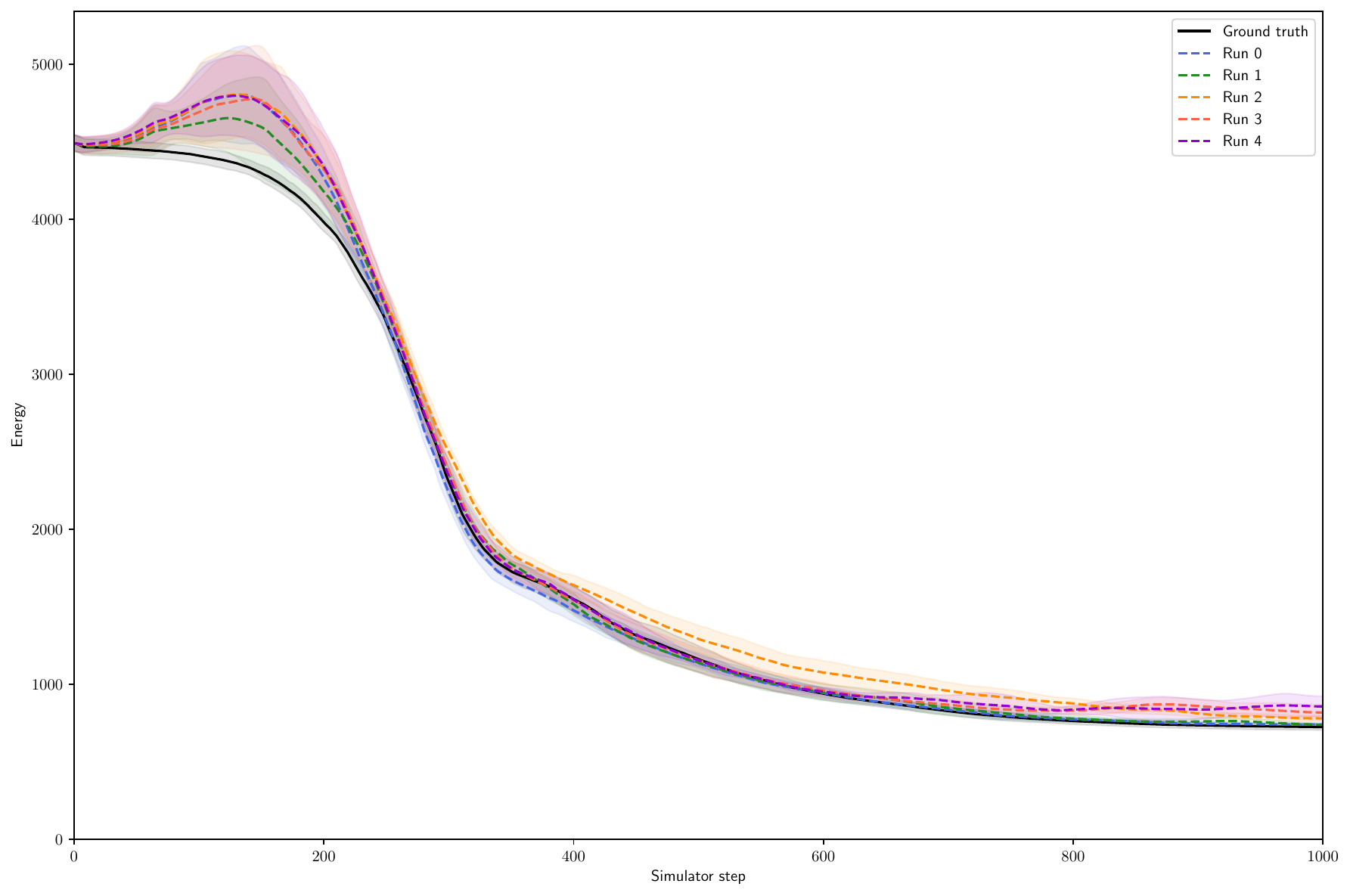}
    \end{center}
    \caption{For each run we plot the total energy of the system at 
    each state in a 1000 step simulation, averaged over different initial states. 
    Energy values are presented in reduced units, where a particle's mass and diameter are taken as the fundamental units of mass and length.
    This shows that the fully trained models have a similar behaviour to the ground 
    truth system when used as a simulator.  
    The total energy of 
    the system consists of the sum of the kinetic and gravitational potential 
    energy of each particles plus the potential energy stored during collisions 
    between particles.} \label{fig:model energy curve}
\end{figure}

To verify that the model has actually captured the true physics of the particle 
system in some way we check that it can be used to simulate a particle system 
by repeatedly applying the fully-trained model to its own output. 
For a given initial state $\ve x_0 \in \R^{N \times 2D}$ we can generate 
a sequence of states by setting $\ve x_{n+1} = f(\ve x_n; \ve w _{\text{final}})$
where $\ve w _\text{final}$ are the model's parameters at the end of a training run. 
Simulations generated in this way appear qualitatively correct compared to results 
generated by the ground-truth simulator. In \cref{fig:model energy curve} we 
show the total energy at each state in a simulation generated in this way alongside 
the ground truth, averaged over different initial states. This shows that when the 
model is used as a simulator the resulting simulations are stable (accumulated 
errors do not cause the simulation to gain unrealistic amounts of energy) and that 
on-average the energy evolves in a manner similar to the ground truth system. 

\subsection{Additional contact score plots} \label{sec:additional_contact_score_plots}

\begin{figure}[!h]
    \begin{center}
    \includegraphics[width=0.8\textwidth]{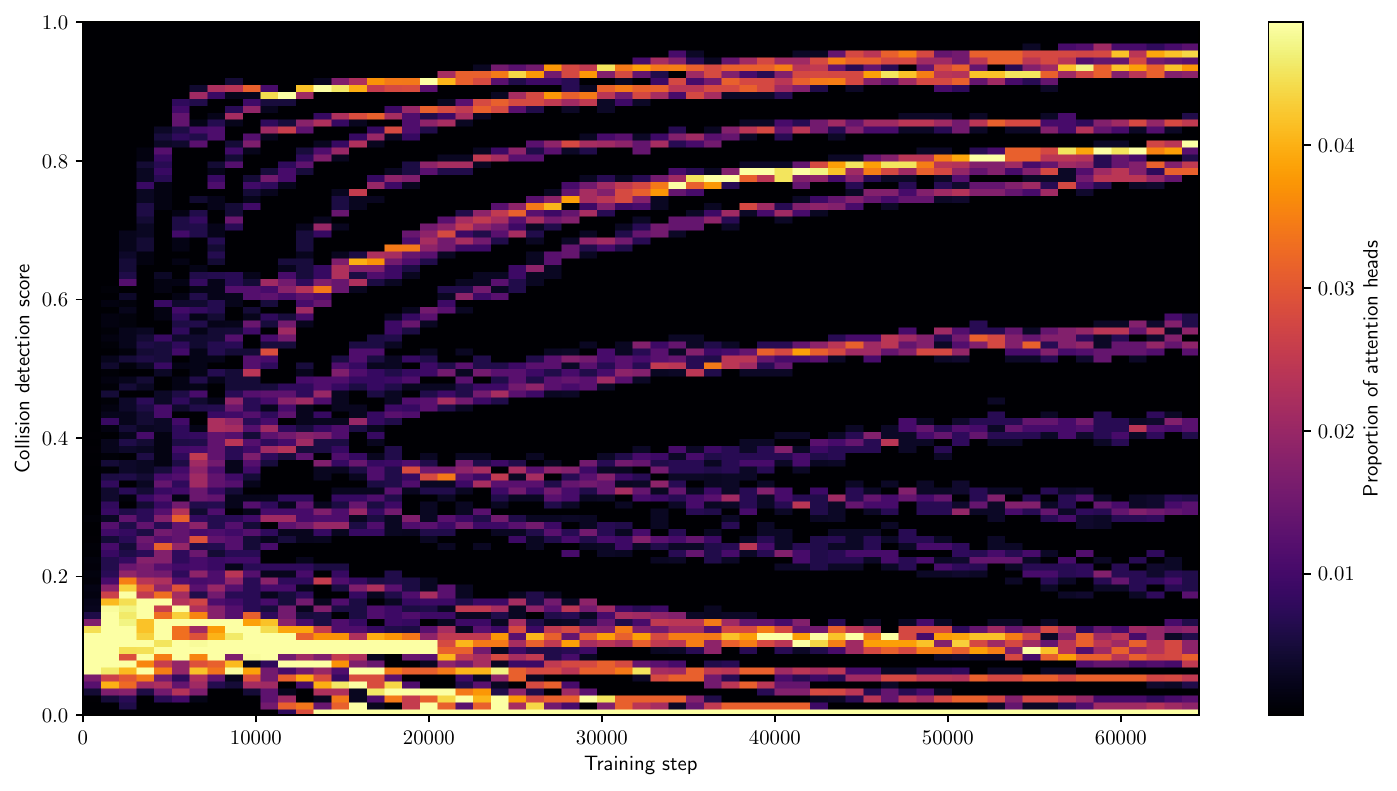}
    \end{center}
    \caption{Distribution of collision detection scores of all attention heads for training run 0.} 
    \label{fig: contact score heatmap run 0}
\end{figure}

\begin{figure}[!h]
    \begin{center}
    \includegraphics[width=0.8\textwidth]{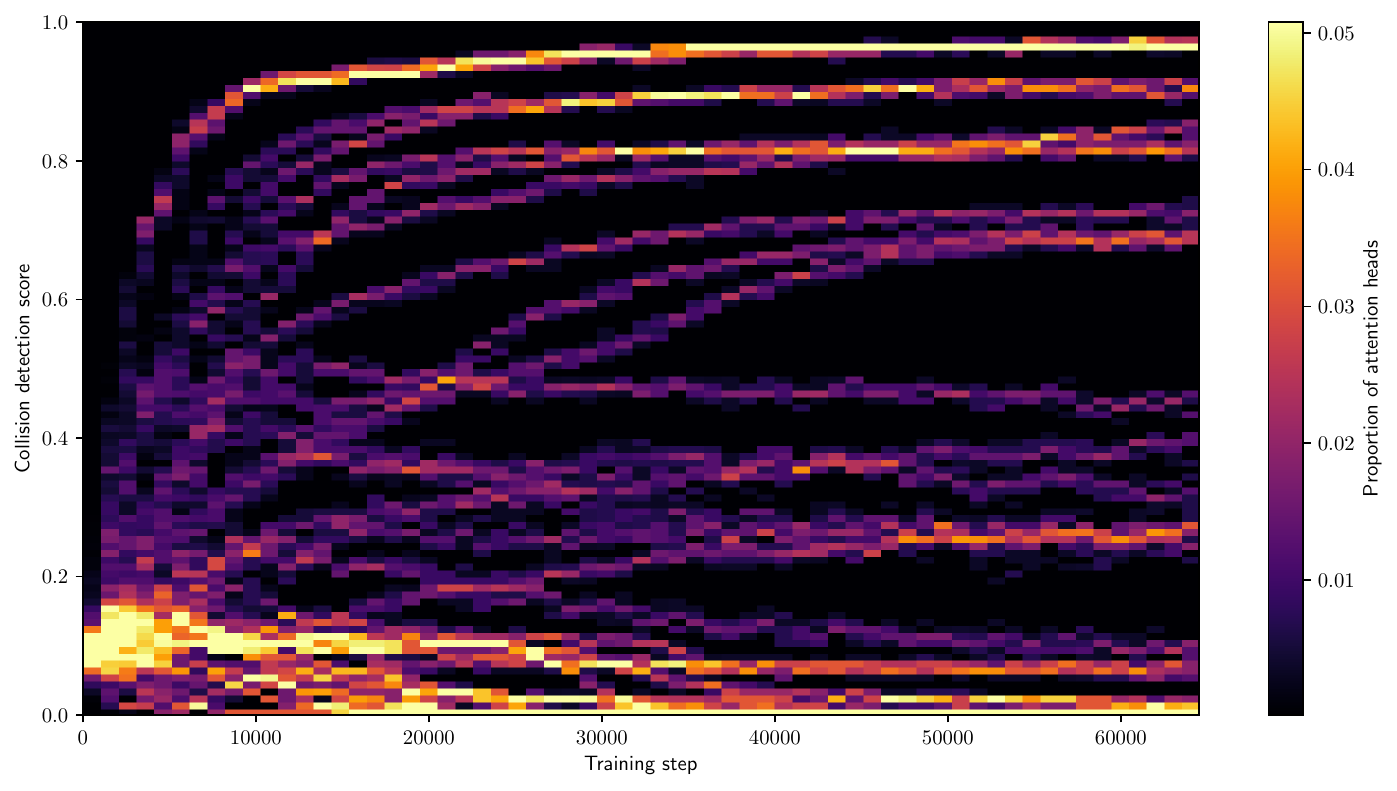}
    \end{center}
    \caption{Distribution of collision detection scores of all attention heads for training run 1.} 
    \label{fig: contact score heatmap run 1}
\end{figure}

\begin{figure}[!h]
    \begin{center}
    \includegraphics[width=0.8\textwidth]{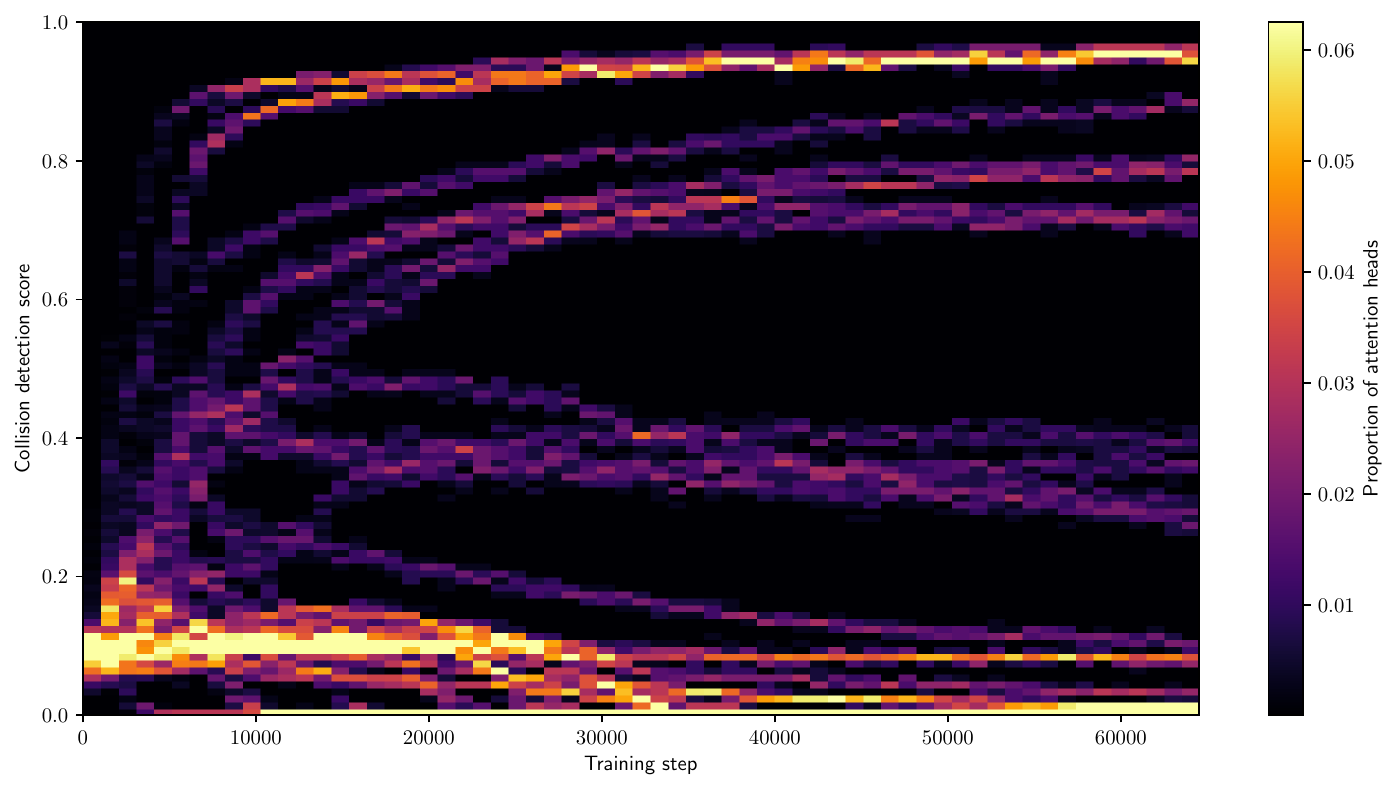}
    \end{center}
    \caption{Distribution of collision detection scores of all attention heads for training run 2.} 
    \label{fig: contact score heatmap run 2}
\end{figure}

\begin{figure}[!h]
    \begin{center}
    \includegraphics[width=0.8\textwidth]{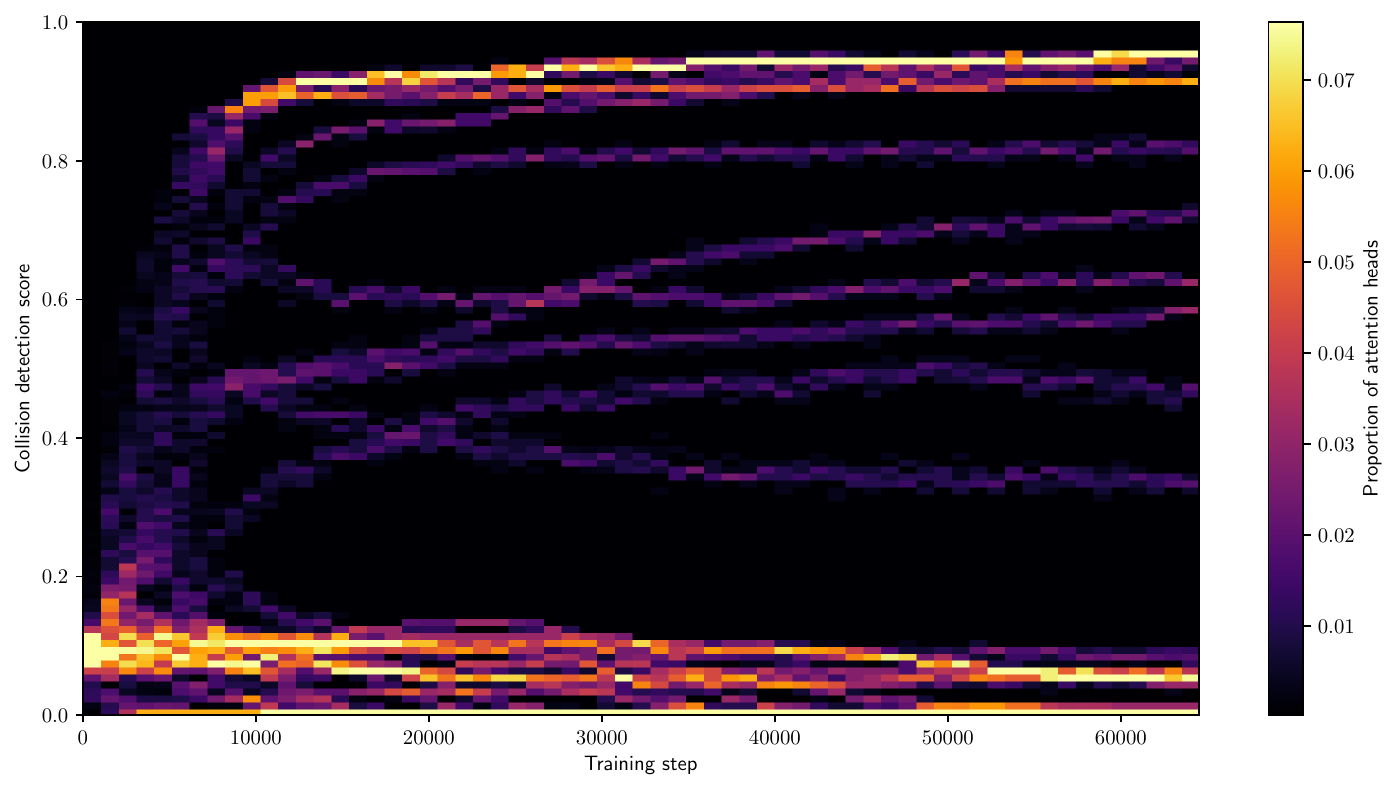}
    \end{center}
    \caption{Distribution of collision detection scores of all attention heads for training run 3.} 
    \label{fig: contact score heatmap run 3}
\end{figure}

\begin{figure}[!h]
    \begin{center}
    \includegraphics[width=0.8\textwidth]{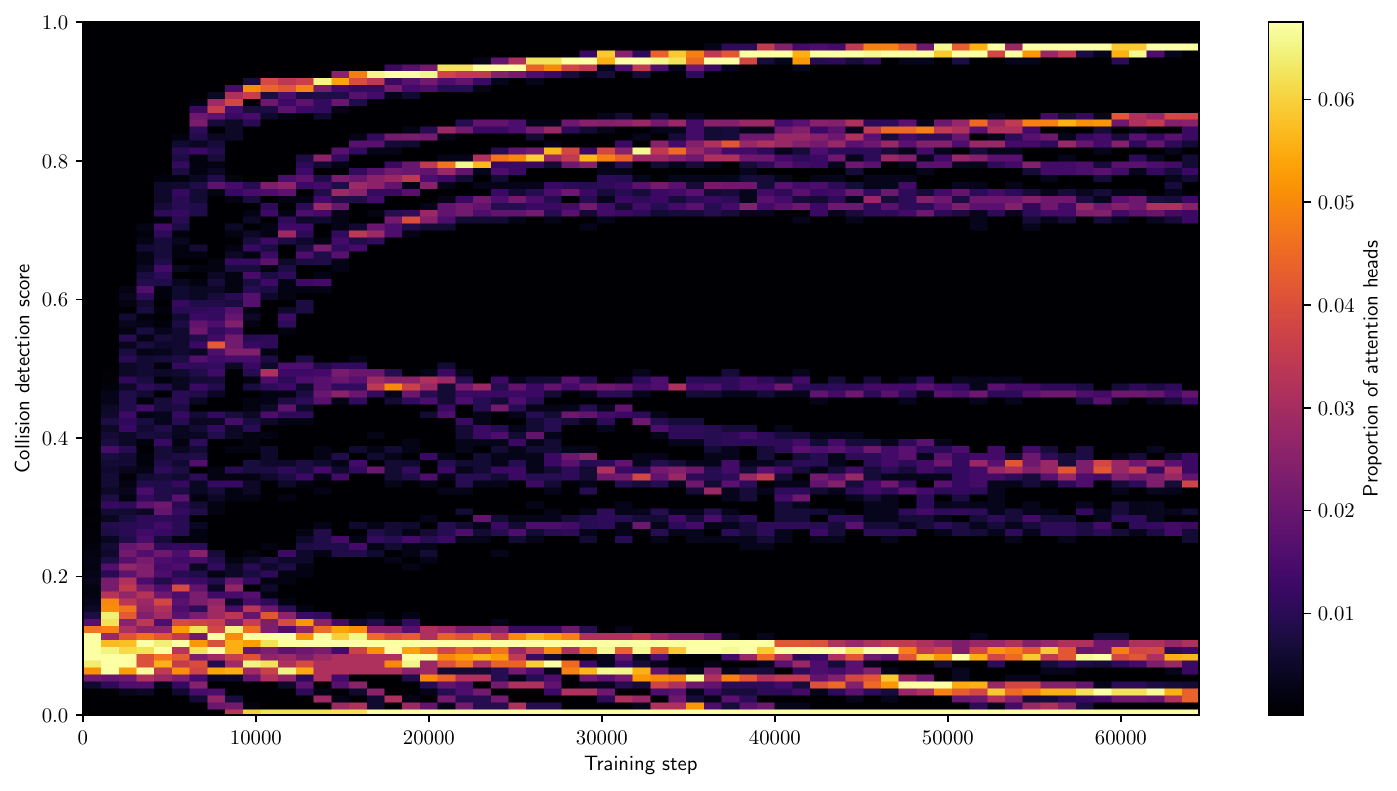}
    \end{center}
    \caption{Distribution of collision detection scores of all attention heads for training run 4.} 
    \label{fig: contact score heatmap run 4}
\end{figure}

In \cref{fig:contact scores} we showed the distribution of collision detection scores evolving 
over training for all attention heads across all five training runs. In this section we present 
a version of that figure for each of the five training runs individually, where we only show the attention 
heads from that training run (Figures \ref{fig: contact score heatmap run 0}-\ref{fig: contact score heatmap run 4}). As in \cref{fig:contact scores}, we see that in each training run 
a cluster of heads with a large collision detection score (true collision detection heads), a 
large cluster of heads with a collision detection score close to zero (heads without collision detection 
behaviour), and several smaller clusters with intermediate collision detection score (partial collision 
detection heads). In particular we observe that all training runs develop a mixture of true and 
partial collision detection heads (see also \cref{fig:head-counts-by-block}).

\FloatBarrier
\subsection{Additional attention plots} \label{sec:additional-attenrtion-plots}

Here we provide more visualisations of attention for the attention heads shown in 
\cref{fig:head behaviour}. These figures were produced by moving a single particle 
around in a given particle state, with all other particles kept in the same 
positions. We visualise the attention scores for the chosen particle only.

\begin{figure}[!h]
    \begin{center}
    \includegraphics[width=0.7\textwidth]{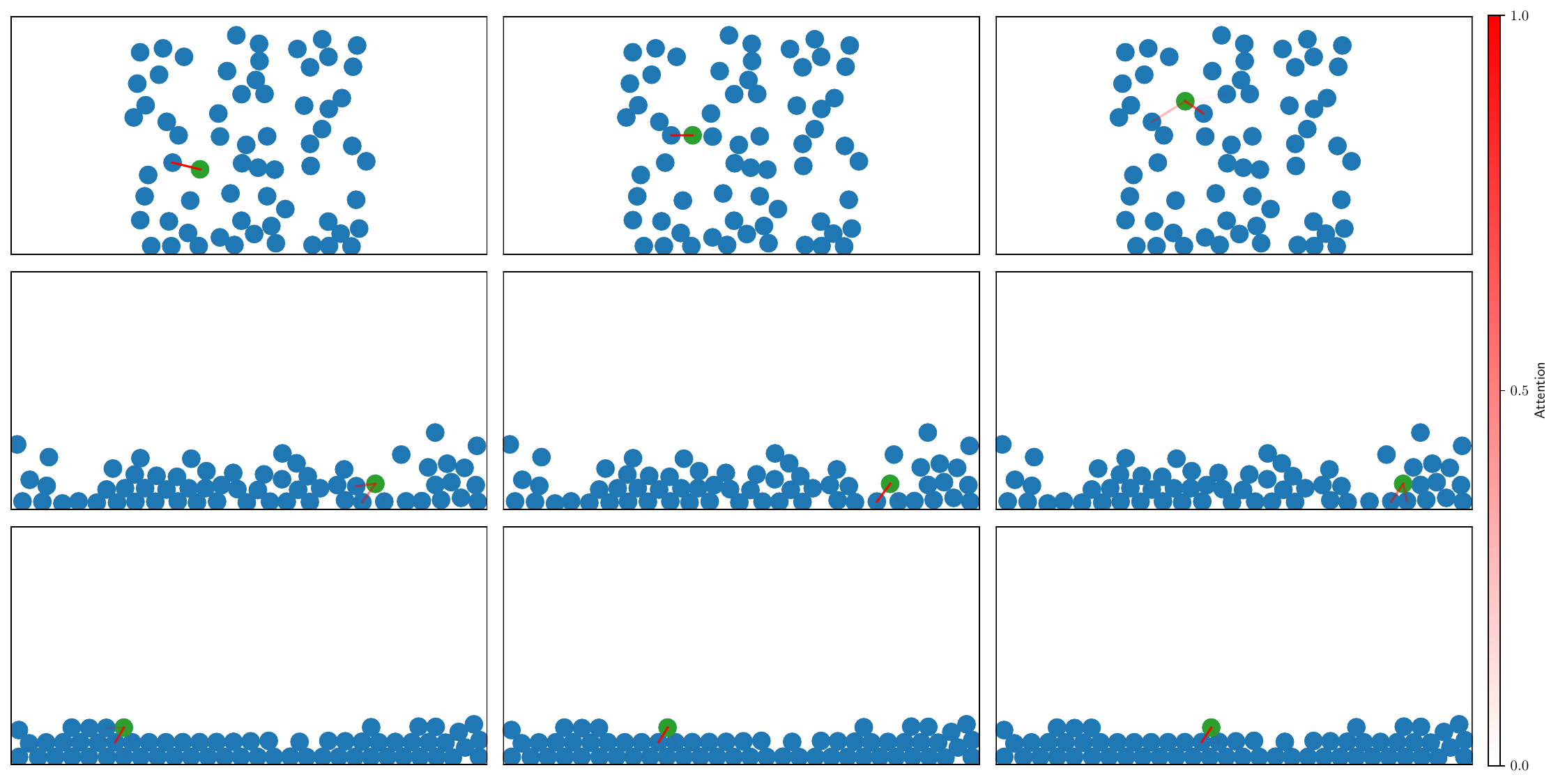}
    \end{center}
    \caption{Attention patterns of a single particle (shown in green) for head 2-7 in run 2.
    This head is a true collision detection head, and is the same head as shown in the left subplot of \cref{fig:head behaviour}.} 
    \label{fig: attn extra 2-7}
\end{figure}

\begin{figure}[h]
    \begin{center}
    \includegraphics[width=0.7\textwidth]{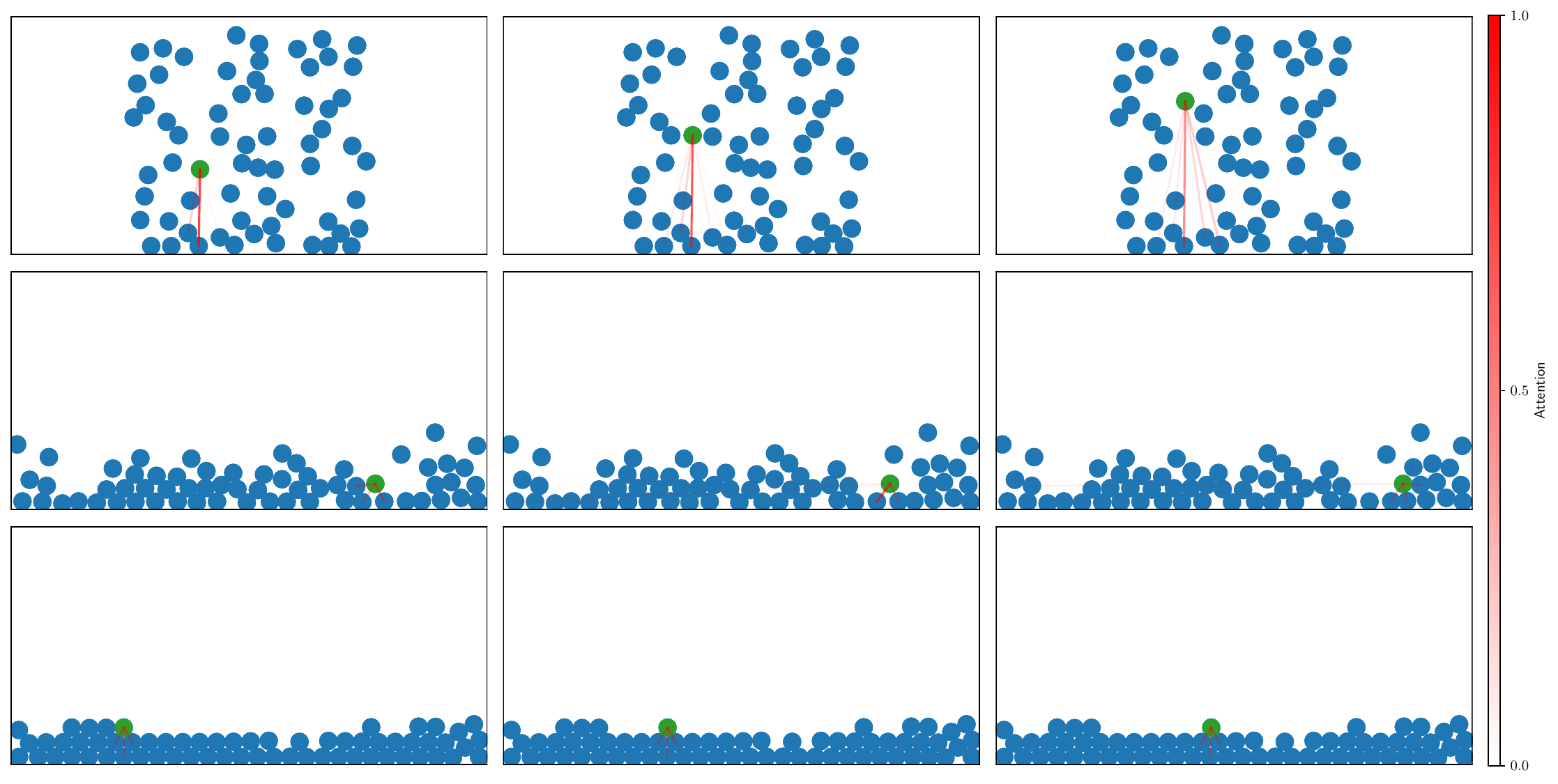}
    \end{center}
    \caption{Attention patterns of a single particle (shown in green) for head 2-5 in run 2. 
    This head is a partial collision detection head, and is the same head as shown in the middle subplot of \cref{fig:head behaviour}.}
    \label{fig: attn extra 2-5}
\end{figure}

\begin{figure}[h]
    \begin{center}
    \includegraphics[width=0.7\textwidth]{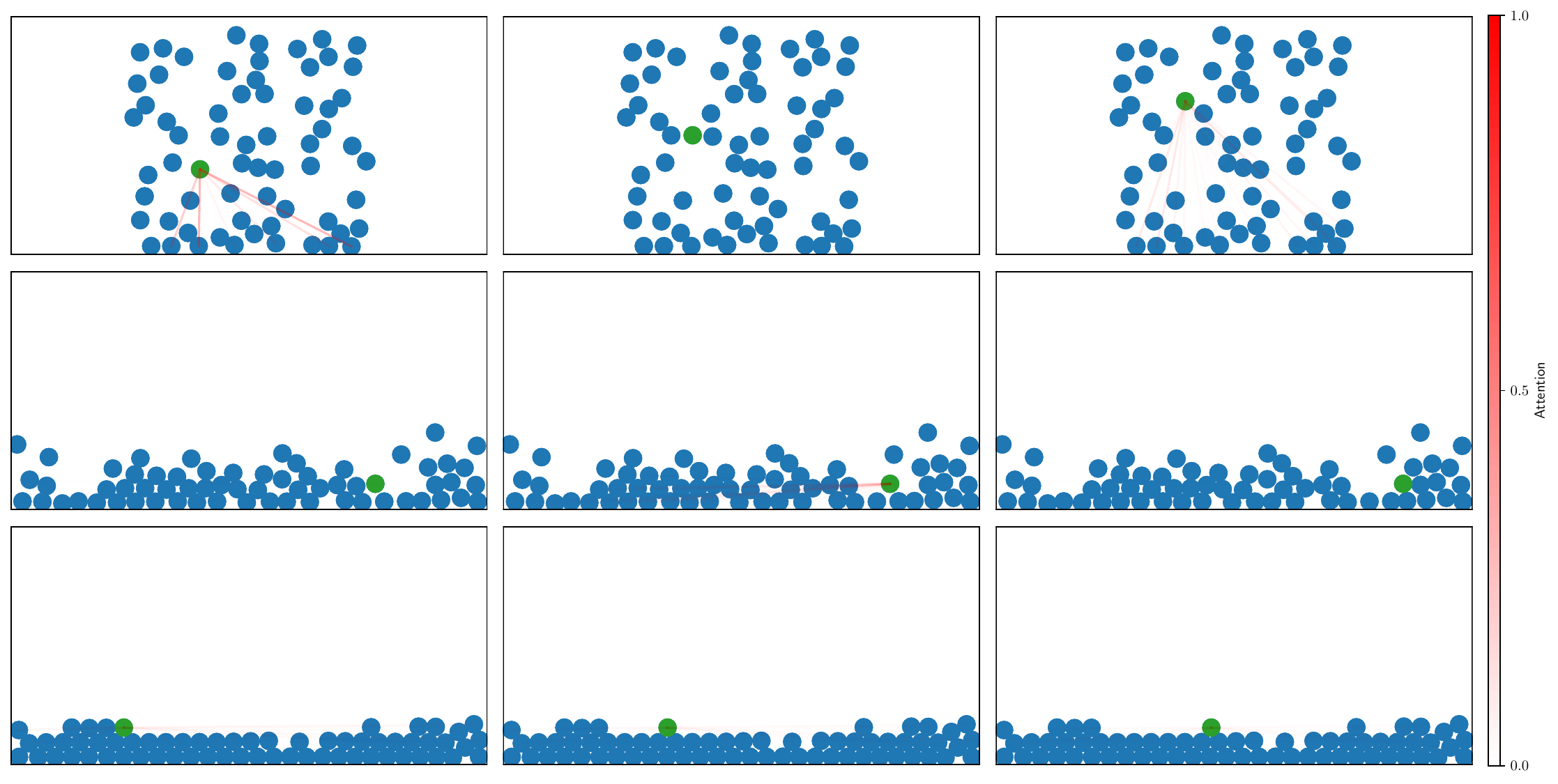}
    \end{center}
    \caption{Attention patterns of a single particle (shown in green) for head 2-5 in run 2. 
    This head is not a collision detection head, and is  the same head as shown in the right subplot of \cref{fig:head behaviour}.}
    \label{fig: attn extra 2-6}
\end{figure}

\clearpage

\subsection{Additional correlation and local learning coefficient results} \label{sec:additional corr and llc}

We give additional attention-distance correlation and local learning coefficient results, similar to 
\cref{fig:heads governed by degenerate potential}. So that LLC estimates can be 
compared we needed to use a single set of LLC estimation 
hyperparameters (given in \cref{table:llc hyperparameters main}) for all attention 
heads. 
It was not possible to find a single set of hyperparamters which produced 
acceptable trace plots for all heads, for all model checkpoints. For this reason 
only we only show plots for a subset of the collision detection heads in this 
section.

\begin{figure}[h]
    \begin{center}
    \includegraphics[width=0.83\textwidth]{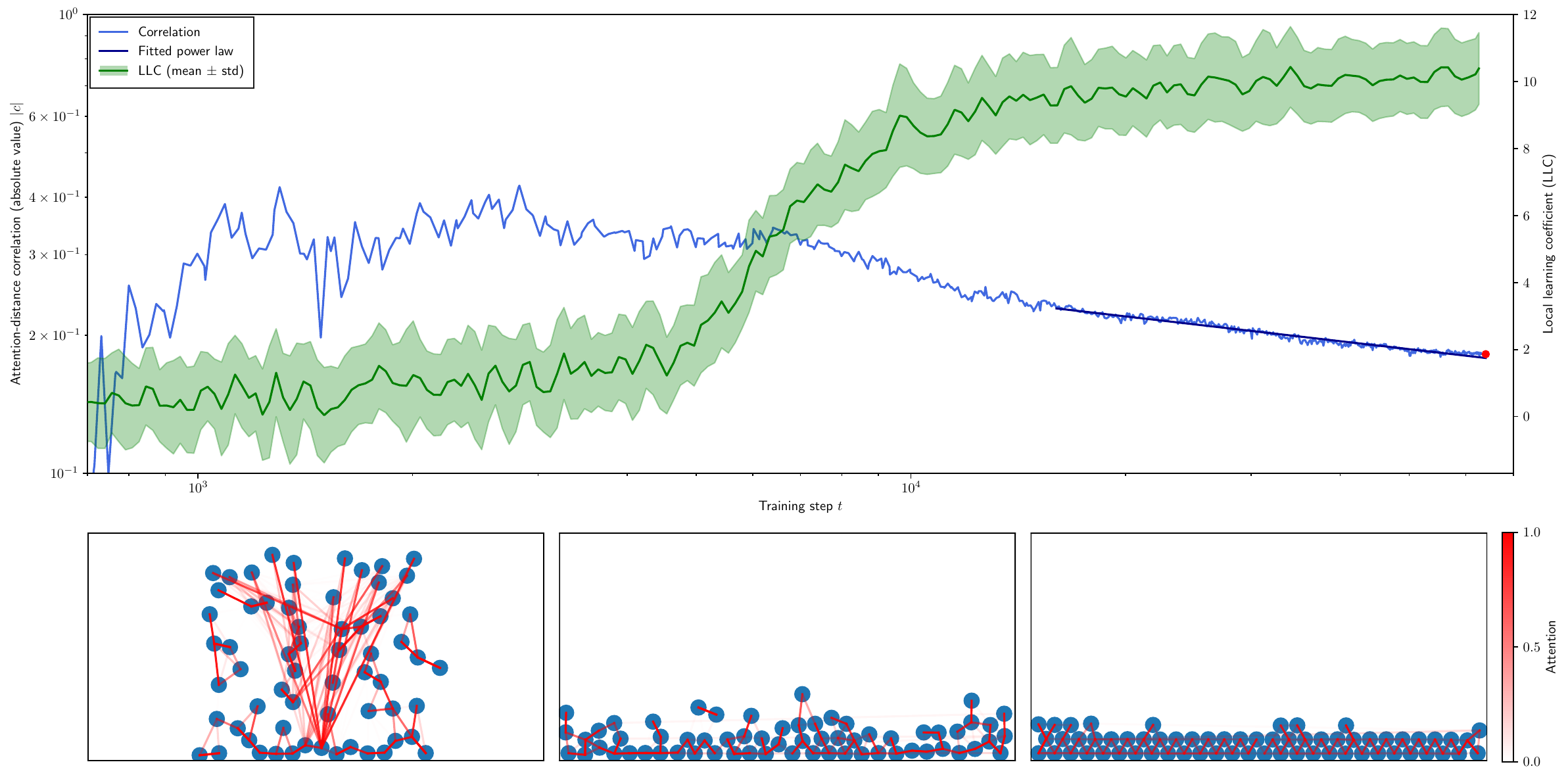}
    \end{center}
    \caption{Attention-distance correlation and local learning coefficient results for head 2-5 in training run 0. The fitted curve is $\log(|c|) = -0.179 \log(t) + 0.263$ with $R^2 = 0.961$ (see \cref{table:power law exponents}).} \label{fig: extra 0-2-5}
\end{figure}

\begin{figure}[h]
    \begin{center}
    \includegraphics[width=0.83\textwidth]{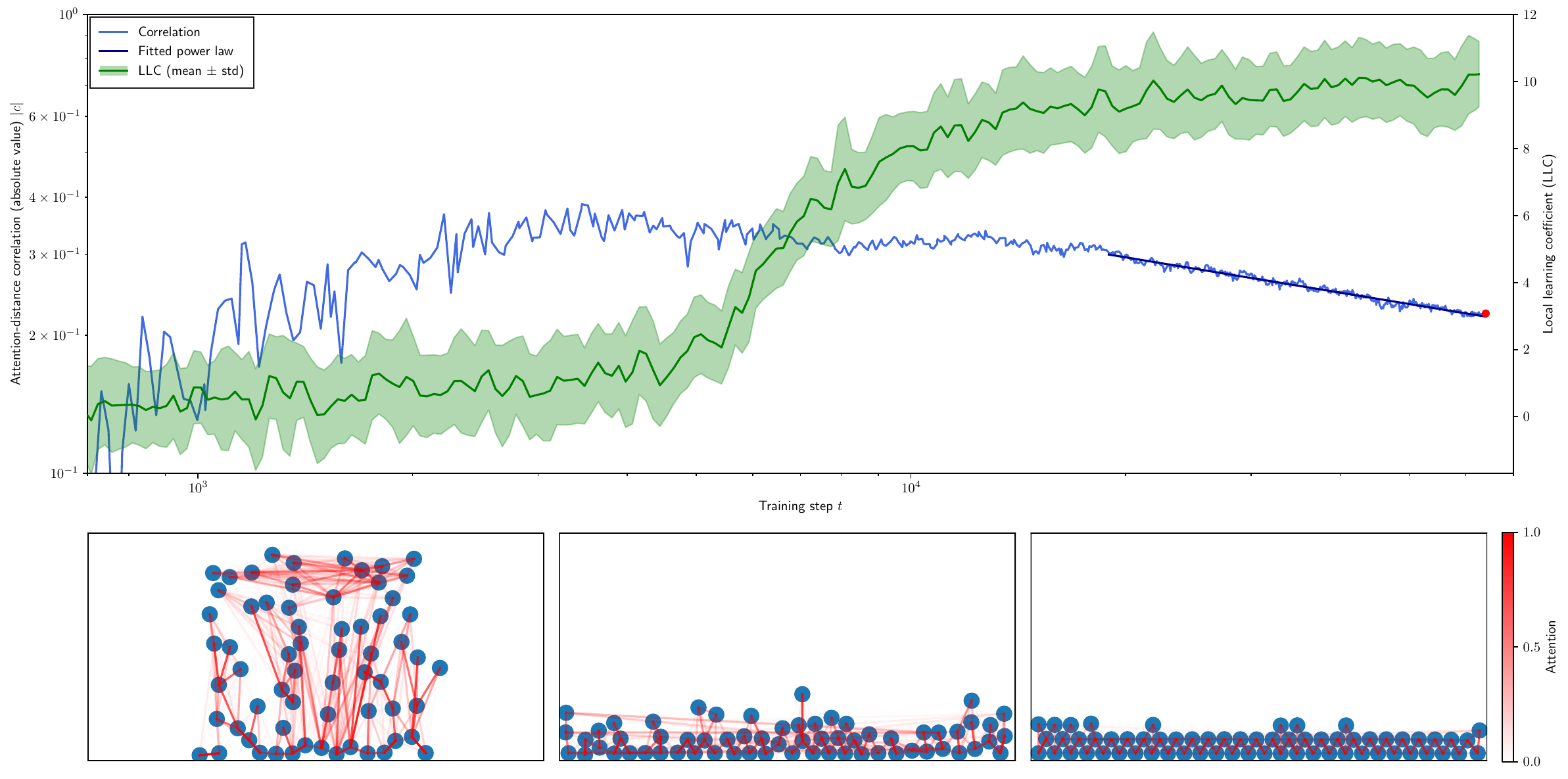}
    \end{center}
    \caption{Attention-distance correlation and local learning coefficient results for head 2-6 in training run 0. The fitted curve is $\log(|c|) = -0.256 \log(t) + 1.317$ with $R^2 = 0.975$ (see \cref{table:power law exponents}).} \label{fig: extra 0-2-6}
\end{figure}

\begin{figure}[h]
    \begin{center}
    \includegraphics[width=0.83\textwidth]{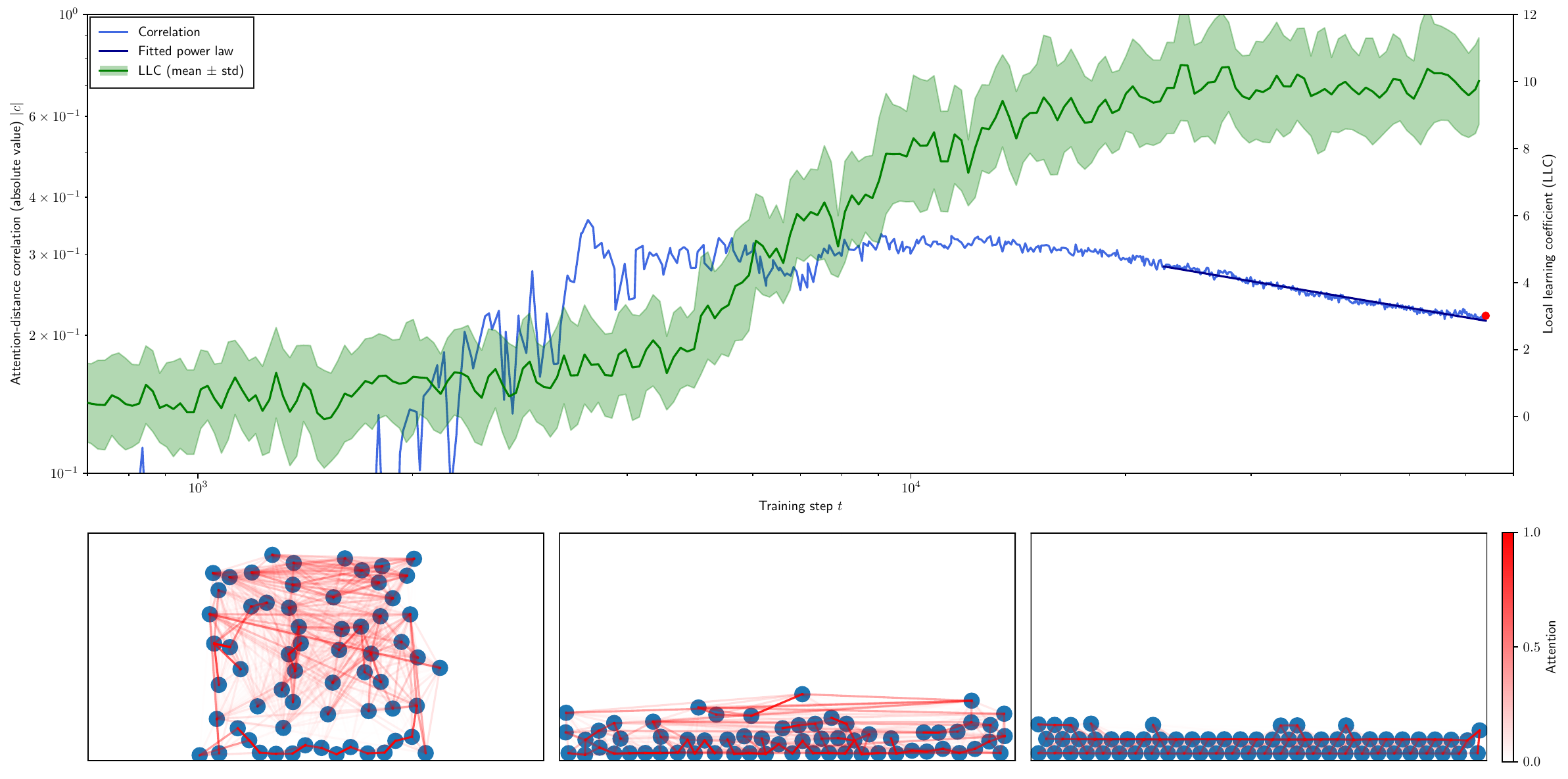}
    \end{center}
    \caption{Attention-distance correlation and local learning coefficient results for head 3-1 in training run 0. The fitted curve is $\log(|c|) = -0.262 \log(t) + 1.363$ with $R^2 = 0.971$ (see \cref{table:power law exponents}).} \label{fig: extra 0-3-1}
\end{figure}

\begin{figure}[h]
    \begin{center}
    \includegraphics[width=0.83\textwidth]{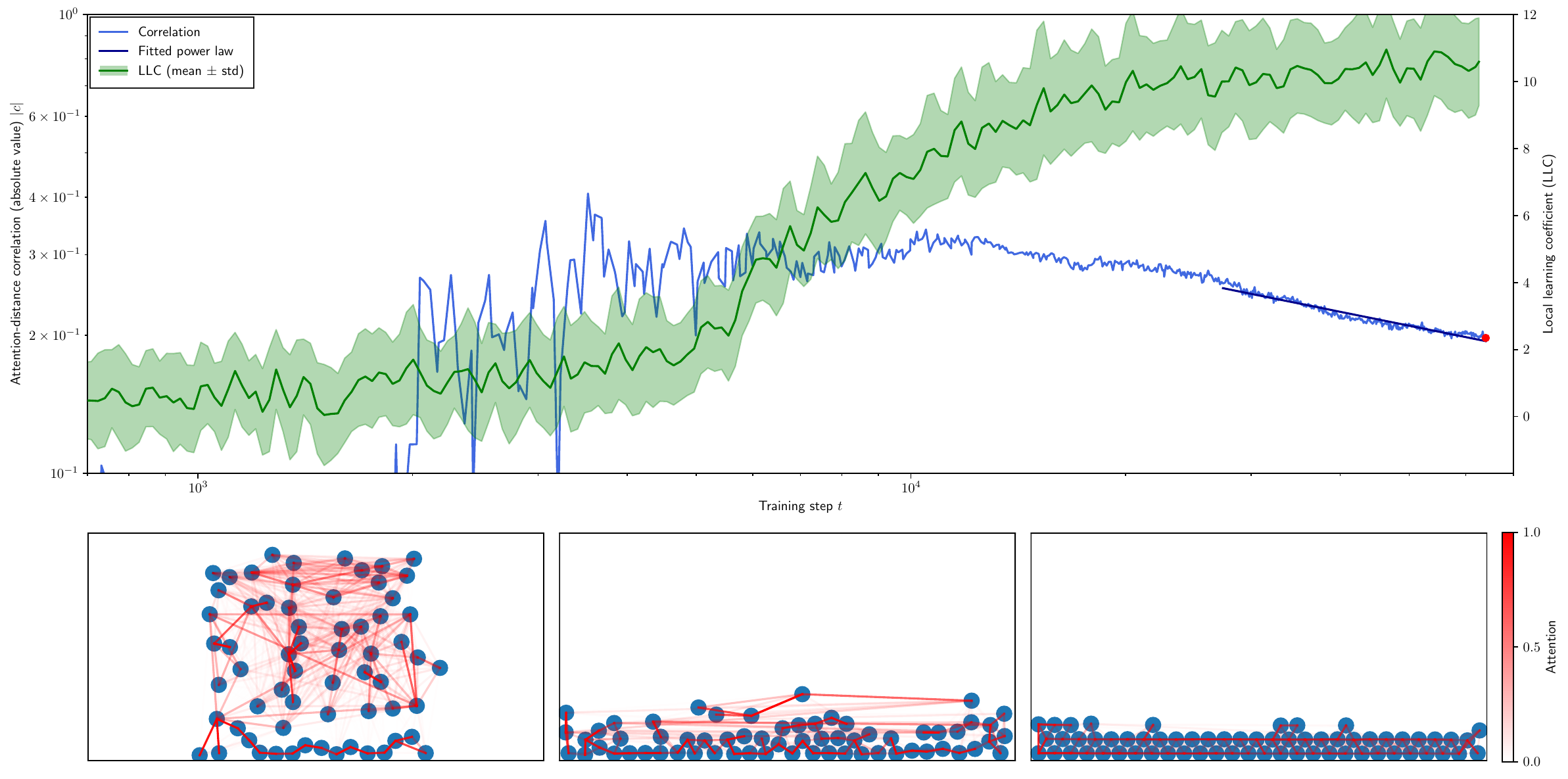}
    \end{center}
    \caption{Attention-distance correlation and local learning coefficient results for head 3-3 in training run 0. The fitted curve is $\log(|c|) = -0.313 \log(t) + 1.827$ with $R^2 = 0.950$ (see \cref{table:power law exponents}).} \label{fig: extra 0-3-3}
\end{figure}

\begin{figure}[h]
    \begin{center}
    \includegraphics[width=0.83\textwidth]{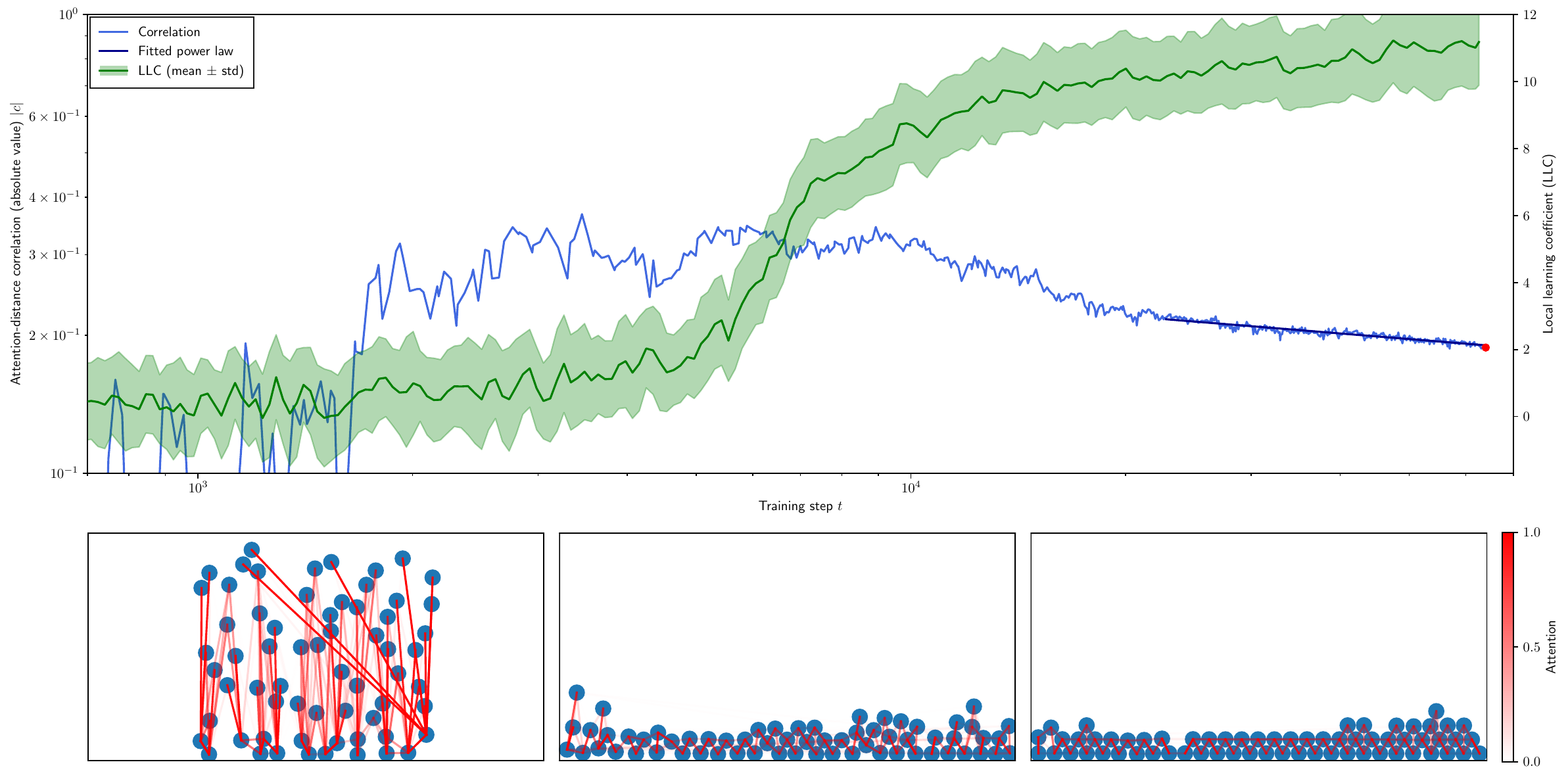}
    \end{center}
    \caption{Attention-distance correlation and local learning coefficient results for head 1-0 in training run 1. The fitted curve is $\log(|c|) = -0.127 \log(t) - 0.259$ with $R^2 = 0.895$ (see \cref{table:power law exponents}).} \label{fig: extra 1-1-0}
\end{figure}

\begin{figure}[h]
    \begin{center}
    \includegraphics[width=0.83\textwidth]{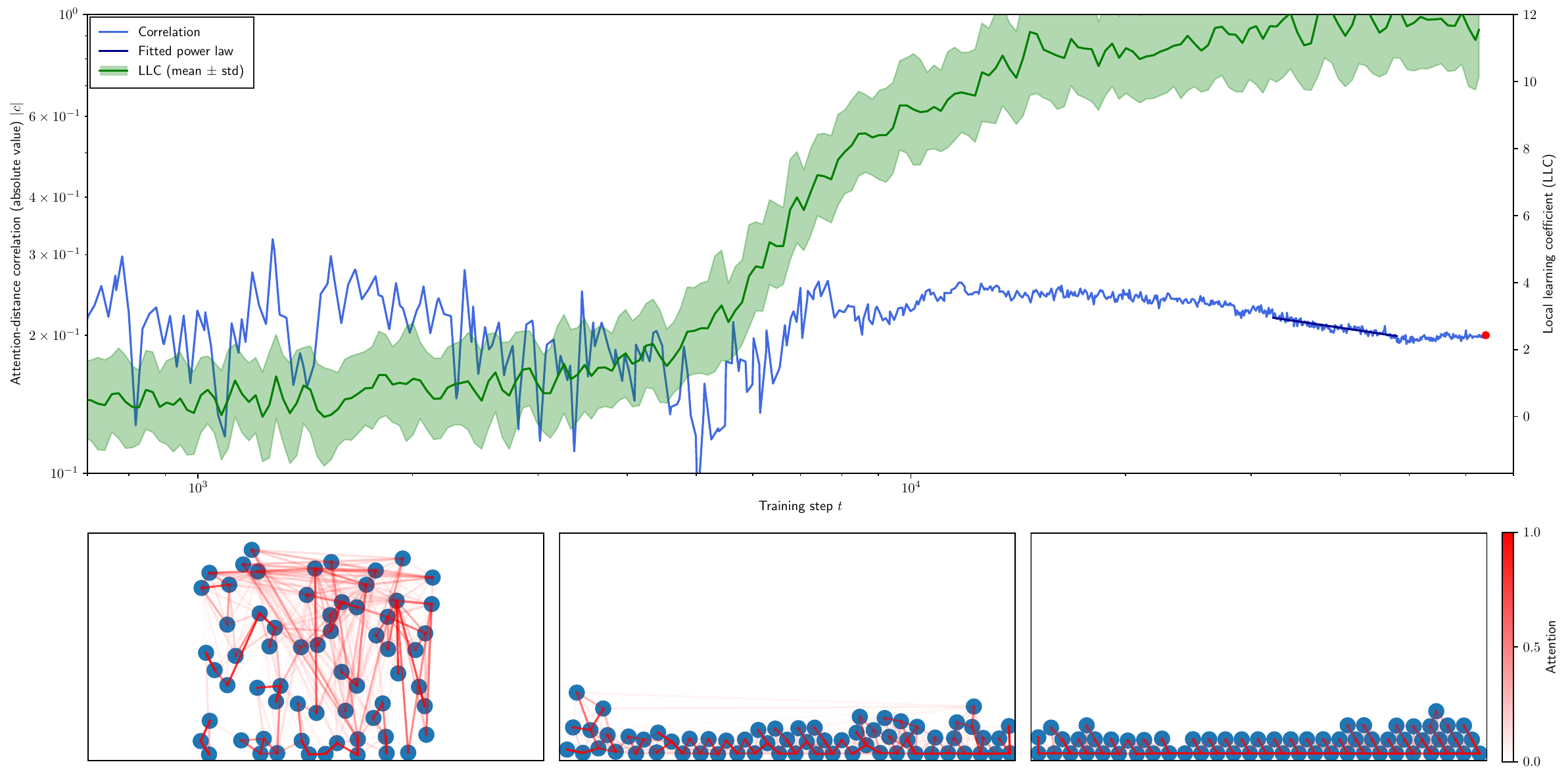}
    \end{center}
    \caption{Attention-distance correlation and local learning coefficient results for head 2-3 in training run 1. The fitted curve is $\log(|c|) = -0.231 \log(t) + 0.879$ with  $R^2 = 0.801$ (see \cref{table:power law exponents}).} \label{fig: extra 1-2-3}
\end{figure}

\begin{figure}[h]
    \begin{center}
    \includegraphics[width=0.83\textwidth]{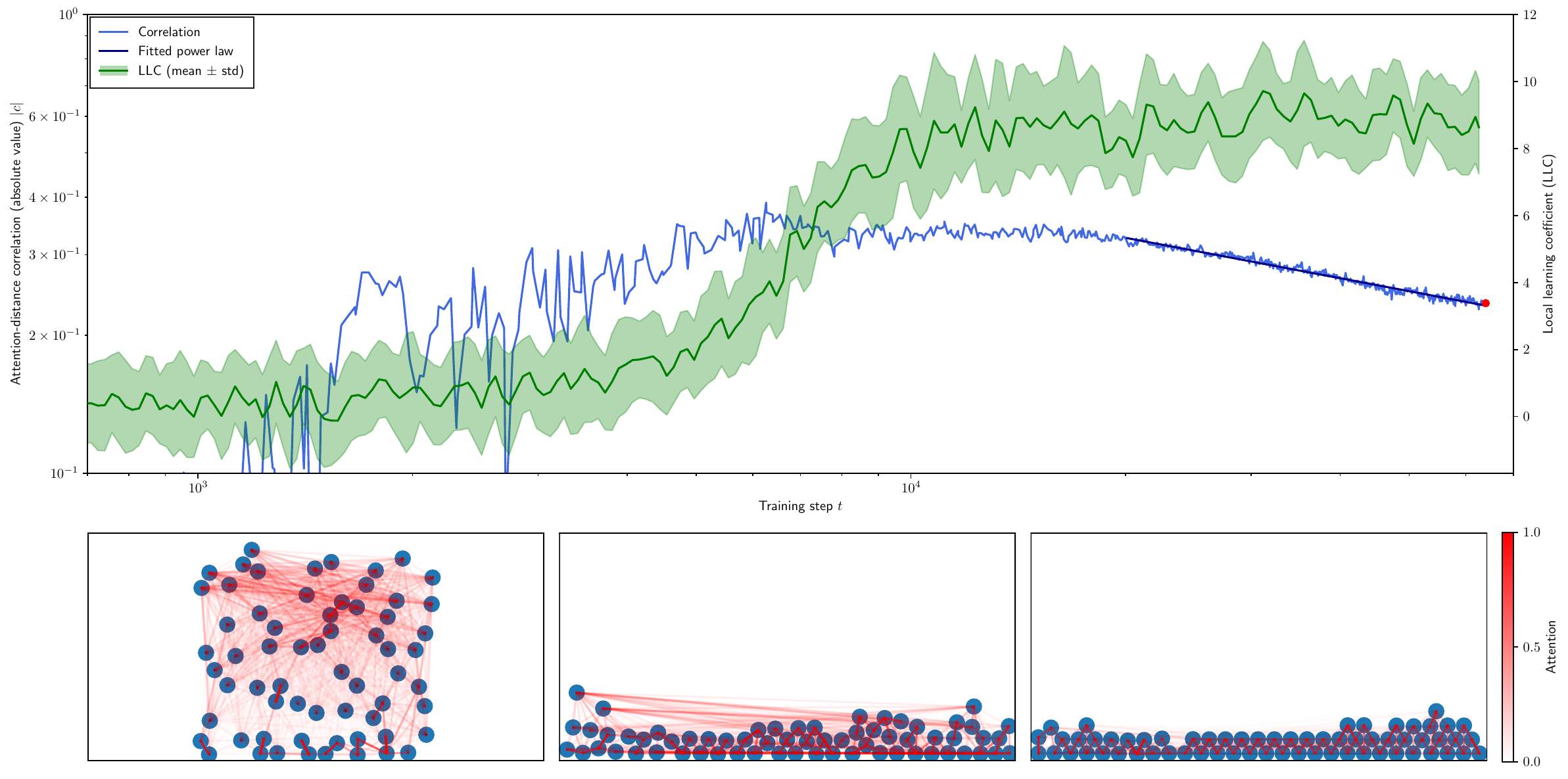}
    \end{center}
    \caption{Attention-distance correlation and local learning coefficient results for head 3-4 in training run 1. The fitted curve is $\log(|c|) = -0.293 \log(t) + 1.777$ with $R^2 = 0.975$ (see \cref{table:power law exponents}).} \label{fig: extra 1-3-4}
\end{figure}

\begin{figure}[h]
    \begin{center}
    \includegraphics[width=0.83\textwidth]{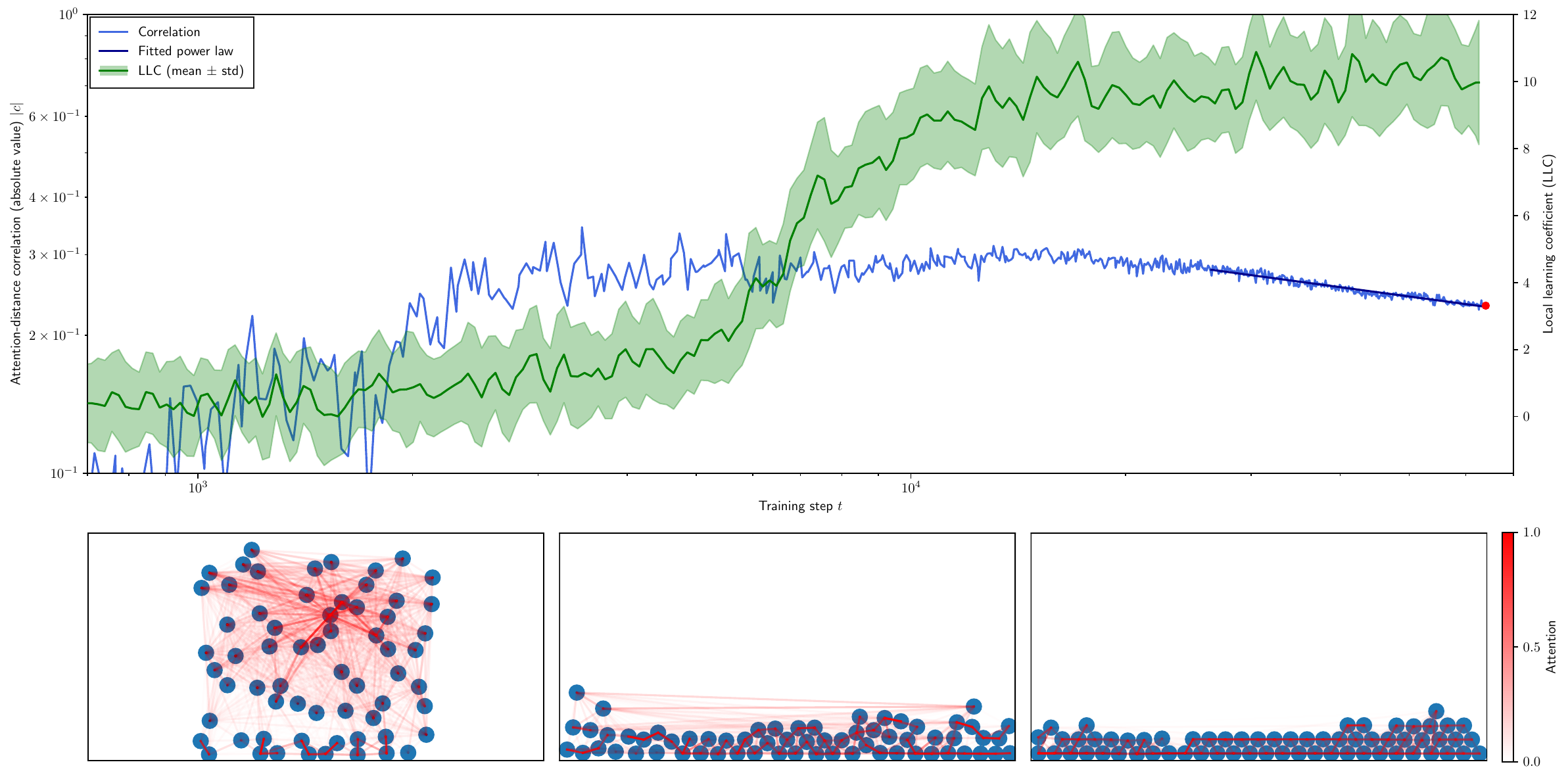}
    \end{center}
    \caption{Attention-distance correlation and local learning coefficient results for head 3-6 in training run 1. The fitted curve is $\log(|c|) = -0.210 \log(t) + 0.879$ with $R^2 = 0.932$ (see \cref{table:power law exponents}).} \label{fig: extra 1-3-6}
\end{figure}

\begin{figure}[h]
    \begin{center}
    \includegraphics[width=0.83\textwidth]{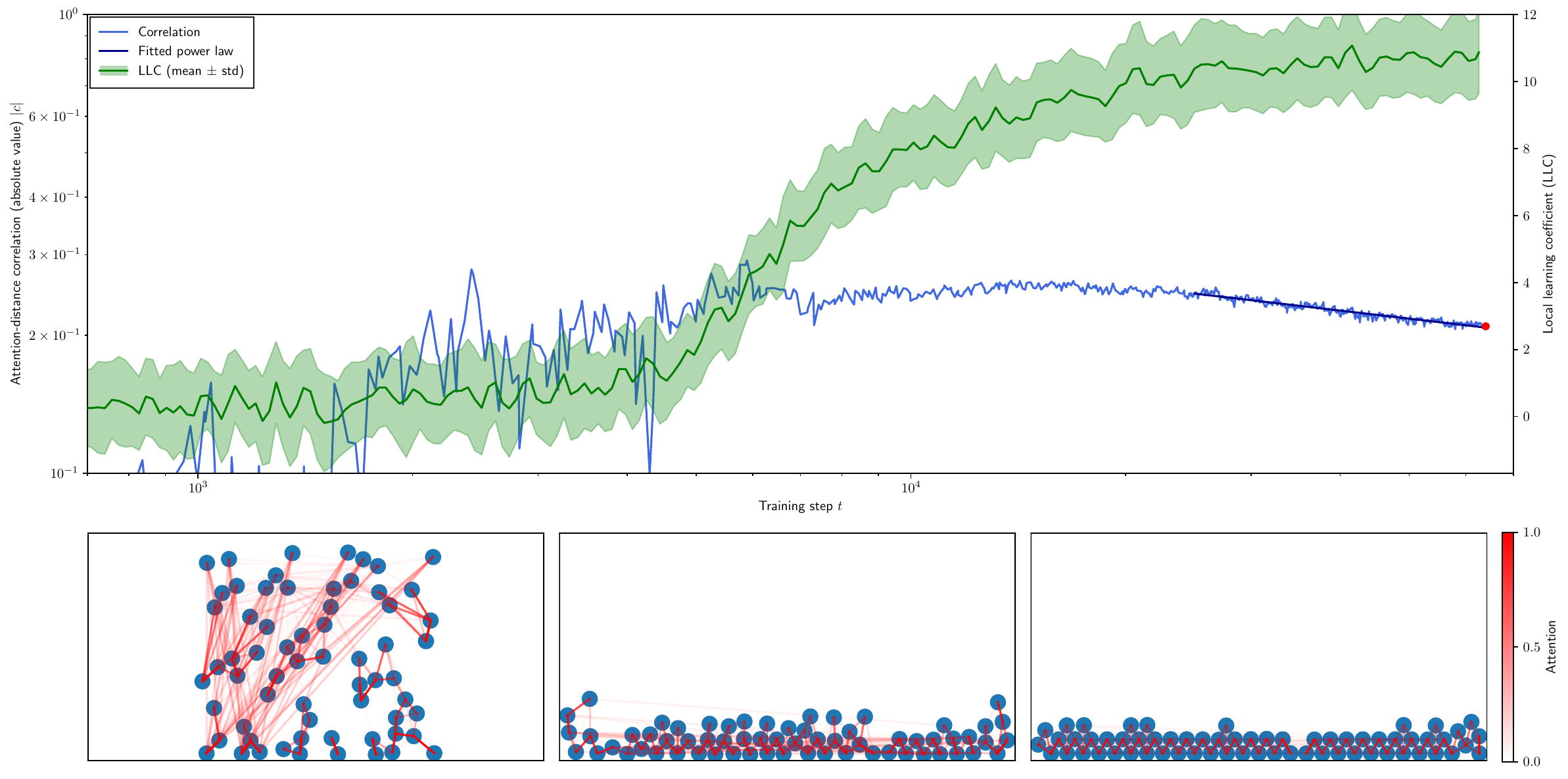}
    \end{center}
    \caption{Attention-distance correlation and local learning coefficient results for head 2-0 in training run 2. The fitted curve is $\log(|c|) = -0.182 \log(t) + 0.438$ with $R^2 = 0.923$ (see \cref{table:power law exponents}).} \label{fig: extra 2-2-0}
\end{figure}

\begin{figure}[h]
    \begin{center}
    \includegraphics[width=0.83\textwidth]{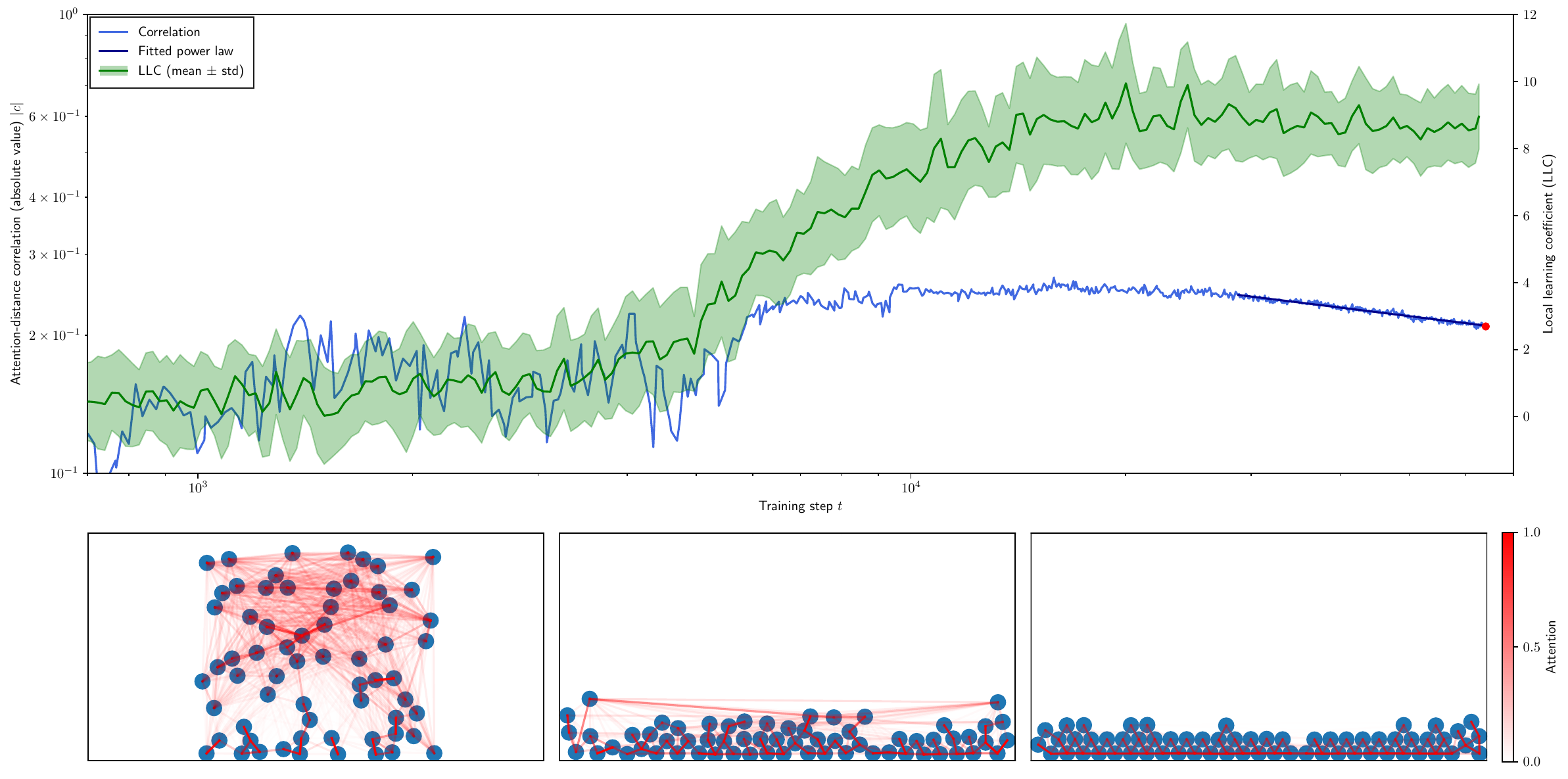}
    \end{center}
    \caption{Attention-distance correlation and local learning coefficient results for head 3-0 in training run 2. The fitted curve is $\log(|c|) = -0.193 \log(t) + 0.574$ with $R^2 = 0.954$ (see \cref{table:power law exponents}).} \label{fig: extra 2-3-0}
\end{figure}

\begin{figure}[h]
    \begin{center}
    \includegraphics[width=0.83\textwidth]{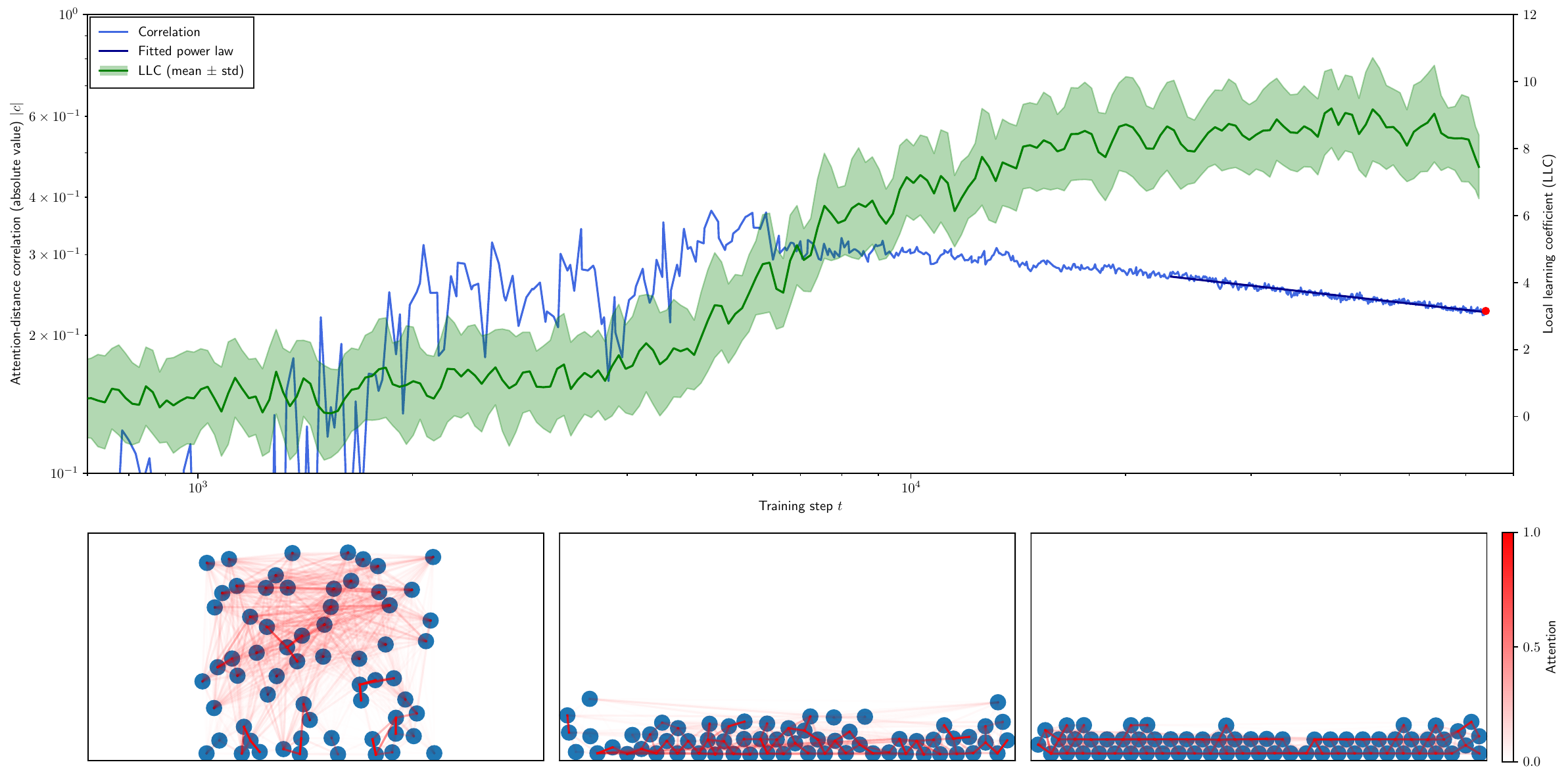}
    \end{center}
    \caption{Attention-distance correlation and local learning coefficient results for head 3-1 in training run 2. The fitted curve is $\log(|c|) = -0.175 \log(t) + 0.448$ with $R^2 = 0.947$ (see \cref{table:power law exponents}).} \label{fig: extra 2-3-1}
\end{figure}

\begin{figure}[h]
    \begin{center}
    \includegraphics[width=0.83\textwidth]{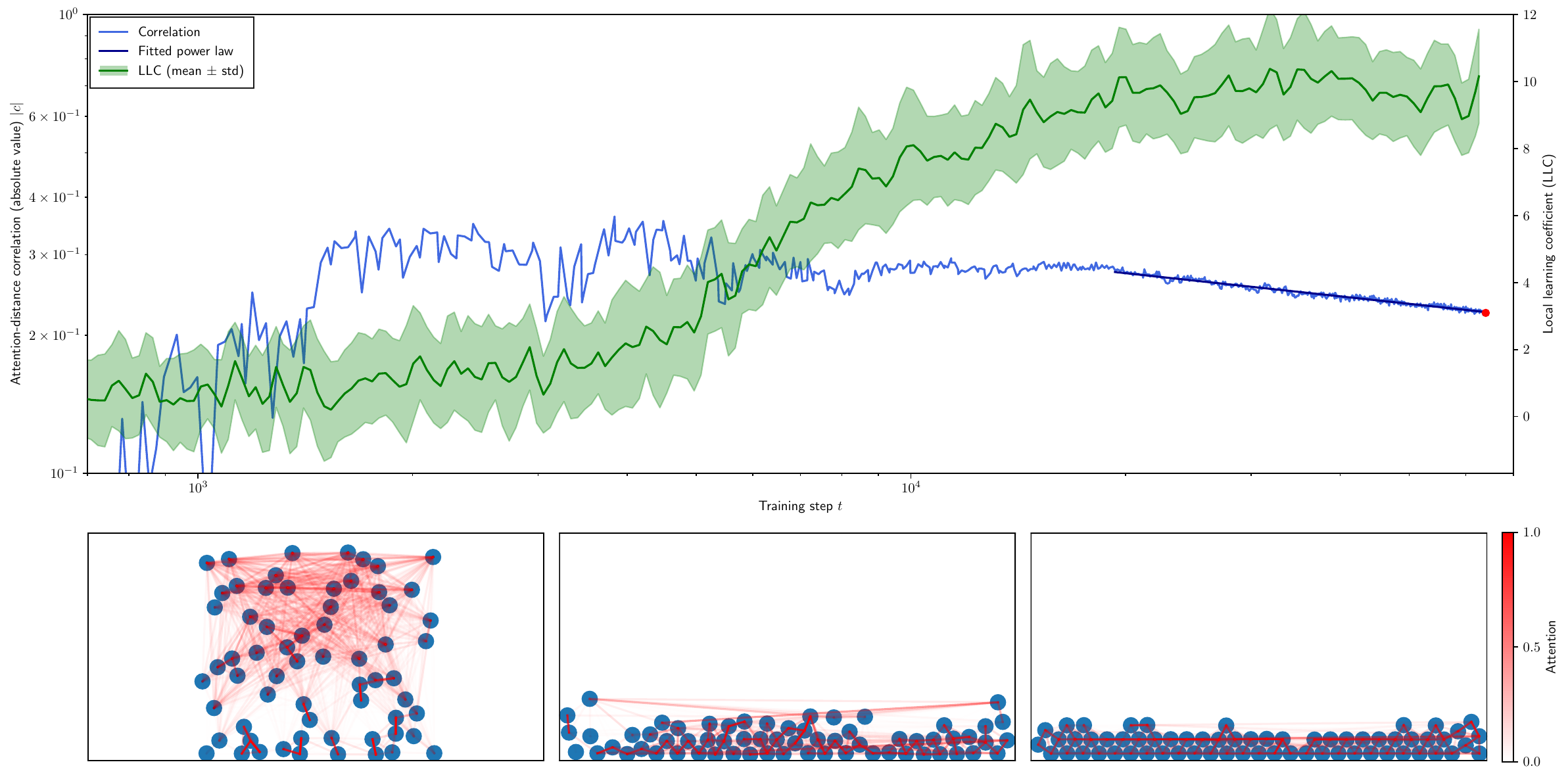}
    \end{center}
    \caption{Attention-distance correlation and local learning coefficient results for head 3-6 in training run 2. The fitted curve is $\log(|c|) = -0.168 \log(t) + 0.364$ with $R^2 = 0.964$ (see \cref{table:power law exponents}).} \label{fig: extra 2-3-0}
\end{figure}

\begin{figure}[h]
    \begin{center}
    \includegraphics[width=0.83\textwidth]{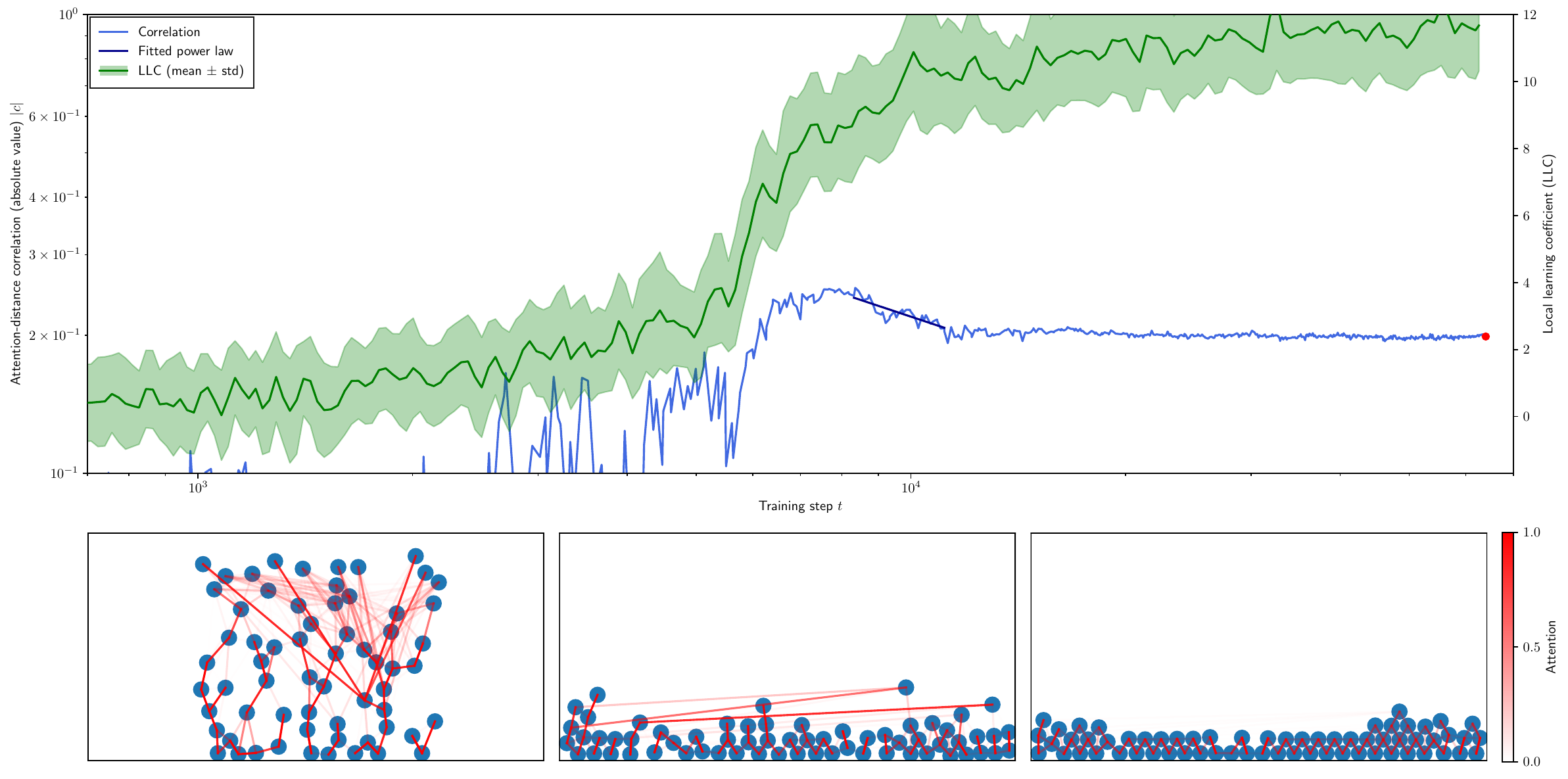}
    \end{center}
    \caption{Attention-distance correlation and local learning coefficient results for head 2-2 in training run 3. The fitted curve is $\log(|c|) = -0.516 \log(t) + 3.240$ with $R^2 = 0.715$ (see \cref{table:power law exponents}).} \label{fig: extra 3-2-2}
\end{figure}

\begin{figure}[h]
    \begin{center}
    \includegraphics[width=0.83\textwidth]{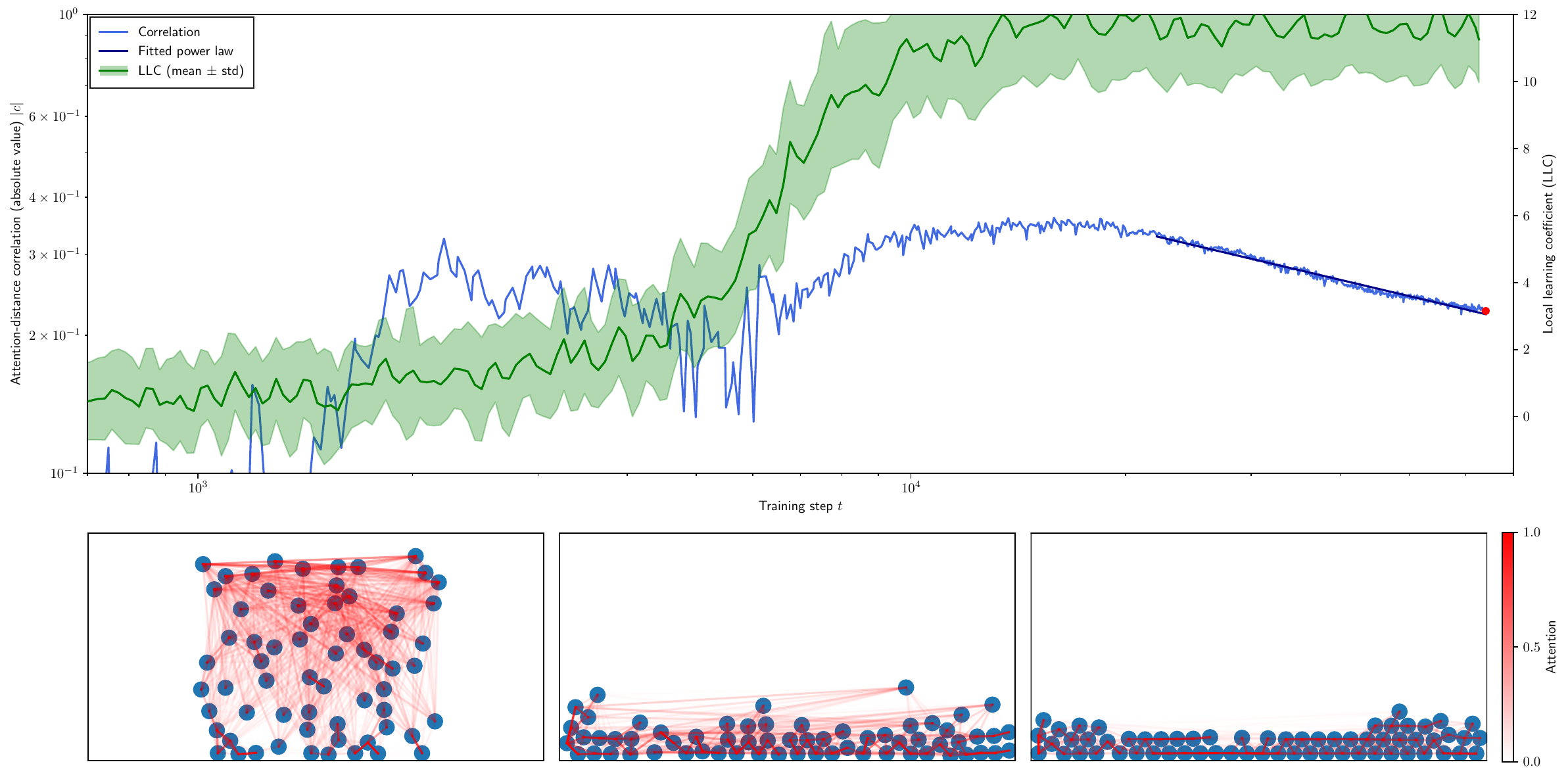}
    \end{center}
    \caption{Attention-distance correlation and local learning coefficient results for head 3-2 in training run 3. The fitted curve is $\log(|c|) = -0.367 \log(t) + 2.559$ with $R^2 = 0.982$ (see \cref{table:power law exponents}).} \label{fig: extra 3-3-2}
\end{figure}

\begin{figure}[h]
    \begin{center}
    \includegraphics[width=0.83\textwidth]{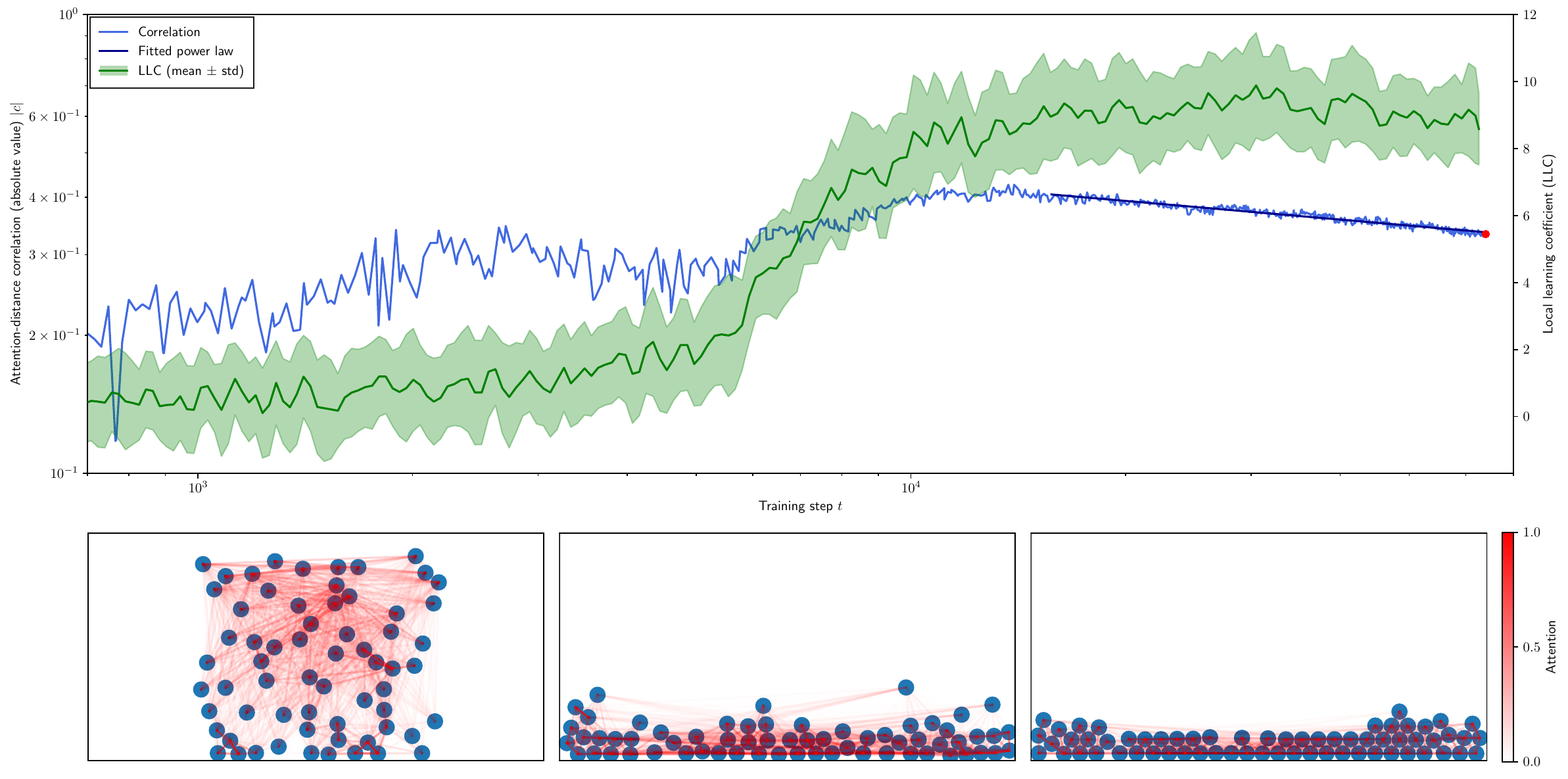}
    \end{center}
    \caption{Attention-distance correlation and local learning coefficient results for head 3-3 in training run 3. The fitted curve is $\log(|c|) = -0.135 \log(t) + 0.405$ with $R^2 = 0.936$ (see \cref{table:power law exponents}).} \label{fig: extra 3-3-3}
\end{figure}

\begin{figure}[h]
    \begin{center}
    \includegraphics[width=0.83\textwidth]{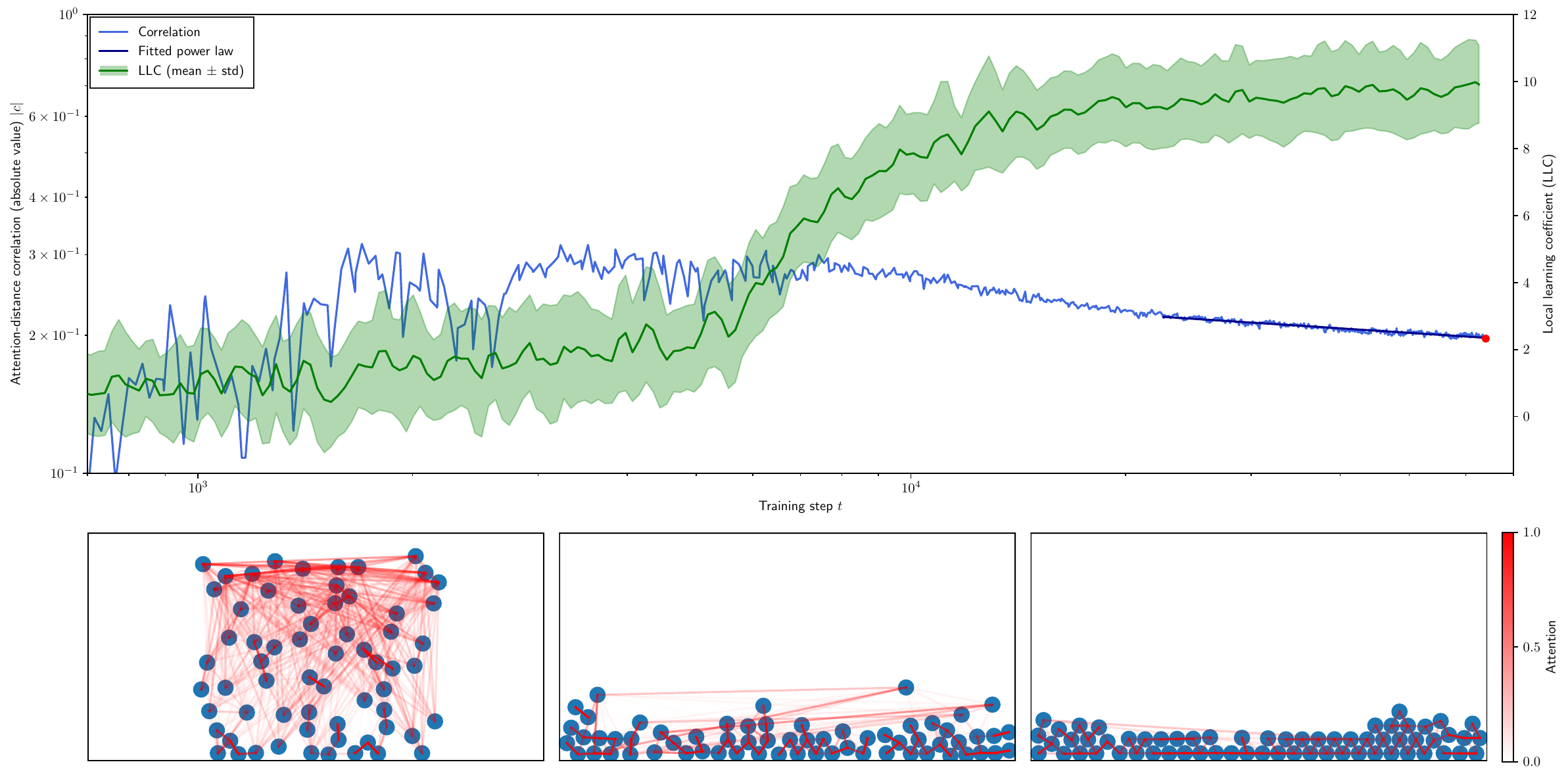}
    \end{center}
    \caption{Attention-distance correlation and local learning coefficient results for head 3-6 in training run 3. The fitted curve is $\log(|c|) = -0.101 \log(t) -0.505$ with $R^2 = 0.904$ (see \cref{table:power law exponents}).} \label{fig: extra 3-3-6}
\end{figure}

\begin{figure}[h]
    \begin{center}
    \includegraphics[width=0.83\textwidth]{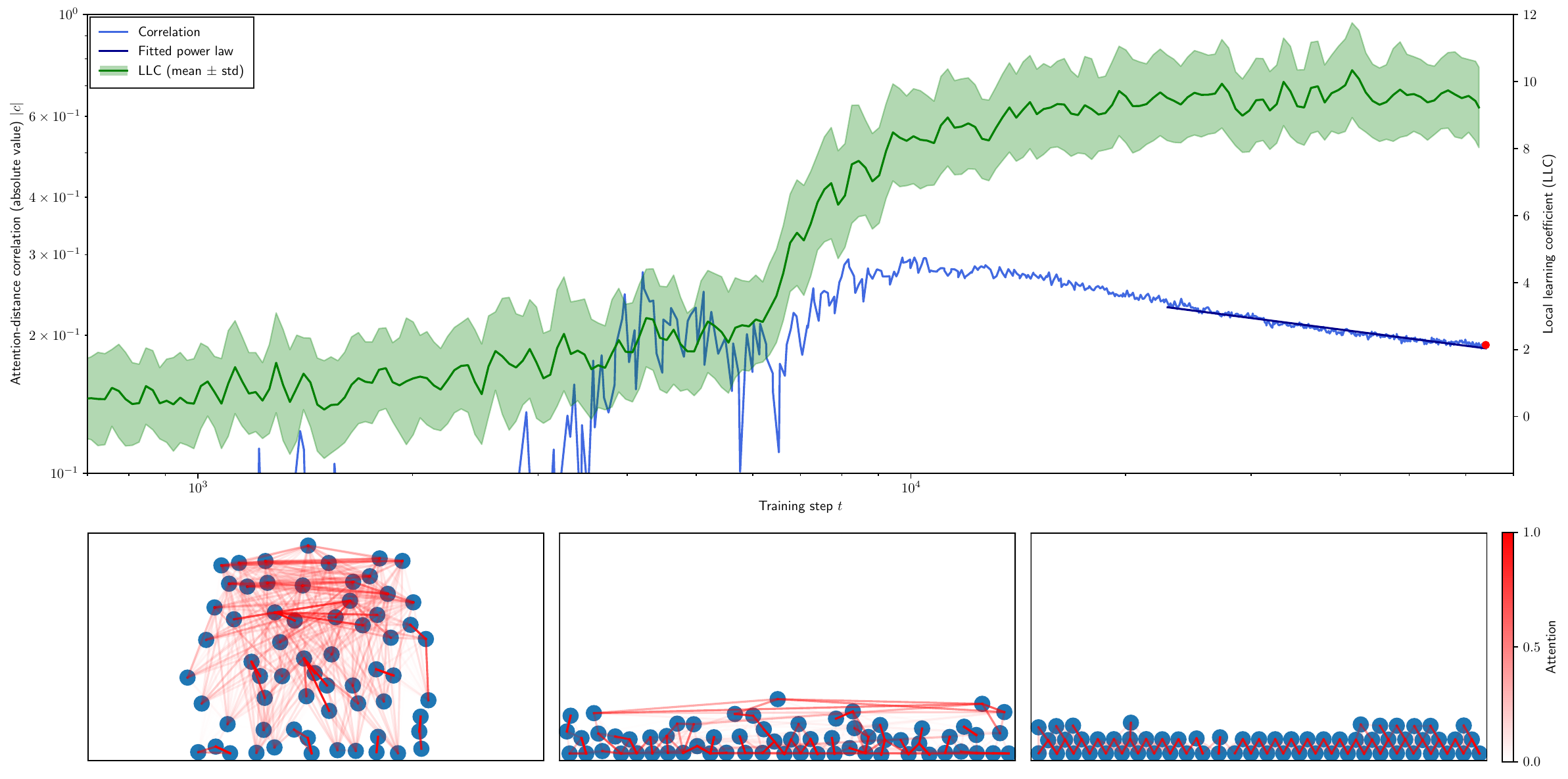}
    \end{center}
    \caption{Attention-distance correlation and local learning coefficient results for head 3-1 in training run 4. The fitted curve is $\log(|c|) = -0.201 \log(t) +0.545$ with $R^2 = 0.953$ (see \cref{table:power law exponents}).} \label{fig: extra 4-3-1}
\end{figure}

\begin{figure}[h]
    \begin{center}
    \includegraphics[width=0.83\textwidth]{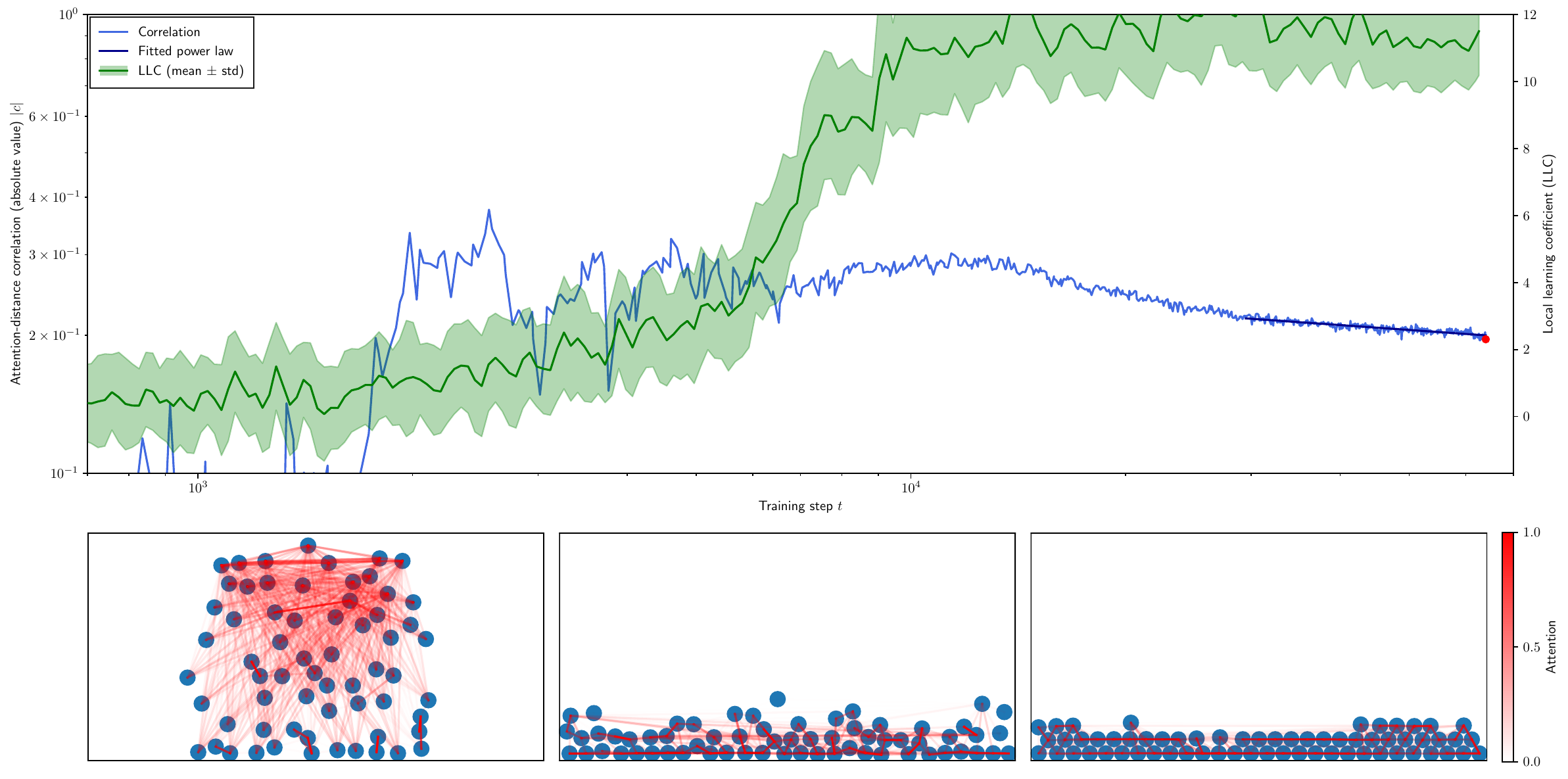}
    \end{center}
    \caption{Attention-distance correlation and local learning coefficient results for head 3-2 in training run 4. The fitted curve is $\log(|c|) = -0.110 \log(t) -0.394$ with $R^2 = 0.775$ (see \cref{table:power law exponents}).} \label{fig: extra 4-3-2}
\end{figure}

\begin{figure}[h]
    \begin{center}
    \includegraphics[width=0.83\textwidth]{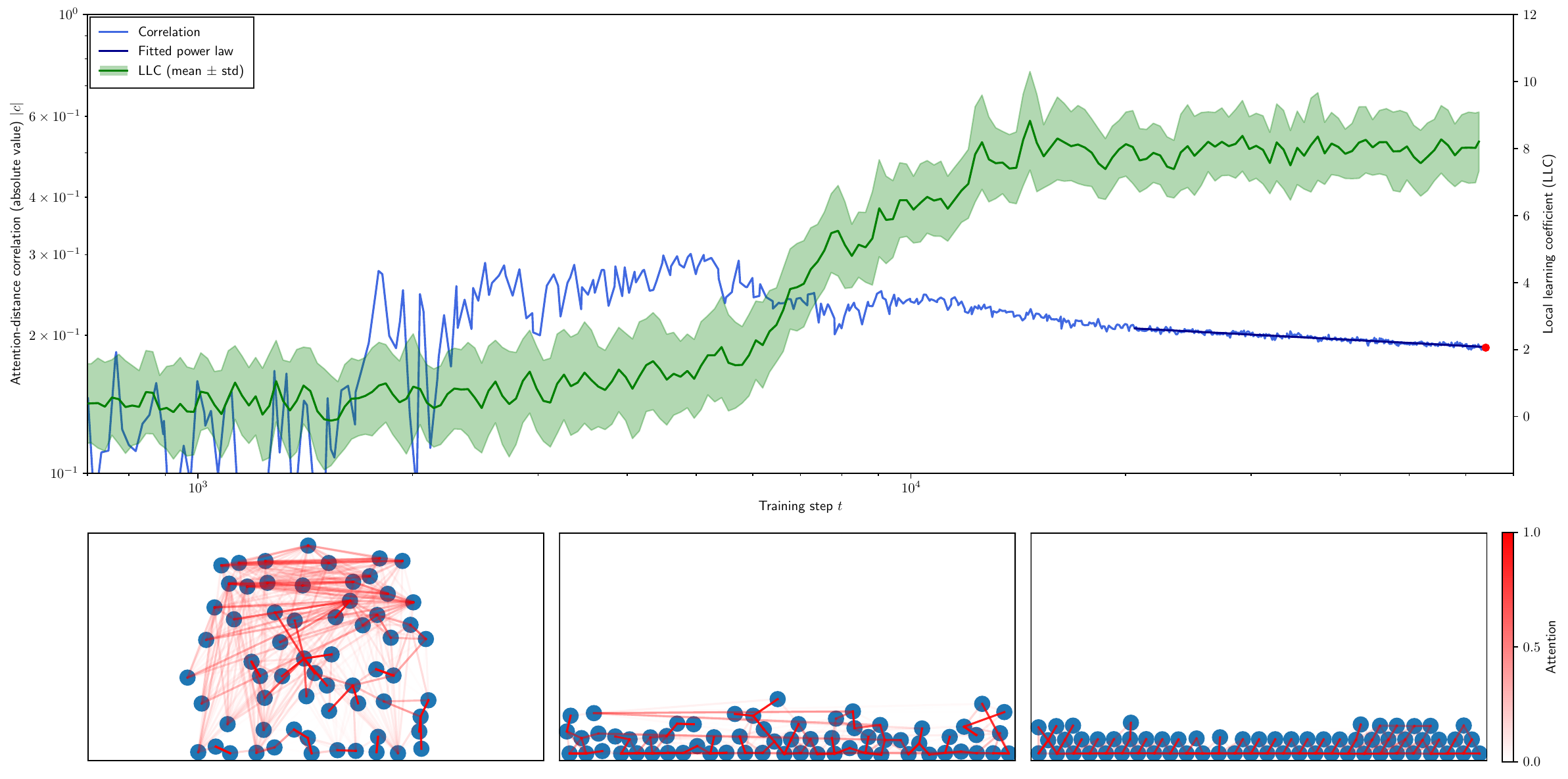}
    \end{center}
    \caption{Attention-distance correlation and local learning coefficient results for head 3-5 in training run 4. The fitted curve is $\log(|c|) = -0.082 \log(t) -0.760$ with $R^2 = 0.911$ (see \cref{table:power law exponents}).} \label{fig: extra 4-3-5}
\end{figure}

\begin{figure}[h]
    \begin{center}
    \includegraphics[width=0.83\textwidth]{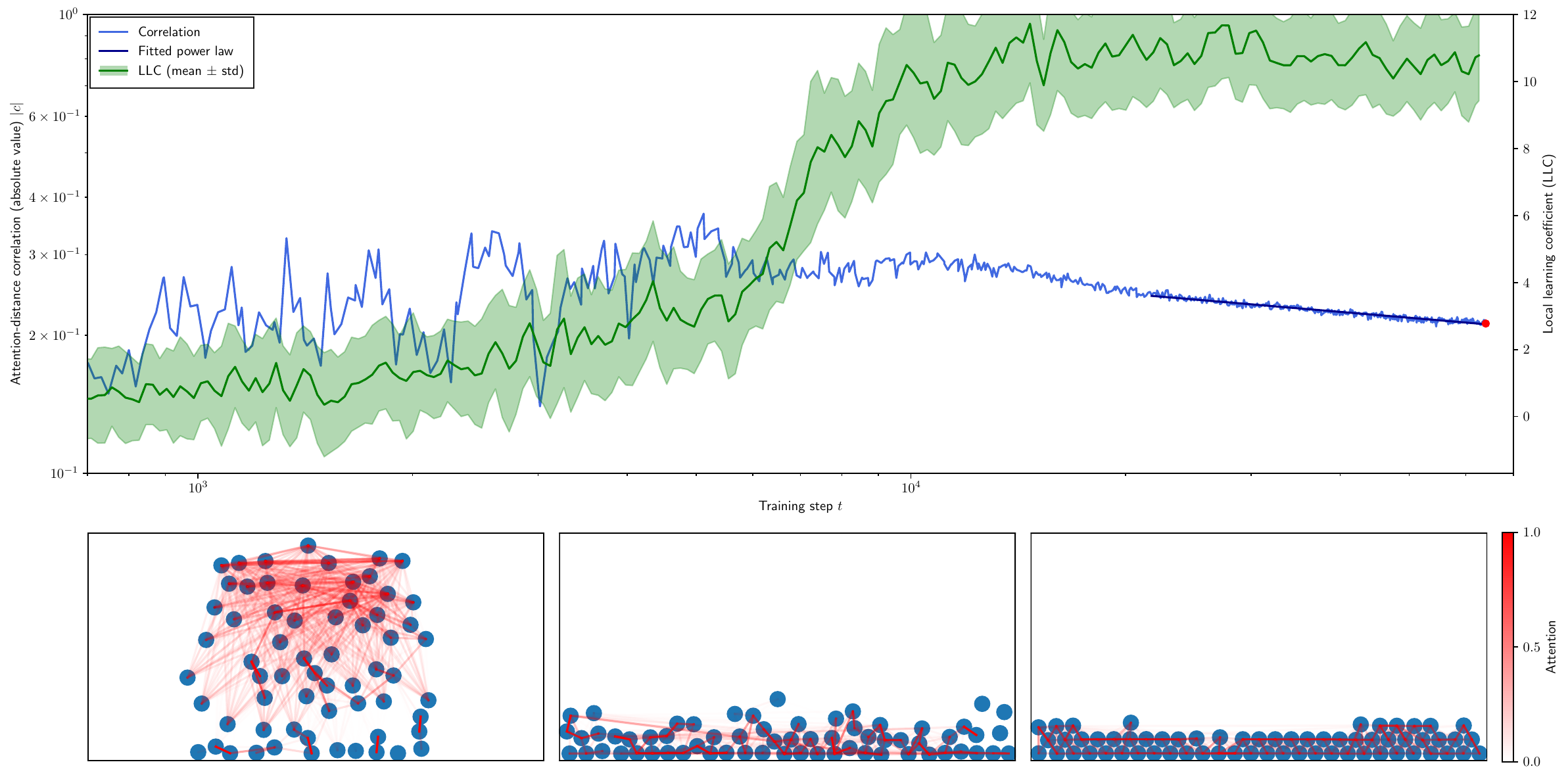}
    \end{center}
    \caption{Attention-distance correlation and local learning coefficient results for head 3-7 in training run 4. The fitted curve is $\log(|c|) = -0.132 \log(t) -0.094$ with $R^2 = 0.930$ (see \cref{table:power law exponents}).} \label{fig: extra 4-3-7}
\end{figure}

\end{document}